\documentclass[10pt,twocolumn,letterpaper]{article}

\usepackage{cvpr}              

\newcommand{\name}{\textit{ObjectMorpher}\xspace}

\usepackage{multirow}
\usepackage{siunitx}
\usepackage{setspace} 
\usepackage{comment}
\usepackage{colortbl}  
\usepackage{amsmath}

\definecolor{cvprblue}{rgb}{0.21,0.49,0.74}
\usepackage[pagebackref,breaklinks,colorlinks,allcolors=cvprblue]{hyperref}

\def\paperID{10450} 
\def\confName{CVPR}
\def\confYear{2026}

\title{ObjectMorpher: 3D-Aware Image Editing via Deformable 3DGS}

\author{
\textbf{Yuhuan Xie}\textsuperscript{*}
\ 
\textbf{Aoxuan Pan}\textsuperscript{*}
\ 
\textbf{Yi-Hua Huang}\textsuperscript{$\dagger$}
\ 
\textbf{Chirui Chang}\textsuperscript{}
\ 
\textbf{Peng Dai}\textsuperscript{}
\
\textbf{Xin Yu}\textsuperscript{}
\
\textbf{Xiaojuan Qi}\textsuperscript{$\ddagger$} \\
\textsuperscript{} The University of Hong Kong
}

\begin{document}

\twocolumn[{%
\renewcommand\twocolumn[1][]{#1}%
\maketitle

\begin{center}
    \centering
    \vspace{-5mm}
    \captionsetup{type=figure}
    \includegraphics[width=1\linewidth, trim=0 0 0 0, clip]{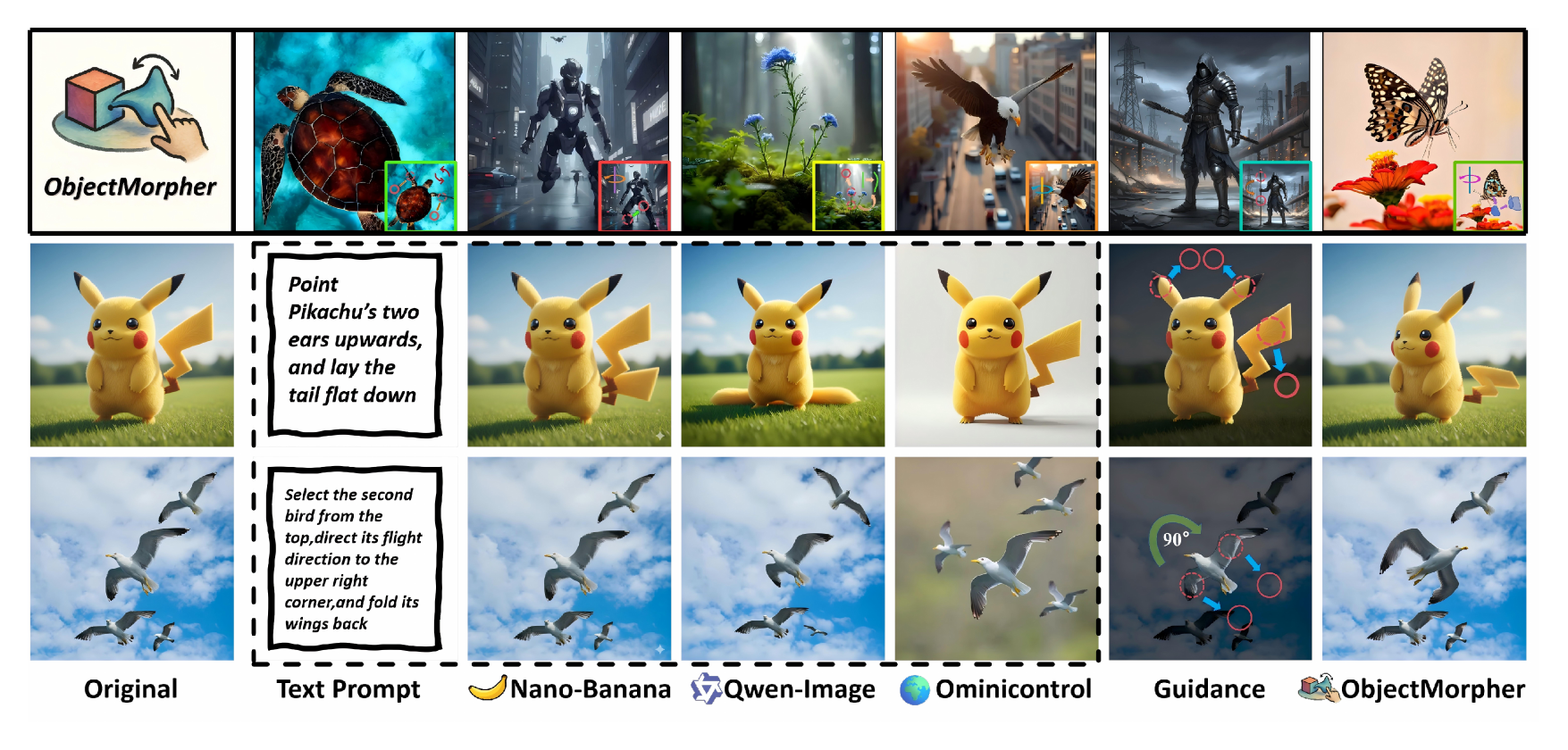}
    \vspace{-8mm}
    \captionof{figure}{Unlike text-based methods that fail to localize subjects or interpret geometry, \textbf{ObjectMorpher} uses \textit{direct 3D manipulation} with real-time interaction. 
This ensures precise edits while preserving the object's identity and background.}
    \vspace{-1mm}
\label{fig:teaser}
\end{center}
}]

\if TT\insert\footins{\noindent\footnotesize{
    \begin{tabular}{@{}l@{\hspace{0.5em}}l}
        $^*$ & Equal contribution. \\
        $^\dagger$ & Project leader. \\
        $^\ddagger$ & Corresponding author.
    \end{tabular}
}}\fi

\begin{abstract}

Achieving precise, object-level control in image editing remains challenging: 2D methods lack 3D awareness and often yield ambiguous or implausible results, while existing 3D-aware approaches rely on heavy optimization or incomplete monocular reconstructions. We present \name, a unified, interactive framework that converts ambiguous 2D edits into geometry-grounded operations. \name\ lifts target instances with an image-to-3D generator into editable 3D Gaussian Splatting (3DGS), enabling fast, identity-preserving manipulation. Users drag control points; a graph-based non-rigid deformation with as-rigid-as-possible (ARAP) constraints ensures physically sensible shape and pose changes. A composite diffusion module harmonizes lighting, color, and boundaries for seamless reintegration. Across diverse categories, \name\ delivers fine-grained, photorealistic edits with superior controllability and efficiency, outperforming 2D drag and 3D-aware baselines on KID, LPIPS, SIFID, and user preference.

\end{abstract}    

\section{Introduction}
\label{sec:intro}

Image editing \cite{Li2022MATMT, Chen2023AnyDoorZO, Song2023ObjectStitchOC, Yu2025ObjectMoverGO, Song2024IMPRINTGO, Winter2024ObjectDropBC} is an important task for generative visual intelligence, enabling artists, designers, and everyday users to manipulate and recompose visual content with increasing precision. Recent foundation image editing models such as Qwen-Image~\cite{wu2025qwen}, Nano-Banana~\cite{google2025nanobanana}, and PixArt-$\alpha$~\cite{chen2023pixartalpha, chen2024pixartdelta} have demonstrated strong capabilities in prompt-based generation and instruction-driven modification. However, despite their impressive fidelity, achieving fine-grained, object-level manipulation- for instance, adjusting the pose of a person, rotating a chair, or bending a flower stem-remains a major challenge as Fig. \ref{fig:teaser} shows. The difficulty stems from the fact that these systems predominantly operate in 2D pixel space, where spatial coherence, geometric consistency, and physical plausibility are difficult to maintain during complex transformations.

A growing body of work ~\cite{pan2023draggan,shi2024dragdiffusion,wu2024draganything, Yu2025ObjectMoverGO, shin2025motionstream} seeks to provide more direct control over local regions. Drag-based editing methods such as DragGAN~\cite{pan2023draggan} and DragDiffusion~\cite{shi2024dragdiffusion} allow users to specify point correspondences to deform image content interactively. Yet, these techniques remain confined to 2D space, often producing ambiguous effects when users intend 3D operations like rotation or depth translation (Fig.\ref{fig:qualitative comparisons}).  3D-aware extensions: OBJECT-3DIT~\cite{Michel2023OBJECT3L}, Neural Assets~\cite{wu2024neural}, Image Sculpting~\cite{Yenphraphai2024ImageSP}, and BlenderFusion~\cite{Chen2025BlenderFusion3V}, inject geometric priors yet still fall short: OBJECT-3DIT and Neural Assets provide only pose-conditioned rigid 6-DoF control, while Image Sculpting relies on slow per-image SDS/textual inversion~\cite{poole2023dreamfusion}, limiting precision and efficiency and requiring over half an hour per image. BlenderFusion further relies on monocular meshes, which are usually incomplete, thereby curtailing large pose changes and deformations.  As a result, these pipelines are compute-heavy, fragile for non-rigid edits, and prone to compositional artifacts. None demonstrates robust, efficient, object-level non-rigid editing.

Three core challenges persist.
(i) Lack of 3D awareness: Purely 2D methods cannot disambiguate geometric transformations, making it difficult to maintain object identity under pose change; 
(ii) Physical plausibility: Without explicit constraints, edits often introduce unnatural distortions or depth inconsistencies;
and (iii) Compositional coherence: After editing, seamlessly reintegrating the modified object back into its original context requires consistent lighting, color, and boundary blending-- an aspect rarely addressed jointly with manipulation. Overcoming these issues demands an image editing paradigm that unifies geometric control, physical realism, and visual harmony within a unified, interactive framework.

We introduce  \name, a 3D-aware image-editing framework expressly designed to resolve the above challenges.  Our core idea is to turn inherently ambiguous 2D edits into geometry-aware operations by lifting objects to an editable 3DGS~\cite{kerbl20233d} proxy, applying physically constrained non-rigid control, and then diffusion-based compositing back into the image for seamless realism.

Concretely, we elevate selected 2D instances into 3D using an foundation image-to-3D generator~\cite{xiang2025structured} to reconstruct each target as a high-quality 3D Gaussian representation, which disambiguates pose/rotation/depth and preserves identity while remaining fast to render and friendly for local parameter editing, directly tackling the lack of 3D awareness.
We then enable intuitive, part-aware manipulation by letting users drag sparse control points in real time; a graph-based non-rigid deformation module with as-rigid-as-possible (ARAP) constraints enforces physical plausibility and prevents unnatural distortions.
Finally, the edited object is re-rendered and harmonized with the original photo via a composite diffusion model, paired with background inpainting where needed, to align lighting, color, and boundaries for photorealistic reintegration, ensuring the final image's compositional coherence.

Extensive experiments across diverse object categories validate the effectiveness of \name. Our method is overwhelmingly preferred by users for its superior controllability and realism (Fig.~\ref{fig:user_study_chart}). 
This is a crucial finding, as quantitative metrics like LPIPS (Table~\ref{tab:baseline}) often fail to capture true editing capability, instead rewarding methods that make minimal changes. \name\ demonstrates the best overall balance: it achieves top-tier user-rated "Guidance Following" (Fig.~\ref{fig:user_study_chart}) and real-time interaction ($<$10s) while maintaining competitive high fidelity (Table~\ref{tab:baseline}), a comprehensive trade-off that baselines fail to solve.  

Our contributions are four-fold:
\begin{itemize}
\item We introduce \name, a 3D-aware image-editing framework that integrates 3D geometric control, physical realism, and visual coherence within a unified pipeline.
\item We propose a 3D Gaussian Splatting (3DGS)–based representation for editable object lifting and deformation, leveraging its flexibility and differentiability for fast, identity-preserving manipulation guided by physically constrained non-rigid control.
\item  We develop a composite diffusion module that harmonizes the edited object with its scene context, achieving photorealistic composition in lighting, color, boundary, and background consistency. 
\item We conduct extensive experiments across diverse categories and edit types, reporting quantitative and qualitative gains, human preference wins in controllability and plausibility, and robustness to large pose changes and non-rigid deformations, consistently outperforming state-of-the-art baselines in both quality and efficiency.
\end{itemize}

\begin{figure*}[t]
\begin{center}
\vspace{-2mm}
\includegraphics[width=.95\linewidth]{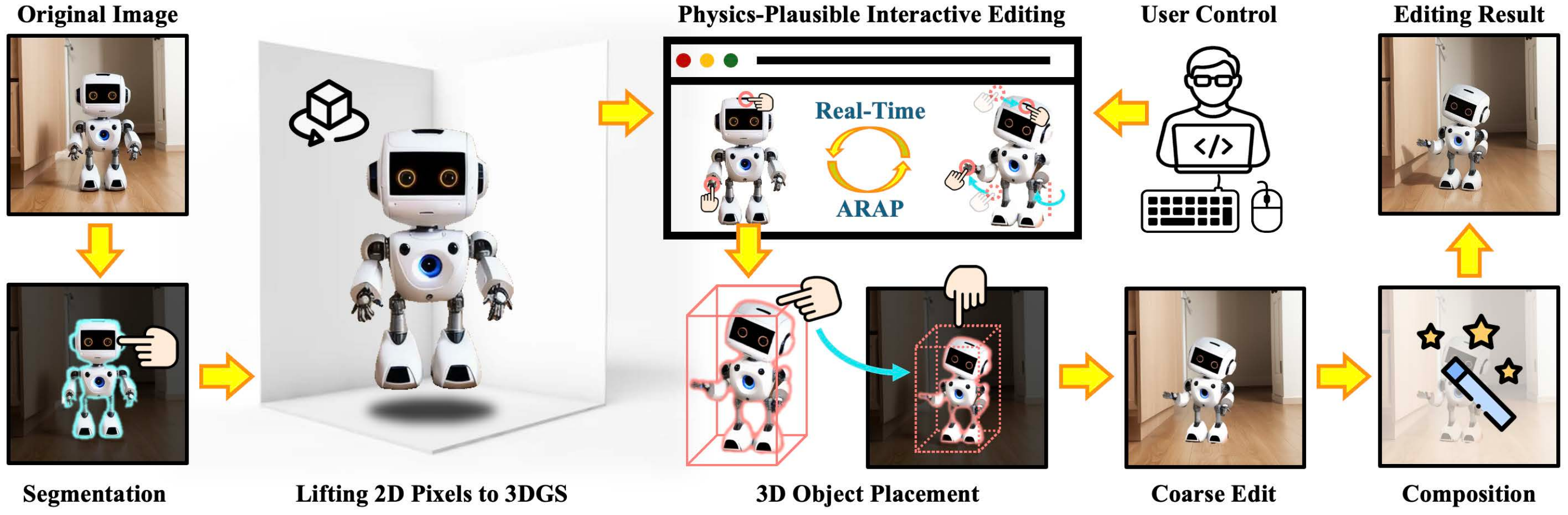}
\vspace{-6mm}
\end{center}
\caption{\textbf{Overview of our image editing pipeline.} The object is lifted from 2D pixels to high-fidelity 3DGS. Real-time editing with local rigidity is applied based on user input. The object is then repositioned, and a generative model refines the edits for harmonious results.}
\label{fig:pipeline}
\vspace{-2mm}
\end{figure*}

\section{Related Work}
\label{sec:related_work}

\noindent \textbf{2D Image Editing.} 2D image editing has made rapid progress with the success of diffusion models \cite{ho2020denoising,sohl2015deep}. Early works focus on localized operations such as object insertion, removal, and movement \cite{avrahami2022blended, suvorov2021resolution, lugmayr2022repaint, meng2021sdedit, saharia2022palette, lu2023tficon, zhang2023controlcom, anydoor, paint_by_example, objectstitch, imprint}. Recent advances in multimodal large language models (MLLMs) such as GPT-4o \citep{hurst2024gpt} and Gemini-2.5-Flash-Image (Nano-Banana) \citep{comanici2025gemini} further show strong fidelity and generalization, allowing users to modify attributes, replace objects, or restyle regions with high realism directly from textual instructions.
These methods work well largely because they operate under strong 2D priors: the model hallucinates plausible pixels consistent with the surrounding context, and in many cases can even preserve identity or appearance when introducing new content \cite{qian2025pico}. However, this paradigm implicitly assumes that the edited object does not change its geometric relationship with the scene. When the user intent involves precise spatial deformation, purely 2D methods must infer missing 3D structure. This often leads to artifacts such as texture stretching, incorrect foreshortening, or shading and shadow inconsistencies \cite{sarkar2024shadows, chang2025far}, since the model has no explicit 3D representation of objects.

Interactive drag-based approaches \cite{drag_anything, pan2023drag, mou2023dragondiffusion} let users place sparse control points for intuitive 2D deformations, but because they still edit in image space, 3D-intended motions easily break rigidity or semantics. Recent data-driven methods \cite{michel2023object, yu2025objectmover} alleviate boundary and illumination inconsistencies, but still do not guarantee geometry-aware object deformation when the configuration changes in 3D.
Our work targets this gap: we keep the user-driven flavor of 2D editing, but lift the target instance into an explicit 3D representation, apply geometry-aware deformation to realize the intended spatial change, and then use a generative module to composite it back. Thus edits that are ill-posed in pure 2D become well-defined geometric operations, setting us apart from prior 2D-only methods.

\vspace{0.1in}\noindent\textbf{3D-Aware Editing.}
Recent approaches enable 3D editing through natural language instructions. Instruct-NeRF2NeRF~\cite{haque2023instruct} enables modifying NeRF scenes according to textual commands, while GO-NeRF~\cite{dai2025go} synthesizes and inserts new objects into NeRFs based on descriptive prompts. Object 3DIT~\cite{Michel2023OBJECT3L} further leverages pretrained diffusion models to achieve language-driven 3D-aware editing. Although these methods offer flexible semantic control, the ambiguity of text supervision often results in limited precision, weak viewpoint consistency, and unintended changes to geometry or appearance.
To improve controllability, another line of work converts implicit scene representations into explicit intermediates, such as meshes or geometric proxies. NeRF-Editing~\cite{yuan2022nerf} performs local geometric deformation through proxy-based constraints, and NeuMesh~\cite{yang2022neumesh} reconstructs editable mesh surfaces to disentangle geometry and texture. 3D-FixUp~\cite{cheng20253d} and BlenderFusion~\cite{chen2025blenderfusion} integrate 3D priors or professional modeling software to support user-guided refinements, while GeoDiffuser~\cite{sajnani2025geodiffuser} and Shape-for-Motion~\cite{liu2025shape} enforce geometric coherence in image or video editing.

Despite improved edit accuracy and interactive manipulation, these methods inherently lose high-frequency details during the mesh conversion process, and often limit deformation to rigid and small transformations. More recently, Gaussian-based rendering has enabled high-fidelity and fast 3D manipulation. DragGaussian~\cite{shen2024draggaussian} introduces drag-style control for adjusting object poses within Gaussian splats, and SC-GS~\cite{huang2024sc} incorporates sparse constraints to flexibly modify dynamic content. While these methods preserve appearance detail and allow smooth updates, they primarily operate on isolated objects and do not explicitly address seamless integration with surrounding environments. In contrast, our method provides a holistic 3D editing framework that jointly considers editing precision, visual fidelity, interaction efficiency, and harmonious integration with the broader scene, ensuring that edited objects remain structurally plausible and naturally embedded within their original environments.


\section{Method}
Our goal is to achieve flexible image editing with \textbf{(1)} precise 3D control, \textbf{(2)} high-fidelity rendering, \textbf{(3)} physical plausibility, \textbf{(4)} real-time interaction, and \textbf{(5)} visually harmonious results. At the core lies the ability to manipulate 2D image objects directly within a 3D space. We accomplish this by lifting 2D instances into 3D through a diffusion-based generation model~\cite{xiang2025structured}, which reconstructs each object as a high-quality 3D Gaussian Splatting (3DGS) representation~\cite{kerbl20233d}.
Building upon this editable 3D representation, we propose an ARAP-constrained~\cite{sorkine2007rigid} non-rigid editing framework that performs physically plausible deformations, treating each object as a deformable rigid body. The edited 3D object is composited with an inpainted background to form a coarse edited result, followed by refinement using a composite diffusion model that ensures seamless blending and visual coherence with the original scene.  An overview of our pipeline is shown in Fig~\ref{fig:pipeline}.

In this section, we detail the 2D-to-3D lifting process in Sec.~\ref{sec:2d3dlifting}, the 3DGS-based rigid-constrained editing in Sec.~\ref{sec:arap}, and the composite diffusion model in Sec.~\ref{sec:composite}.

\subsection{Lifting 2D Pixels to 3DGS}
\label{sec:2d3dlifting}

Unlike conventional 2D editing methods based on GANs~\cite{goodfellow2020generative} or diffusion models~\cite{ho2020denoising}, our approach enables manipulation of both 6-DoF pose and non-rigid shape within 3D space. Given an input image, users select target objects interactively via click prompts, which are segmented by SAM~\cite{kirillov2023segment}. The cropped object regions are then processed by the off-the-shelf 3D generation model TRELLIS~\cite{xiang2025structured}, which reconstructs the object as a 3D Gaussian Splatting (3DGS) representation~\cite{kerbl20233d}.  We choose the 3DGS output over the mesh decoder of TRELLIS for its better fidelity and editability: the mesh decoder, post-trained with a frozen encoder, is optimized mainly for geometric normal consistency and limited color resolution, whereas the 3DGS decoder jointly optimizes appearance and geometry through volumetric rendering loss, achieving higher realism and smoother gradients for manipulation (Fig.~\ref{fig:ablation on representations}).

The 3DGS representation consists of a set of Gaussian primitives, denoted as ${\mathcal{G}_i : (\mu_i, o_i, s_i, q_i, c_i)}$. Here, $\mu_i \in \mathbb{R}^3$ specifies the position of the Gaussian center, $o_i \in \mathbb{R}^+$ represents the opacity, $s_i \in \mathbb{R}^{+3}$ defines the scaling factors along the three axes in 3D space, $q_i \in \mathbb{R}^4$ is the quaternion of the $SO(3)$ Gaussian rotation, and $c_i \in \mathbb{R}^3$ is the Gaussian color.
The rendering process for 3DGS applies the EWA projection~\cite{zwicker2002ewa}. First, the 3D Gaussians are projected onto the image plane using the projection matrix, resulting in 2D Gaussians. Then each pixel color is computed through $\alpha$-composition of the 2DGS ordered by depth.

\subsection{Physics-Plausible Interactive Editing}
\label{sec:arap}

Given the lifted 3DGS representation $\mathcal{G}_i$, we enable real-time object editing that respects physical constraints using As-Rigid-As-Possible (ARAP) deformation~\cite{sorkine2007rigid}. ARAP preserves local rigidity while allowing intuitive, part-aware non-rigid transformations. However, since high-fidelity 3DGS models contain millions of Gaussians, direct optimization is infeasible in real time. 

To address this, we introduce a sparse editing proxy for 3DGS. We first sample 512 control points via farthest-point sampling on the Gaussian cloud, yielding an even, structure-aware subset that forms a deformable graph. This proxy faithfully captures global geometry while remaining sufficiently lightweight for interaction. Users edit the proxy directly; a rigidity-constrained solver then estimates per-node affine motions (translation, rotation, and scale) to produce the deformed graph. Finally, these motions are propagated to the full 3DGS-- serving as guidance for the dense Gaussians-- enabling efficient, high-fidelity non-rigid editing with real-time responsiveness.

\subsubsection{Deformable Graph Construction} 
\label{sec:graph}
\begin{figure}[hb]
\begin{center}
\vspace{-2mm}
\includegraphics[width=\linewidth]{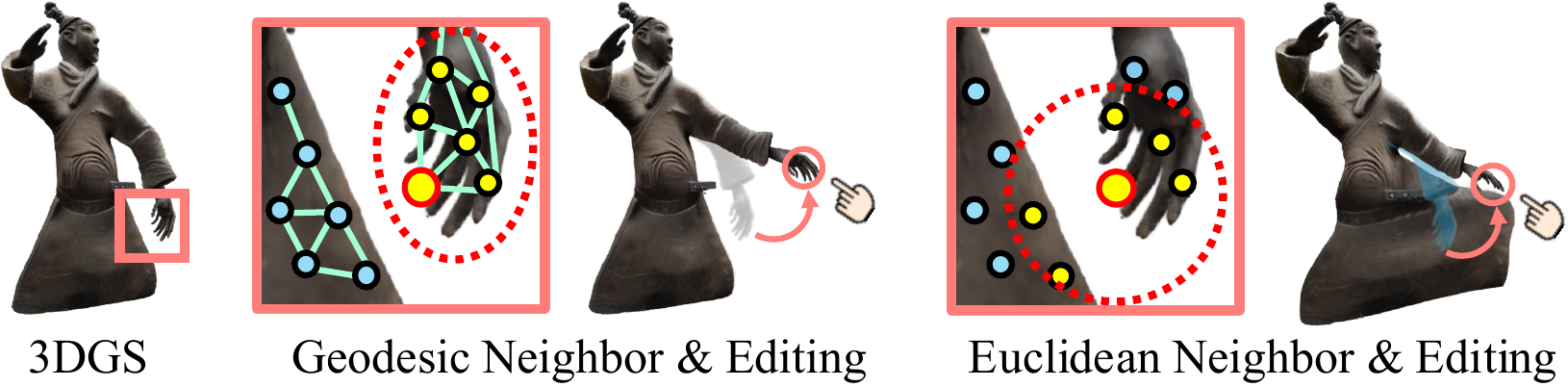}
\vspace{-10mm}
\end{center}
\caption{Comparison of graph connections based on Euclidean distance and geodesic distance.}
\label{fig:geodesic}
\vspace{-2mm}
\end{figure}

Given the sampled control points ${p_i}$, we build a manifold-aware graph that supports rigidity-constrained editing. To guarantee local smoothness and stable deformation, each node must have sufficiently many neighbors-- both for graph optimization and for transferring deformations to the dense 3DGS. We therefore target a connectivity radius proportional to object scale, linking candidates within $0.3D_{\text{scene}}$, where $D_{\text{scene}}$ is the scene diameter (the maximum pairwise control-point distance). Direct Euclidean linking at this radius, however, ignores the surface manifold and can spuriously connect nodes across distant parts (see Fig.~\ref{fig:geodesic}). To respect geometry, we first construct an \emph{auxiliary local graph} by connecting points within a shorter radius $2\bar d_{\text{NN}}$ (with $\bar d_{\text{NN}}$ the mean nearest-neighbor distance). We then estimate \emph{geodesic distances} as shortest paths on this auxiliary graph using the Floyd algorithm~\cite{floyd1962algorithm}. Finally, we form the \emph{deformable graph} by connecting pairs whose \emph{geodesic} distance is below $0.3 D_{\text{scene}}$.
This two-stage procedure preserves local continuity, avoids cross-part shortcuts, and yields a topology that provides robust support for smooth, well-conditioned deformations (Fig.~\ref{fig:geodesic}).

\subsubsection{Rigidity-Constrained Deformation}
Let $\mathbf{p}_i$ and $\mathbf{p}'_i$ denote the original and deformed positions of the $i$-th control point, respectively. During interactive editing, users directly manipulate the 3D control points through a GUI that visualizes the real-time rendered 3DGS. By clicking and dragging, users can reposition $H$ control points to new target positions $\tilde{\mathbf{p}}_{h_i}$, enforcing the hard constraints: 
\begin{equation}
\mathbf p'_{h_i} = \tilde{\mathbf p}_{h_i}, \quad i \in [H],
\end{equation}
where $h_i$ indexes the selected handle points. The remaining unconstrained points are optimized for physically plausible, smooth deformation matching the user’s manipulation. 

\vspace{0.1in}\noindent\textbf{ARAP regularization.} We adopt an \textit{as-rigid-as-possible} (ARAP) deformation model to preserve local shape rigidity while allowing intuitive, non-rigid manipulation. The energy function is defined as:
\begin{equation}
 E(\mathbf p_i', \mathbf R_i) = \sum\limits_i \sum\limits_{j\in \mathcal{N}i} w_{ij} | (\mathbf p'_i - \mathbf p'_j) - \mathbf{R}_i (\mathbf p_i - \mathbf p_j) |^2,
\label{eq:arap_energy}
\end{equation}
where $\mathbf{R}_i$ is the local rotation matrix centered at $\mathbf{p}_i$, $\mathcal{N}i$ denotes its neighborhood, and $w_{ij}$ is a distance-decayed weight. This formulation is invariant to global translation and rotation, penalizing only non-rigid distortions. The optimization alternates between updating $\mathbf{p}'$ (the deformed positions) and estimating $\mathbf{R}_i$ (the local rotations) until convergence~\cite{moon1996expectation}. In practice, three iterations are generally sufficient for visually stable results.


\vspace{0.1in}\noindent\textbf{Position optimization.} Fixing $\mathbf{R}_i$, differentiating Eq.~\eqref{eq:arap_energy} w.r.t. $\mathbf{p}'_i$ yields the linear system:
\begin{equation}
{\mathbf L}\mathbf p' = \mathbf b,
\label{eq:laplacian}
\end{equation}
where $\mathbf b = [\sum\limits_{j \in \mathcal{N}i} \frac{w_{ij}}{2} (\mathbf R_i + \mathbf R_j)(\mathbf p_i - \mathbf p_j)]^T$ is a constant vector, and $\mathbf{L}$ is the Laplacian matrix encoding neighborhood connectivity , which is derived in Sec. \ref{sec:graph}: the deformable graph. 
To respect the user-specified handle constraints, the corresponding rows and columns of $\mathbf{L}$ (and entries in $\mathbf{b}$) are removed. The reduced system is solved using the inverse or pseudo-inverse of $\mathbf{L}$ to obtain the updated unconstrained positions $\mathbf{p}'$.

\vspace{0.1in}\noindent\textbf{Rotation estimation.} 
Given the updated positions, the local rotations $\mathbf{R}_i$ are refined by solving:
\begin{equation}
\mathbf S_i = \sum\limits_{j\in\mathcal{N}i} w_{ij}(\mathbf p_j - \mathbf p_i)^T(\mathbf p'_j - \mathbf p'_i).
\end{equation}
The above is further solved by Singular Value Decomposition (SVD)~\cite{sorkine2017least}: $\mathbf S_i = \mathbf U_i\Sigma_i \mathbf V_i^T$, the local rotation $\mathbf R_i$ is computed as $\mathbf R_i = {\mathbf V}_i{\mathbf U}_i^T$.

\paragraph{Dense Gaussian deformation.} 
After optimizing the sparse control graph, the deformation is propagated to all Gaussians in the dense 3DGS via \textit{Linear Blend Skinning} (LBS)~\cite{magnenat1989joint}. Each Gaussian parameter is updated as: 
\begin{equation}
    \begin{aligned}
        \mu'_i &= \sum\limits_{j\in \tilde{\mathcal{N}}_i} \tilde{w}_{i,j}\left(\mathbf R_j \left(\mathbf \mu_i-\mathbf p_j \right) + \mathbf p'_j \right), \\
        \mathbf q'_i  &= \sum\limits_{j\in \tilde{\mathcal{N}}_i} \tilde{w}_{i,j} \text{Quat}\left(\mathbf R_j \cdot \text{Rot}(\mathbf q_i)  \right), \\
        s_i' &= (\sum\limits_{j\in \tilde{\mathcal{N}}_i} \tilde{w}_{i,j} \cdot\mathbb{E}_{k\in\mathcal{N}_j}(\frac{||\mathbf p_j'- \mathbf p_k'||}{||\mathbf p_j-\mathbf p_k||})) s_i. 
    \end{aligned}
\end{equation}
Here, $\tilde{\mathcal{N}}_i$ denotes the $\tilde{K}=4$ nearest control points, and $\tilde{w}_{ij}$ are normalized distance-based weights. Neighboring control points are determined through a two-stage strategy: (1) identify the nearest control by Euclidean distance, and (2) find its remaining $\tilde{K}-1$ neighbors along the geodesic manifold. This hybrid metric preserves topological continuity and enables smooth, structure-aware deformation transfer across the 3DGS, which is critical for visual realism.

\subsection{Generative Composition and Refinement}
\label{sec:composite}

\begin{figure}[h]
\begin{center}
\vspace{-4mm}
\includegraphics[width=\linewidth]{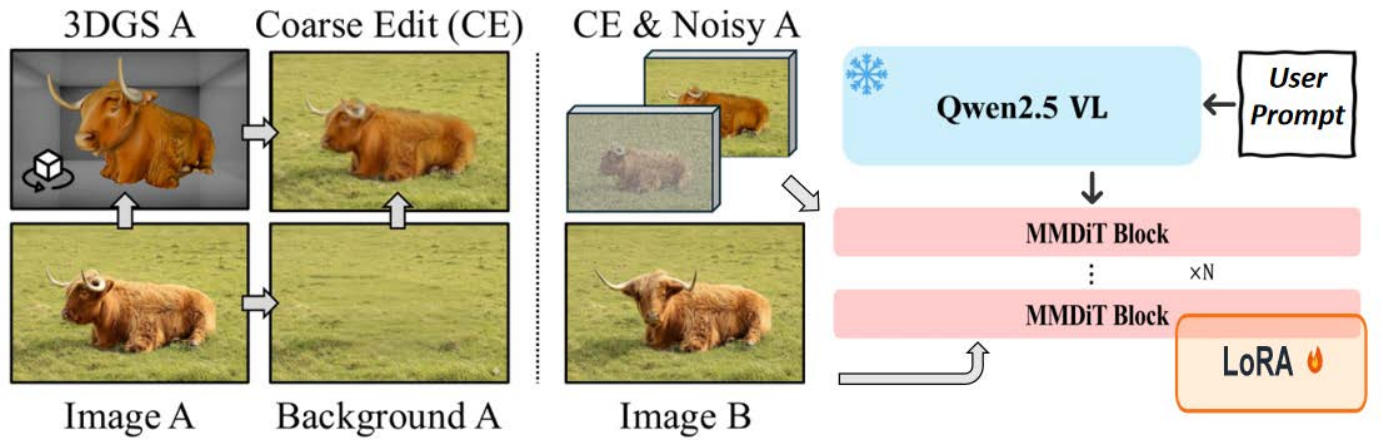}
\vspace{-10mm}
\end{center}
\caption{Illustration of the training data preparation and the pipeline of our generative composition model.}
\label{fig:generative_composition}
\vspace{-6mm}
\end{figure}

After deformation, the edited 3D object is re-rendered into 2D and must be seamlessly reintegrated into the original image. We first inpaint the occluded background region using PixelHacker~\cite{xu2025pixelhacker}, producing a clean background plate. The rendered object is then composited into this plate, followed by a refinement step using a composite diffusion model trained to harmonize lighting, color, and boundaries between object and scene (Fig.~\ref{fig:generative_composition}). 

\vspace{0.1in}\noindent \textbf{Generative Composition.} Our composition module leverages the pre-trained Qwen-Image-Edit (with an MMDiT backbone). Generation is jointly conditioned on two inputs: the original image, which provides robust semantics to preserve identity and lighting, and the coarse edited image, which dictates the geometric layout of the 3D edits. To efficiently harmonize the edited object with the scene without compromising the foundation model's strong generative prior, we fine-tune a task-specific Low-Rank Adaptation (LoRA) module within the attention layers, ensuring seamless and photorealistic integration.


\vspace{0.1in}\noindent \textbf{Training Data Construction.} Training data are constructed from paired images of the same object under different poses and deformations, collected from Subjects200K~\cite{tan2025ominicontrol} and videos generated by KlingAI. Coarse edit pairs are synthesized using TRELLIS~\cite{xiang2025structured} to obtain 3DGS reconstructions, rendered from multiple views, with poses predicted by $\pi^3$~\cite{wang2025pi}. The model learns to translate these coarse composites into photorealistic results aligned with the ground-truth originals, yielding smooth object-- scene integration and lighting consistency. As shown in Fig.~\ref{fig:generative_composition}, the paired triplets consist of the coarse edit A, image B, and image A. The first two serve as conditions for the composition model to generate the last.


\begin{figure*}[t]
    \centering

    \setlength{\tabcolsep}{2pt}
    \def\imgwidth{0.13\textwidth} 

    \begin{tabular}{cccccccc}
    
        & 
        \small Origin &
        \small Guidance &
        \small DragDiffusion &
        \small AnyDoor &
        \small Sculpting &
        \small DragAnything &
        \small Ours \\
        \addlinespace[3pt] 

        \raisebox{14pt}{\rotatebox{90}{\small Butterfly}} & 
        \includegraphics[width=\imgwidth]{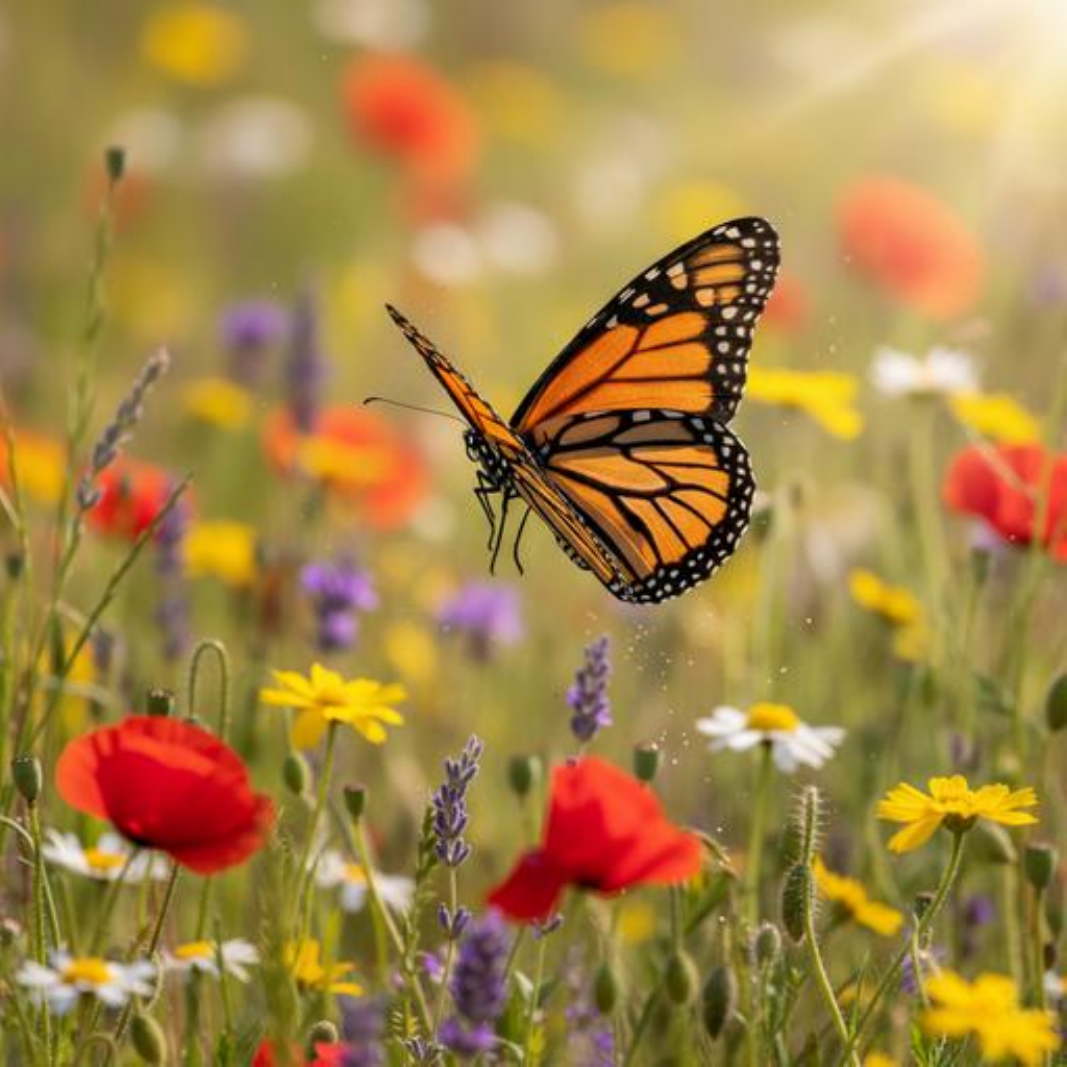} &
        \includegraphics[width=\imgwidth]{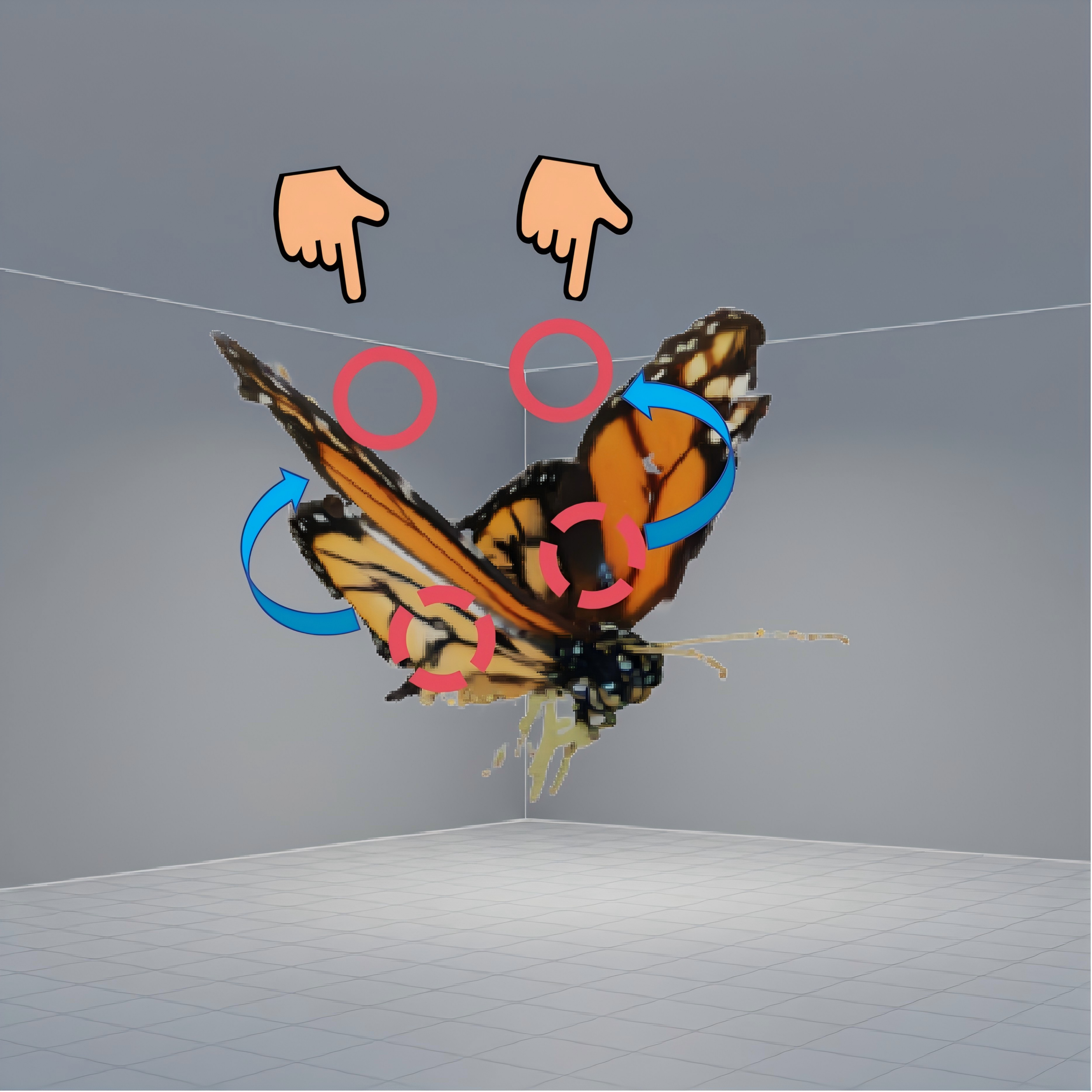} &
        \includegraphics[width=\imgwidth]{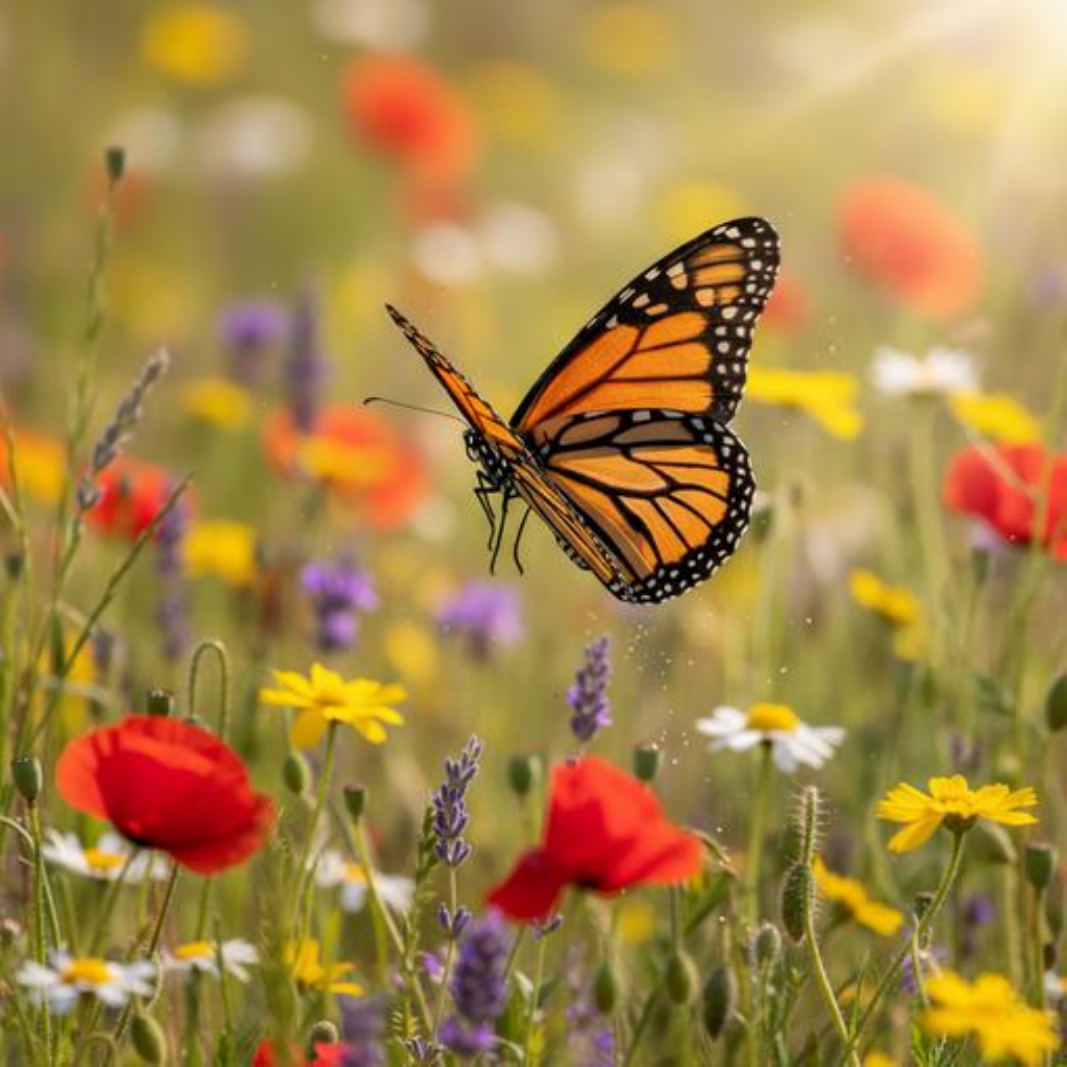} &
        \includegraphics[width=\imgwidth]{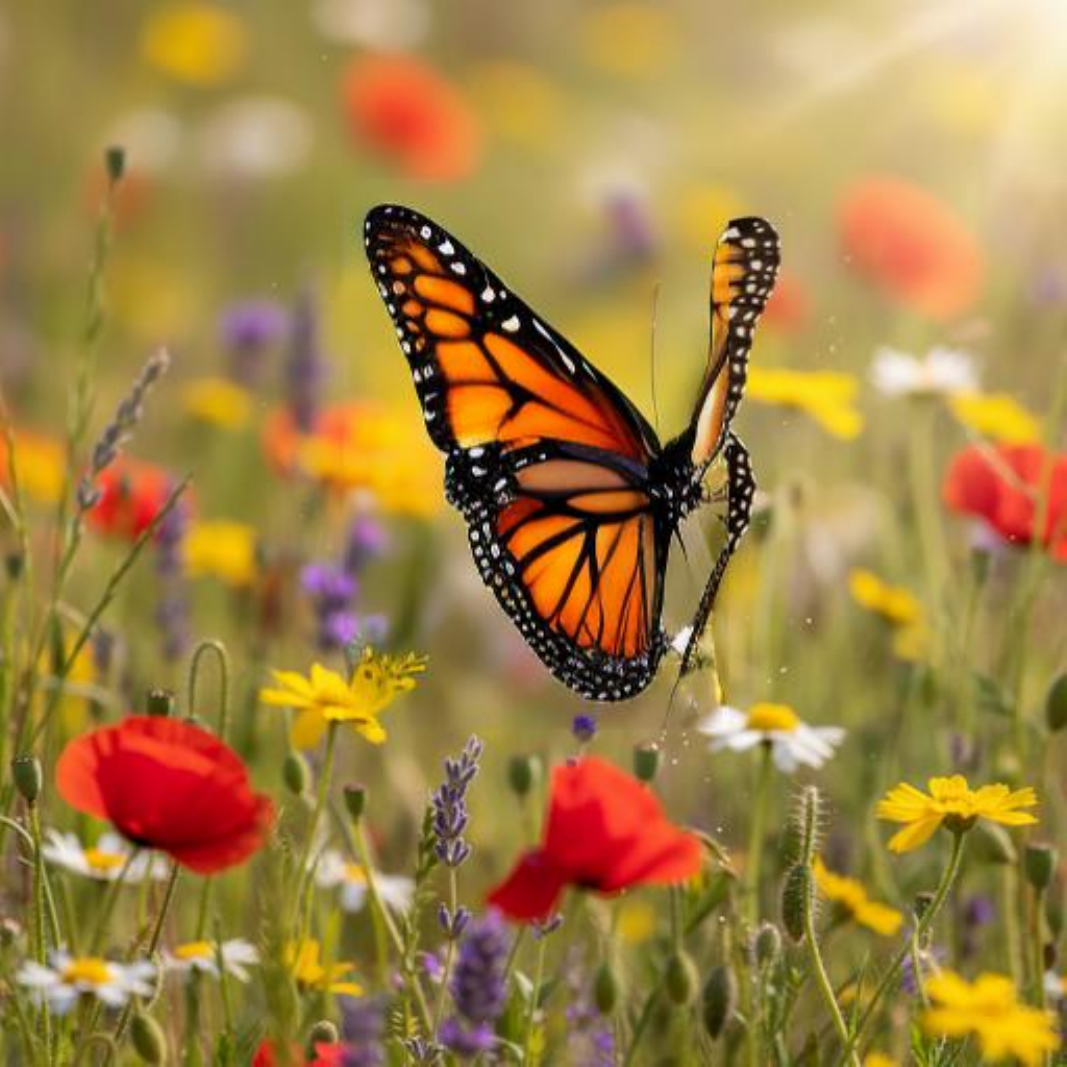} &
        \includegraphics[width=\imgwidth]{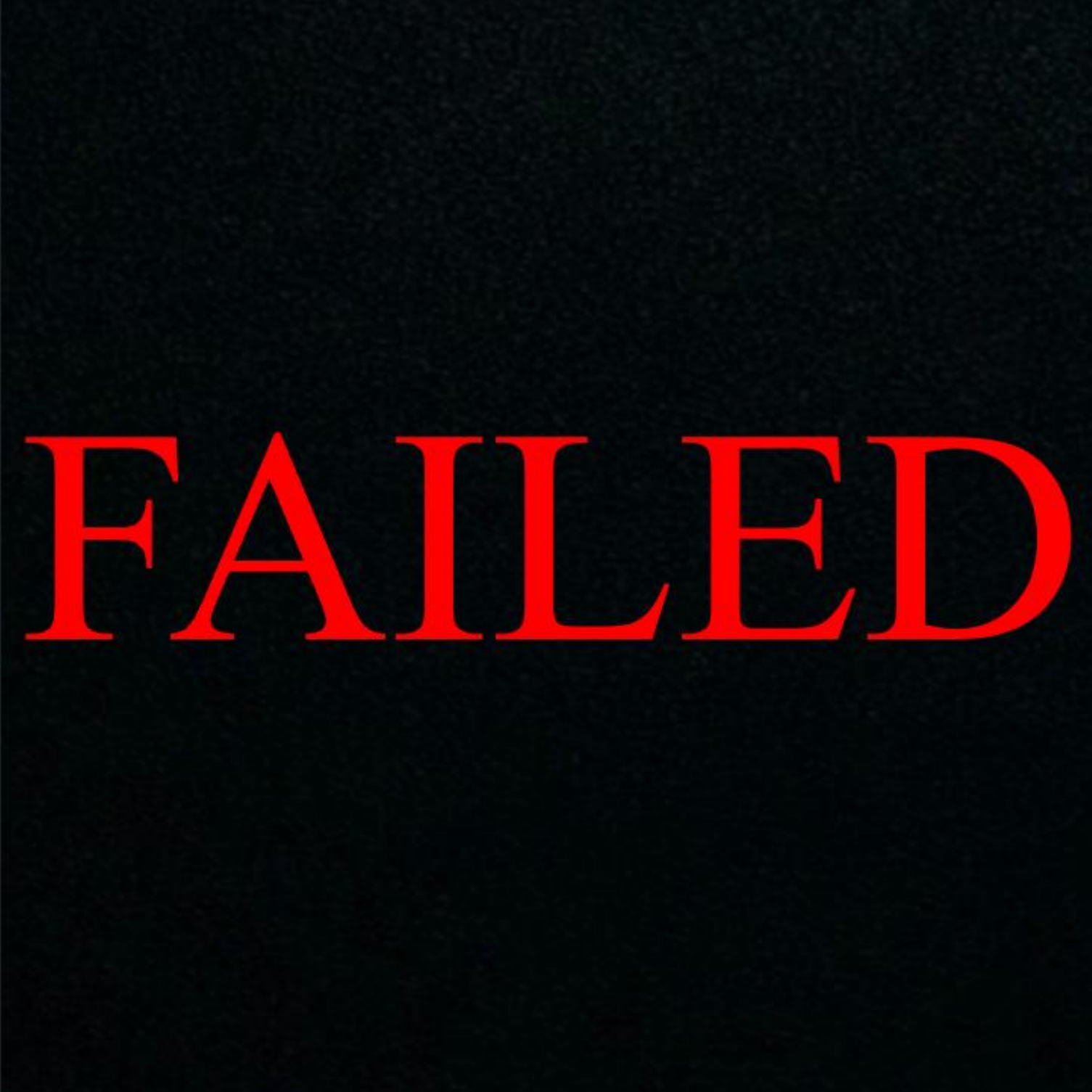} &
        \includegraphics[width=\imgwidth]{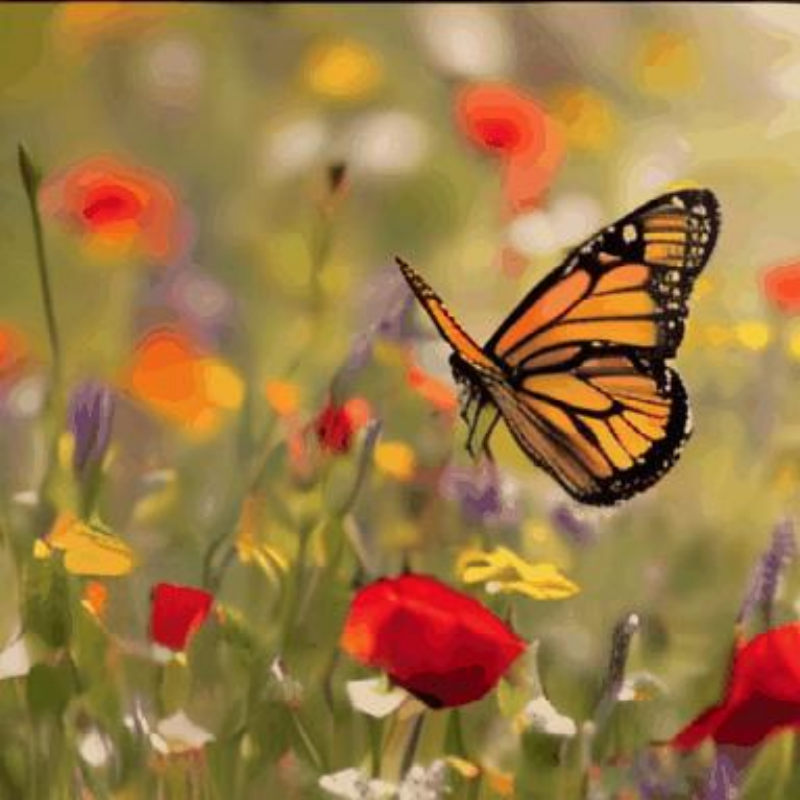} &
        \includegraphics[width=\imgwidth]{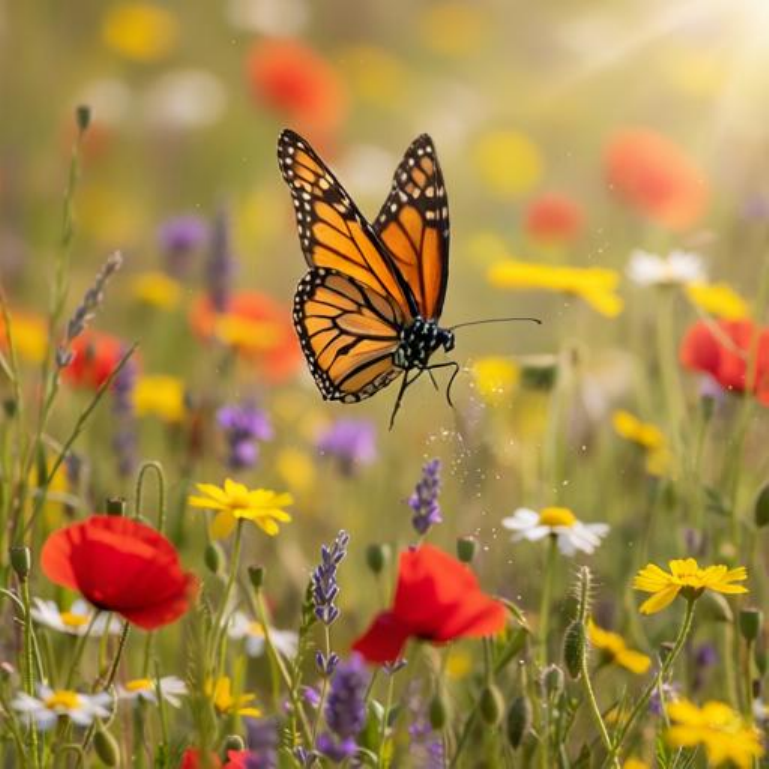} \\
        
        \raisebox{14pt}{\rotatebox{90}{\small Cartoon Dog}} &
        \includegraphics[width=\imgwidth]{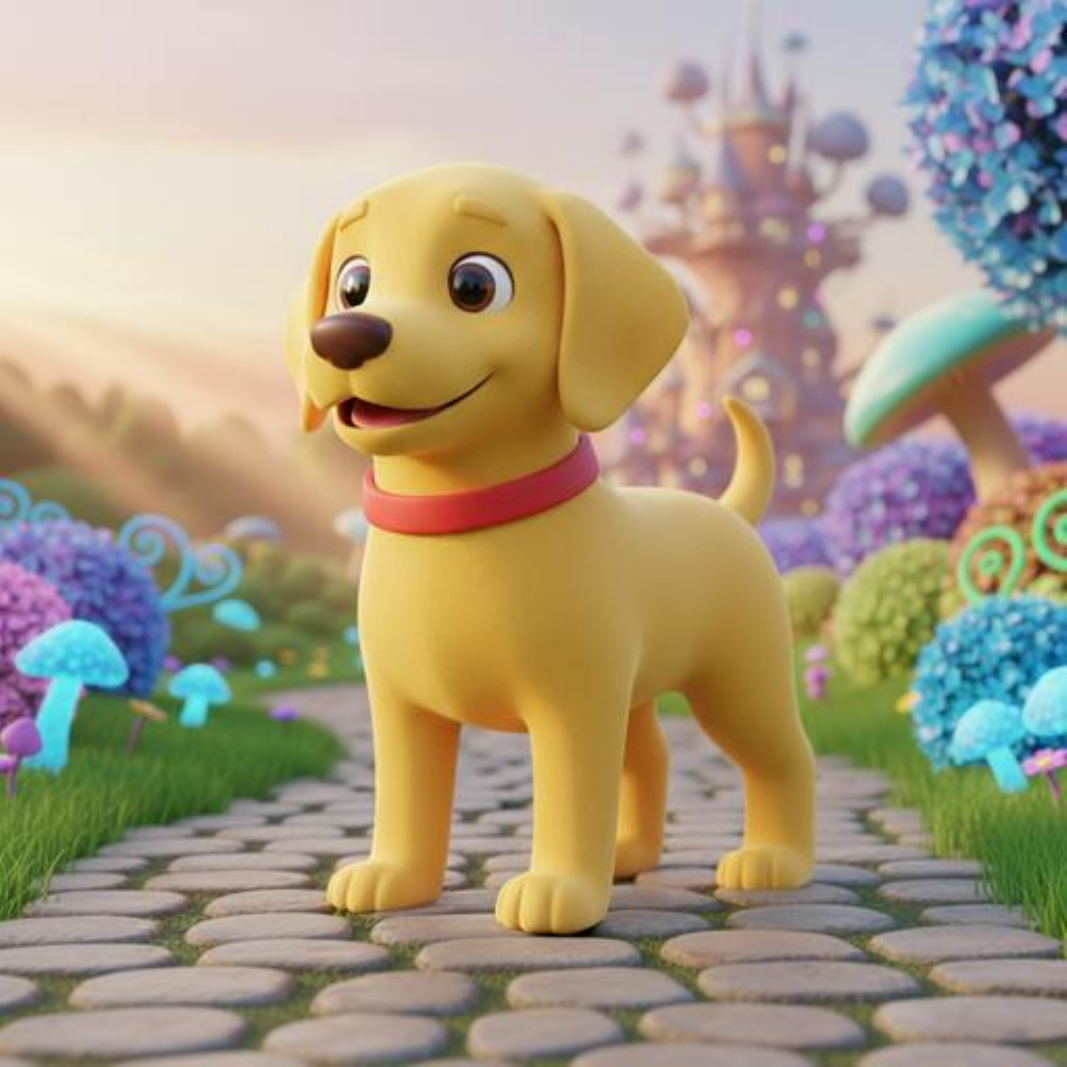} &
        \includegraphics[width=\imgwidth]{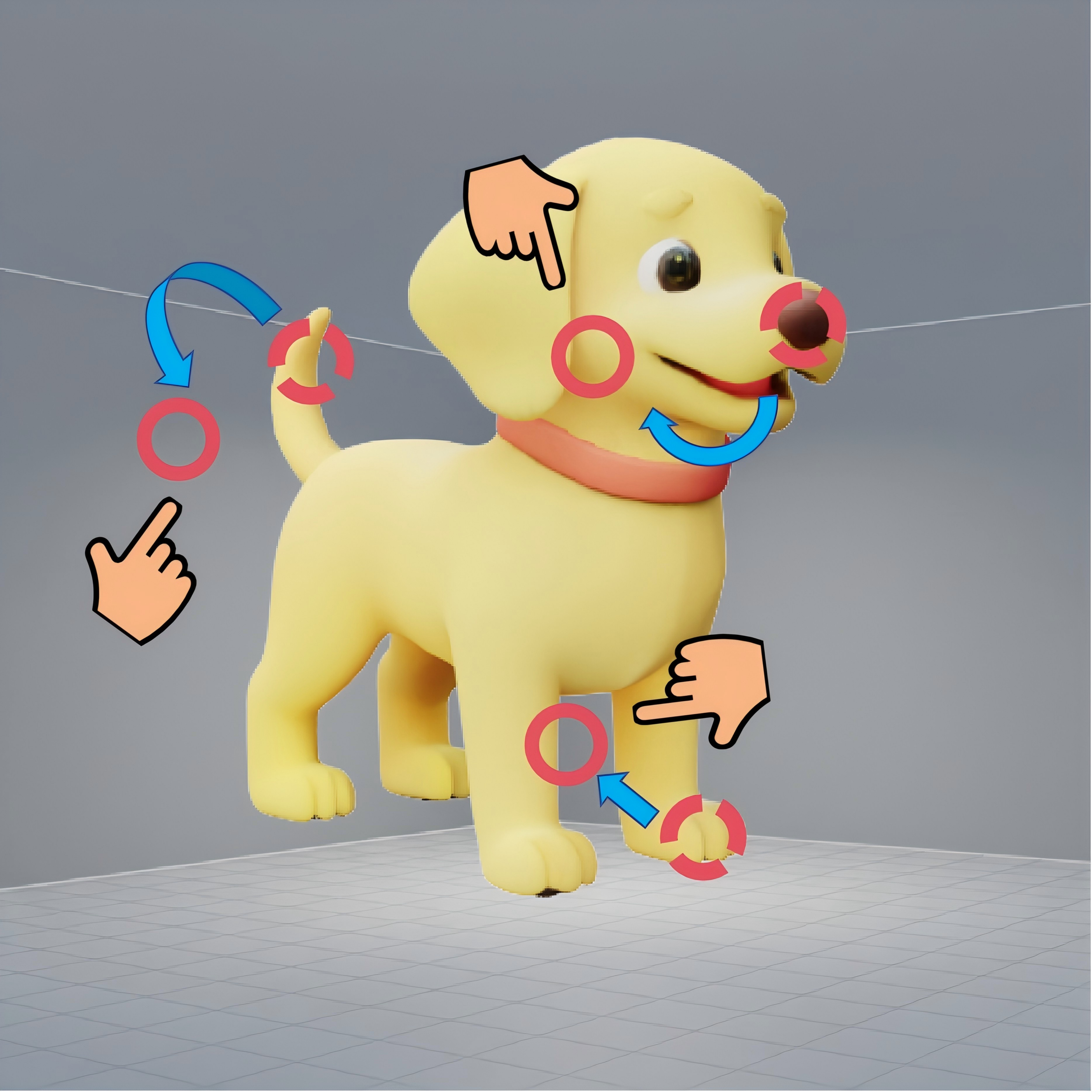} &
        \includegraphics[width=\imgwidth]{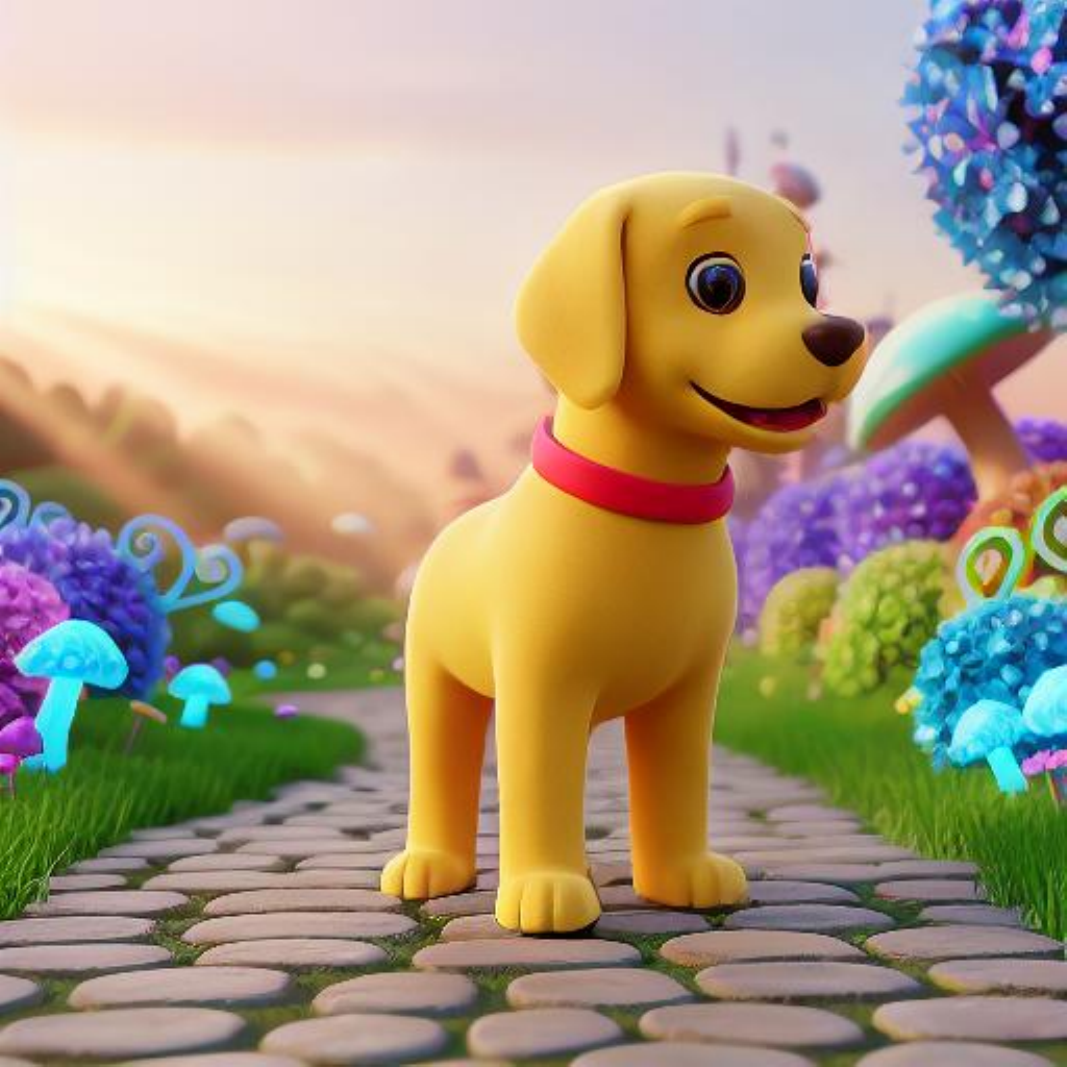} &
        \includegraphics[width=\imgwidth]{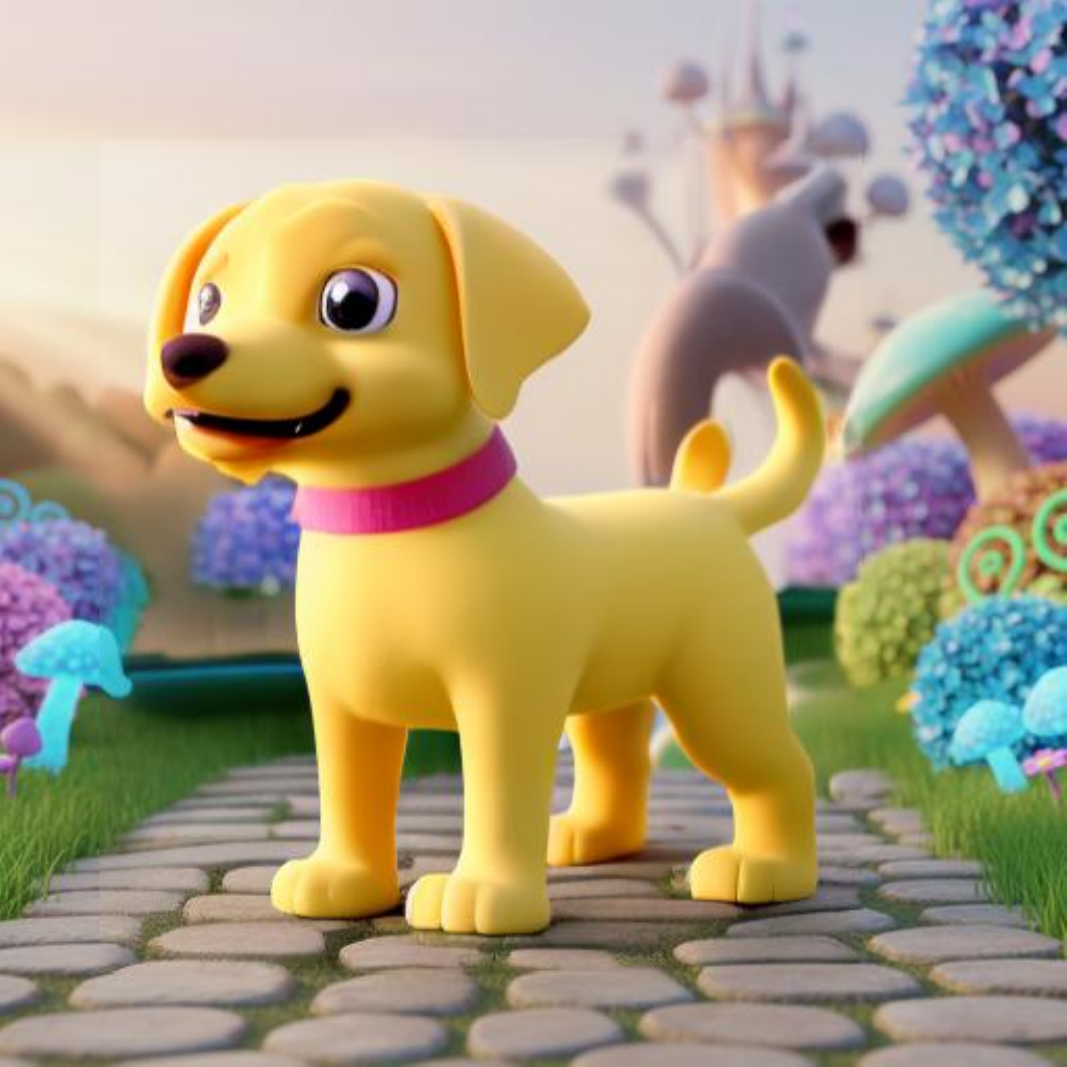} &
        \includegraphics[width=\imgwidth]{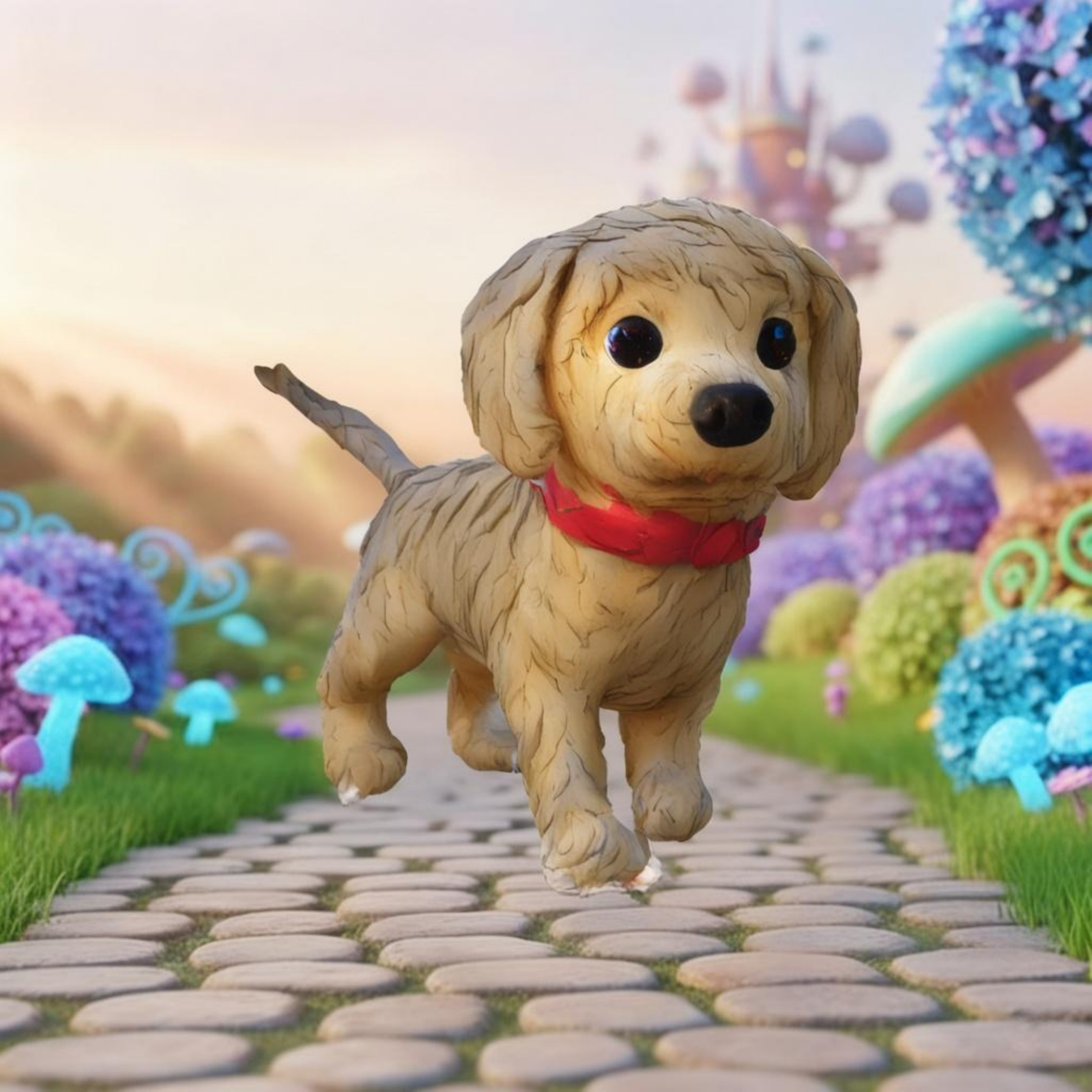} &
        \includegraphics[width=\imgwidth]{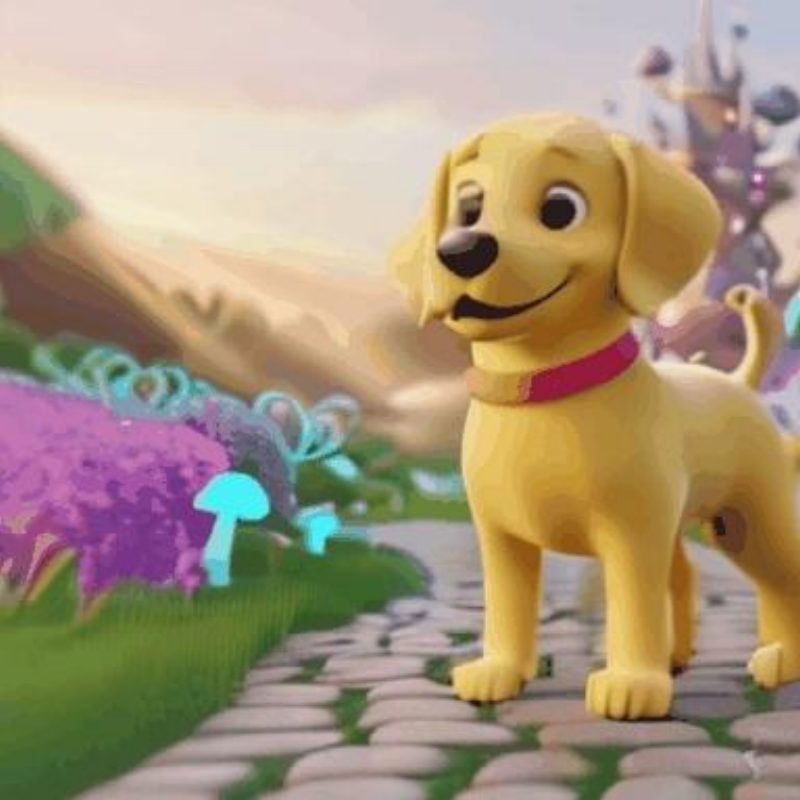} &
        \includegraphics[width=\imgwidth]{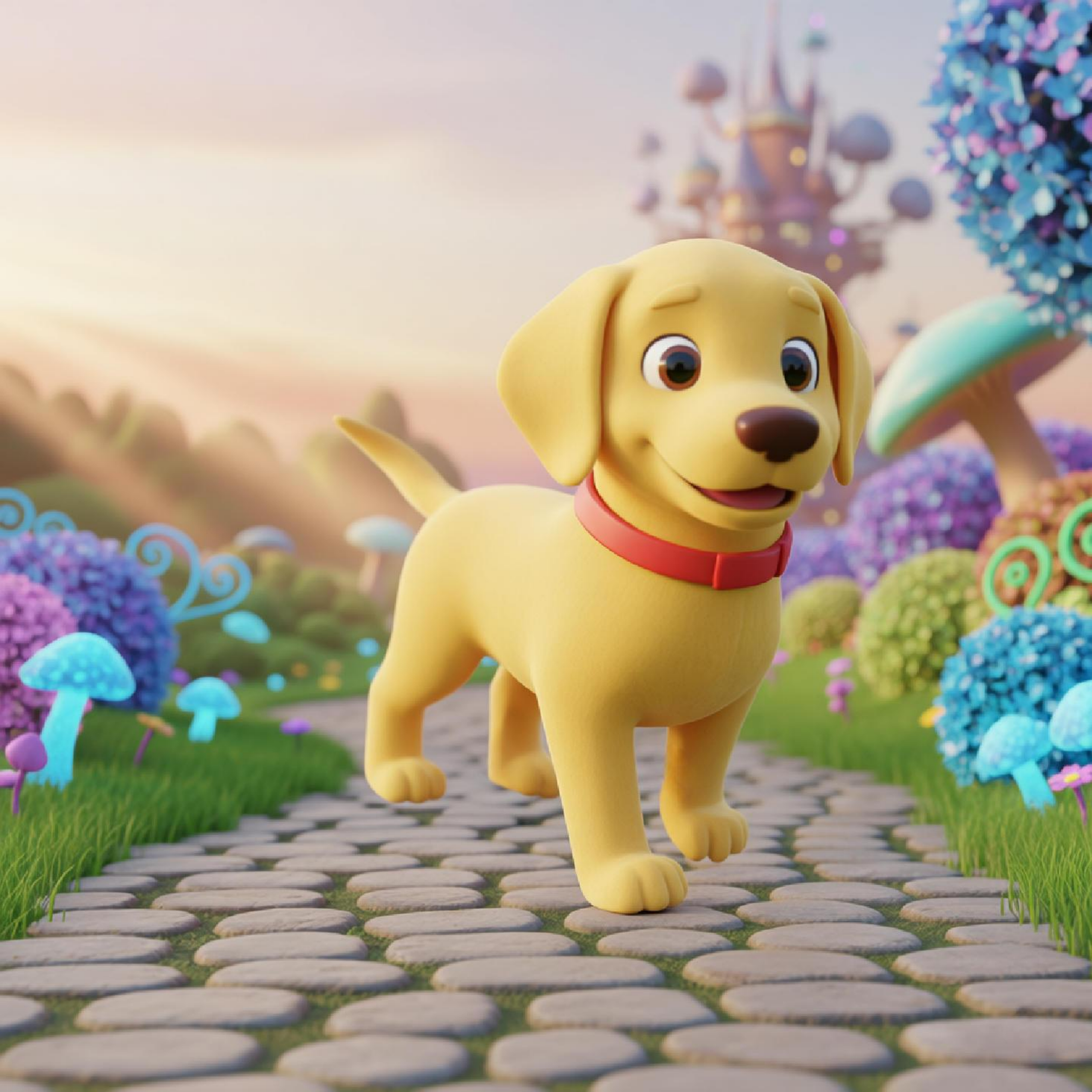} \\
        
        \raisebox{14pt}{\rotatebox{90}{\small Knight}} &
        \includegraphics[width=\imgwidth]{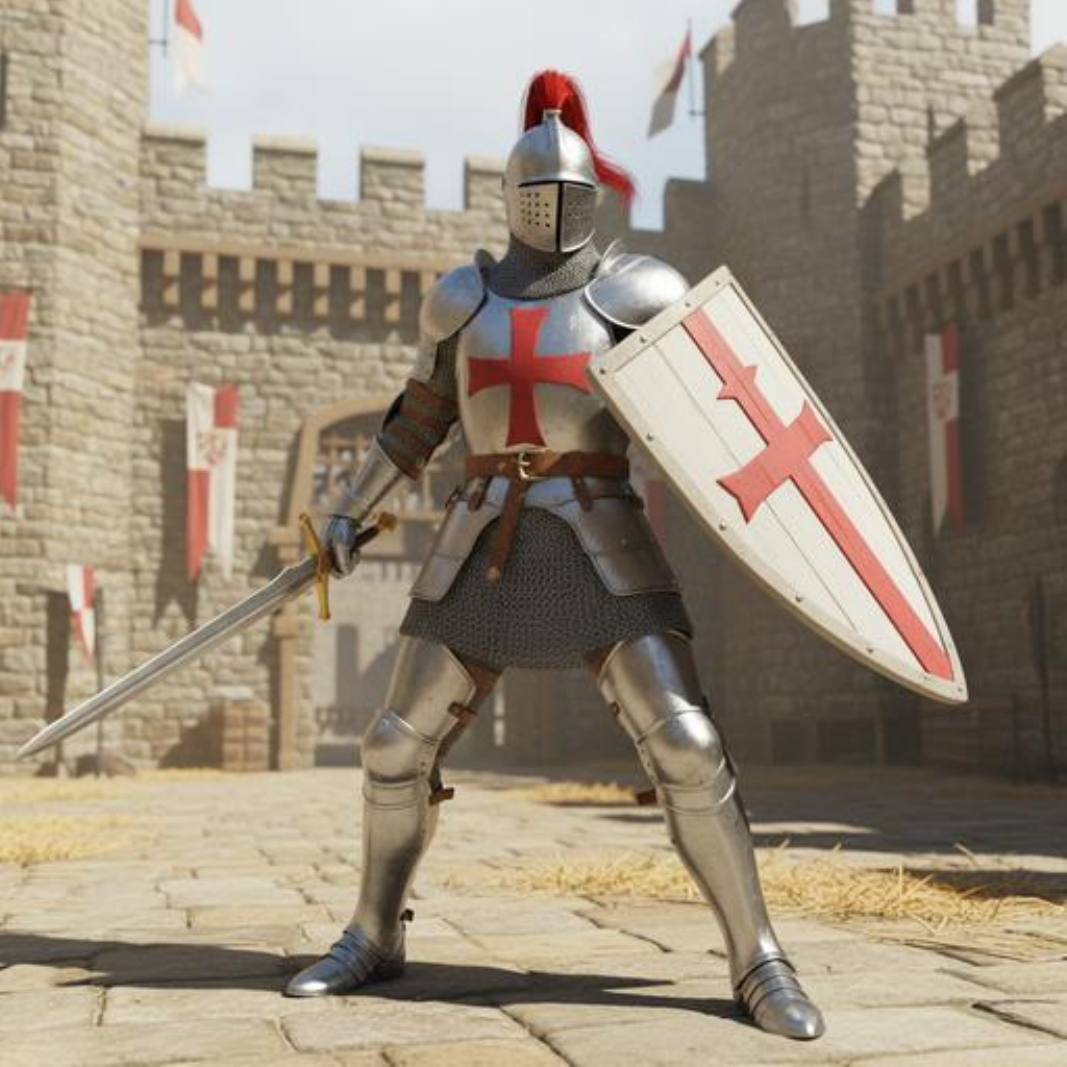} &
        \includegraphics[width=\imgwidth]{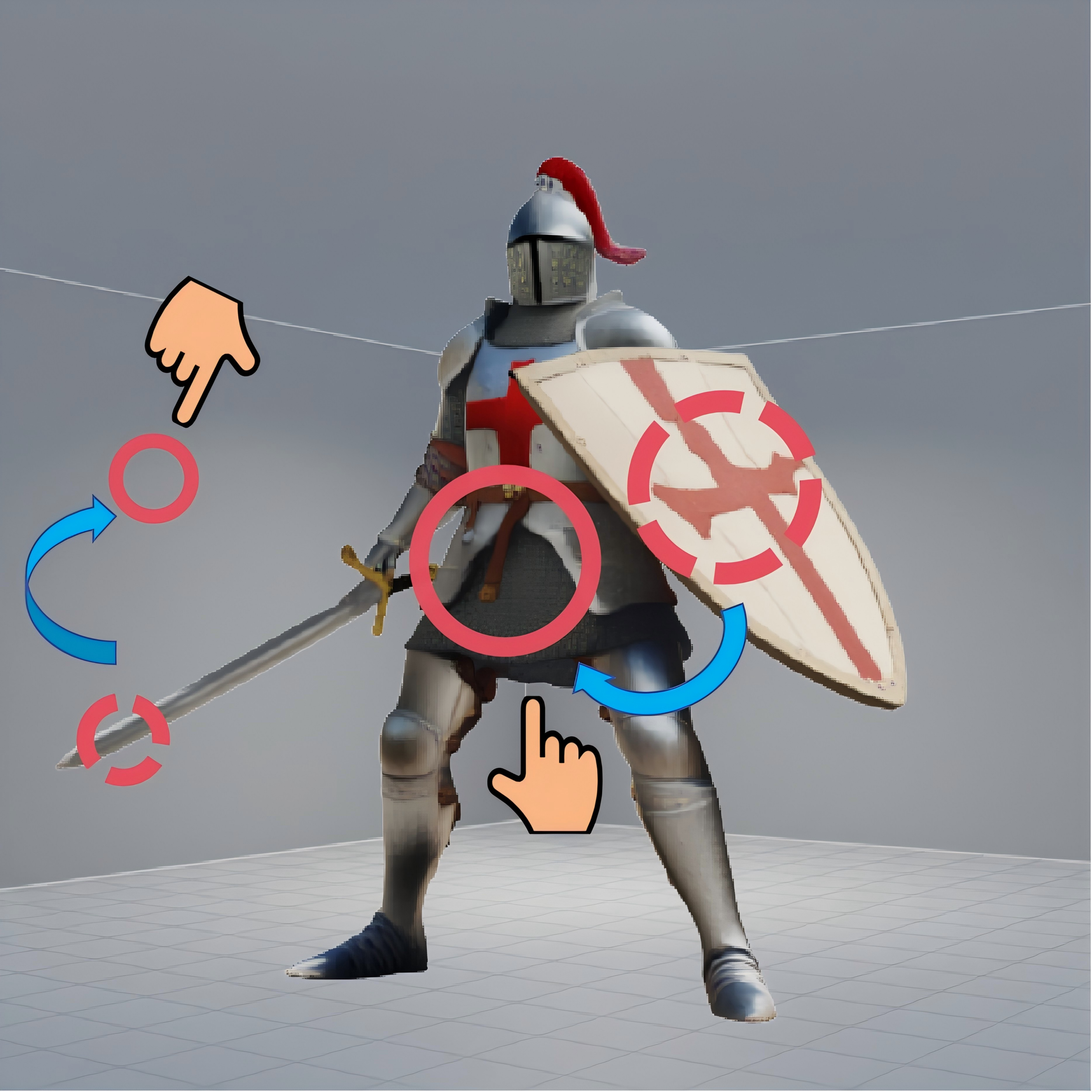} &
        \includegraphics[width=\imgwidth]{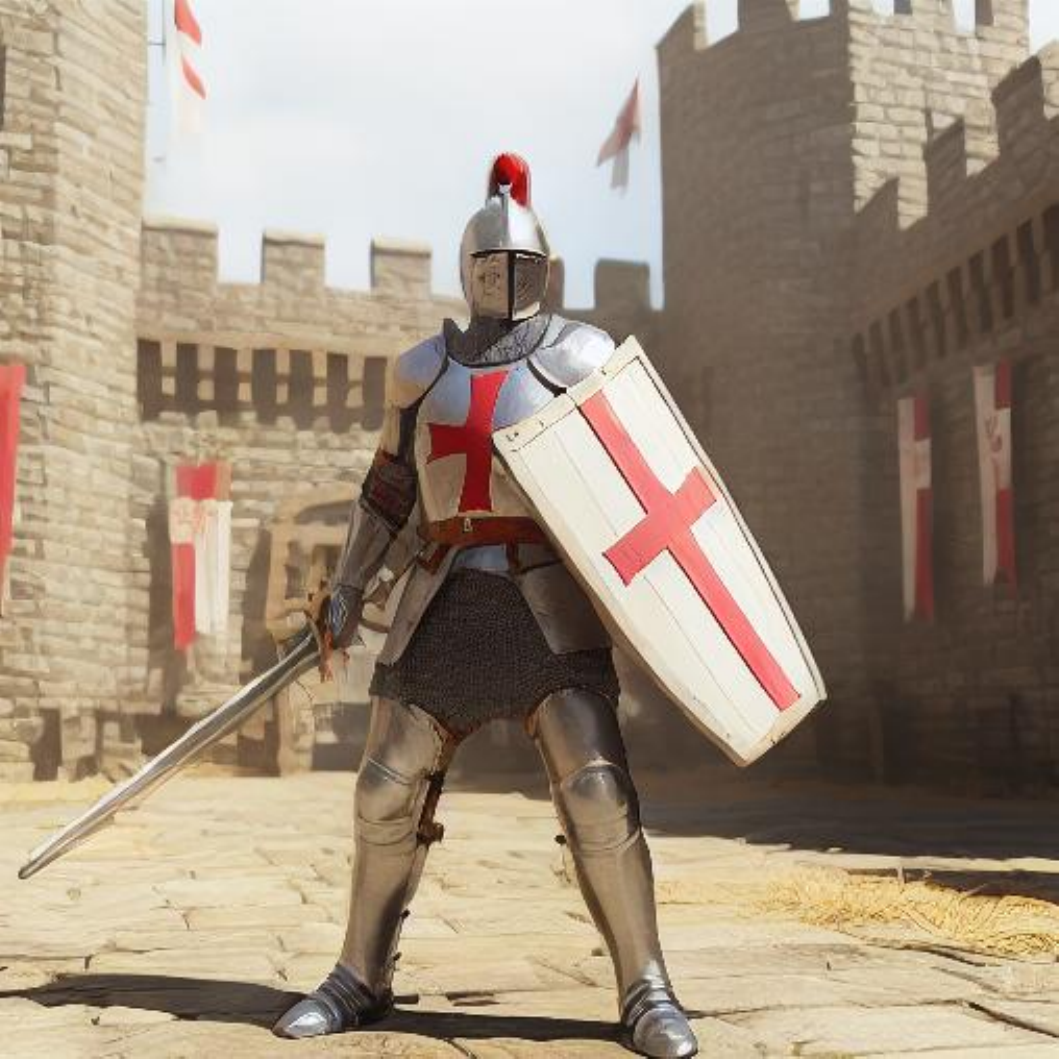} &
        \includegraphics[width=\imgwidth]{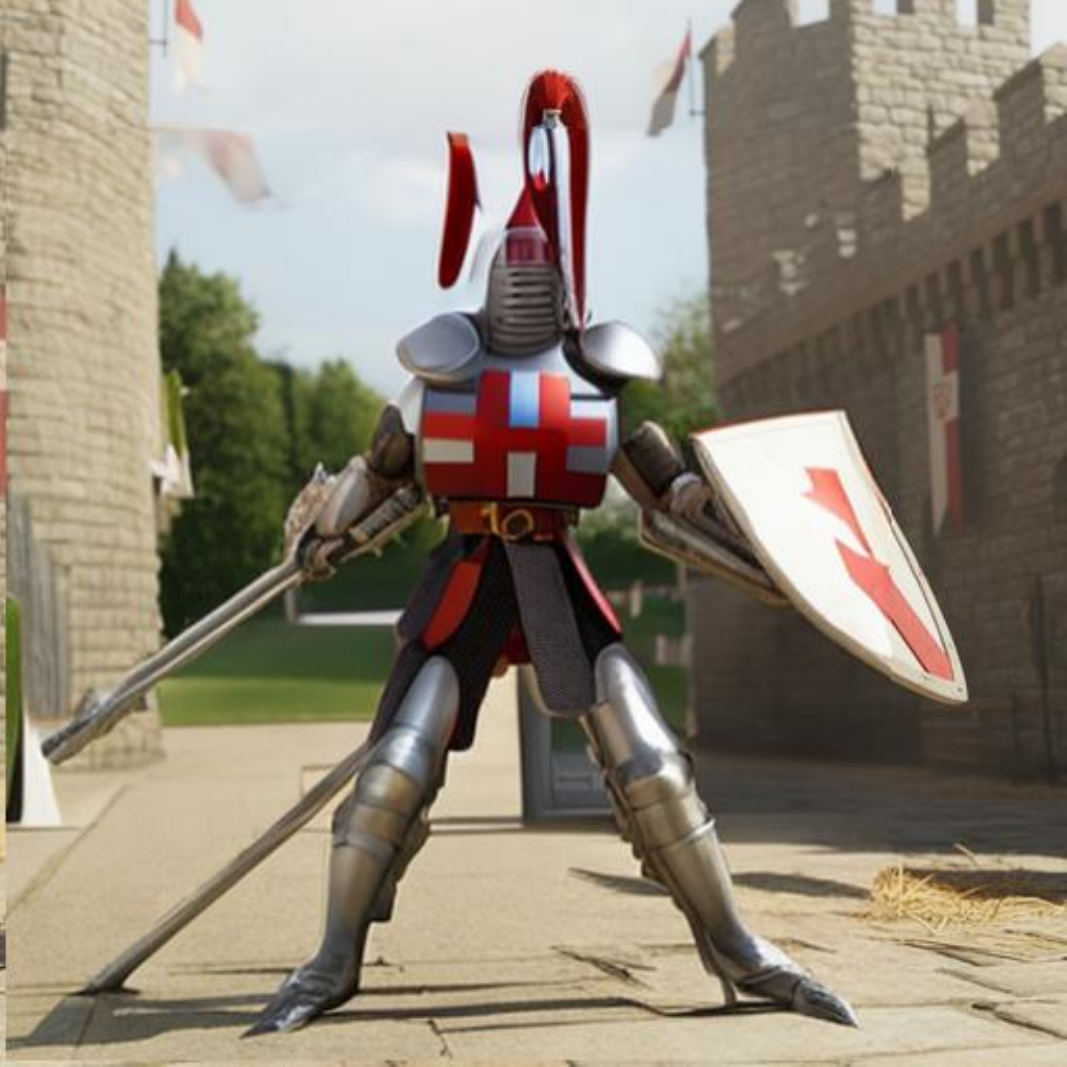} &
        \includegraphics[width=\imgwidth]{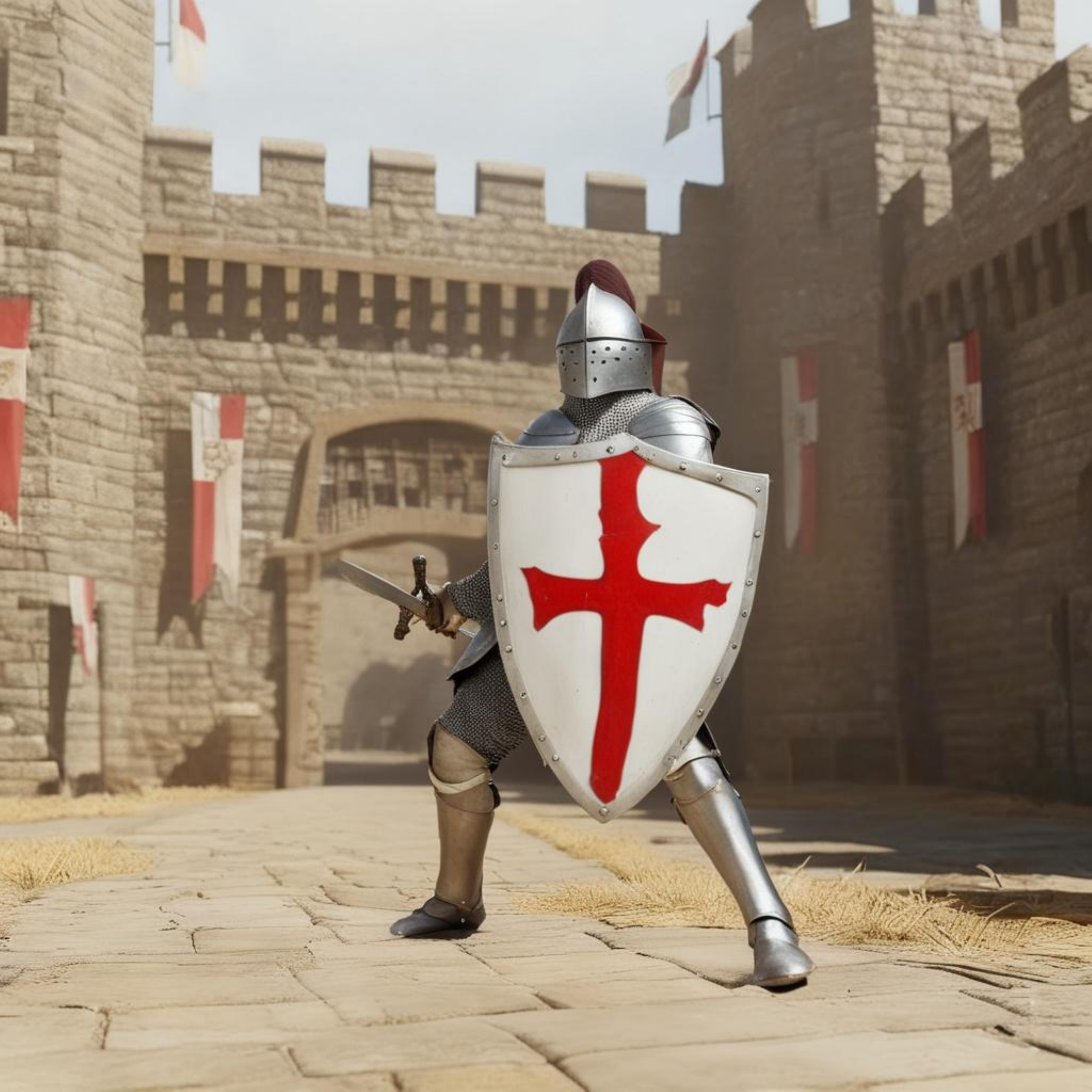} &
        \includegraphics[width=\imgwidth]{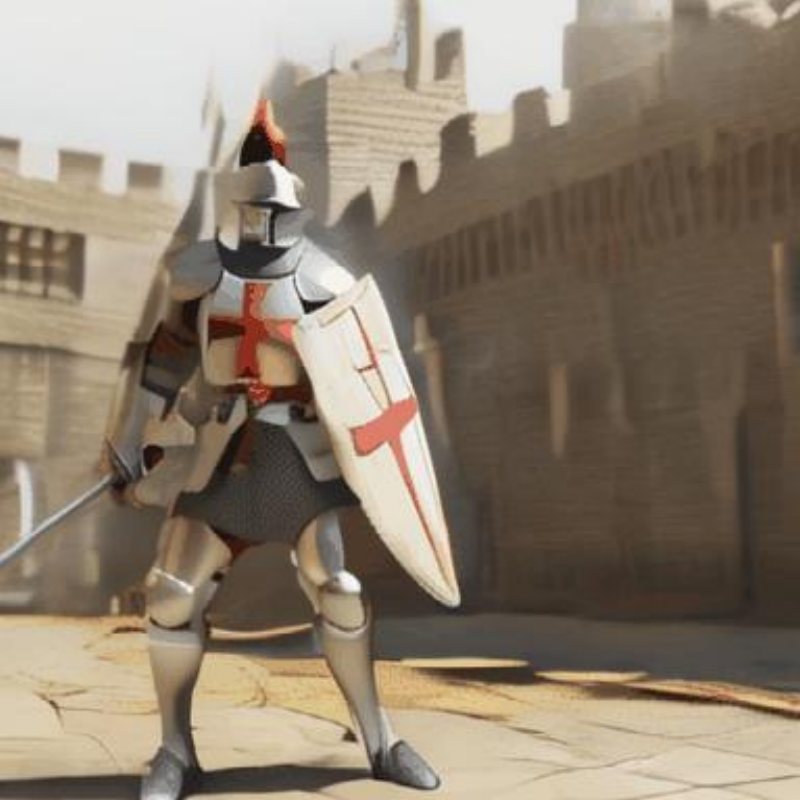} &
        \includegraphics[width=\imgwidth]{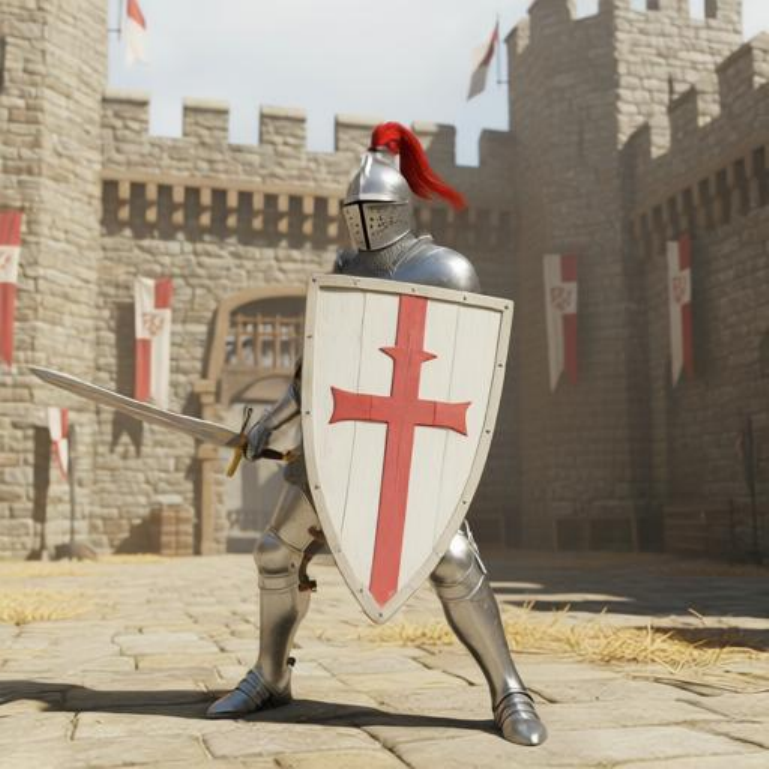} \\
        
        \raisebox{14pt}{\rotatebox{90}{\small Chair}} &
        \includegraphics[width=\imgwidth]{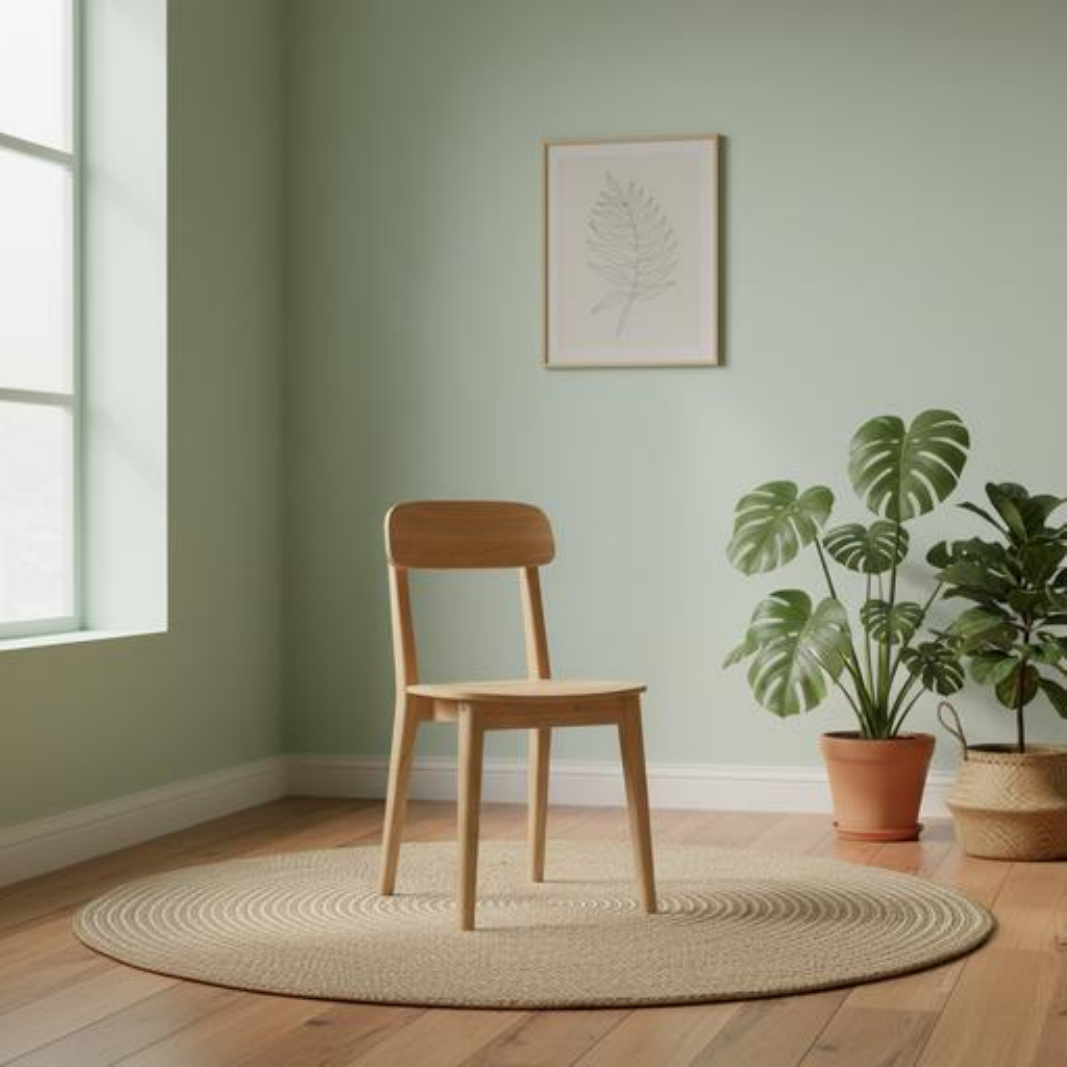} &
        \includegraphics[width=\imgwidth]{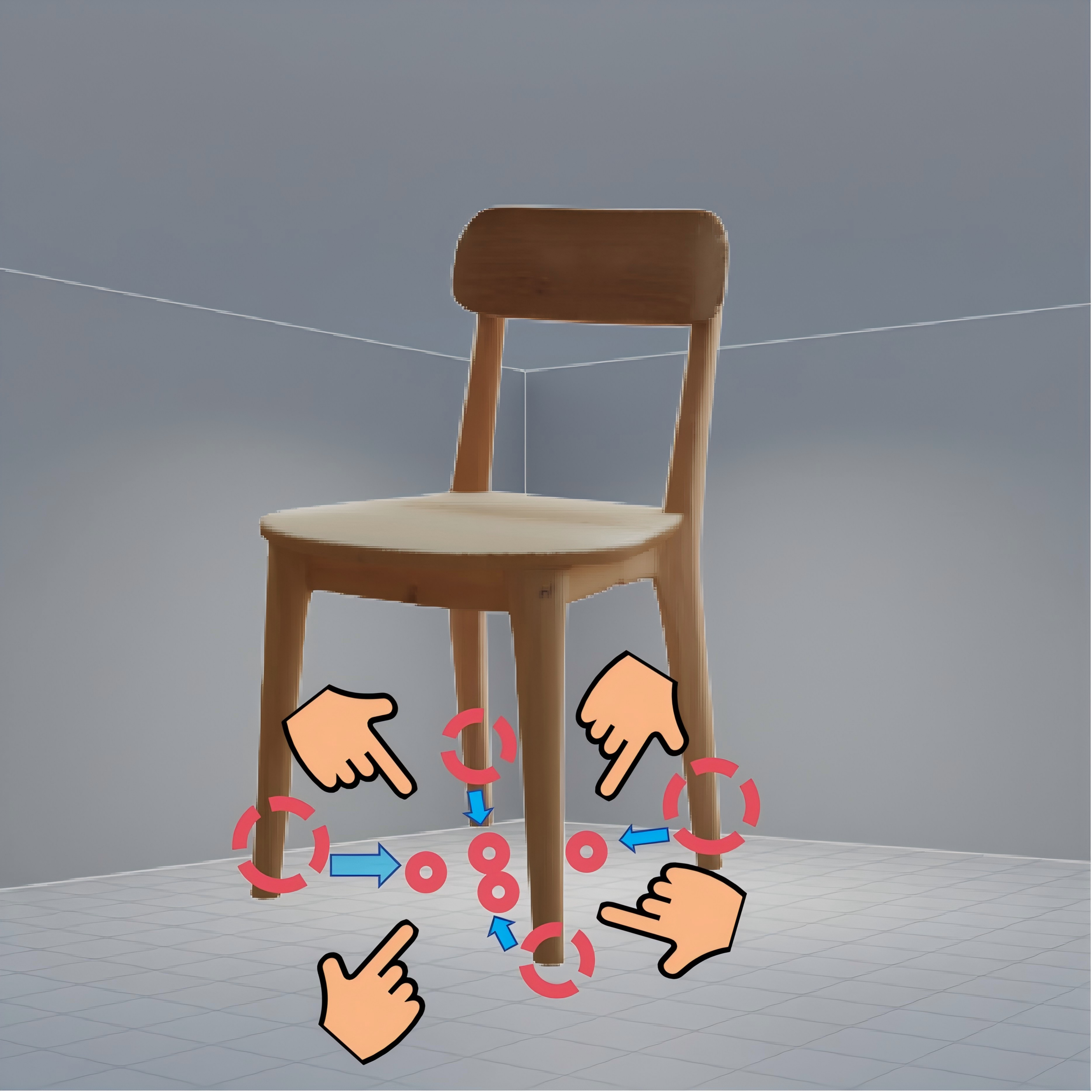} &
        \includegraphics[width=\imgwidth]{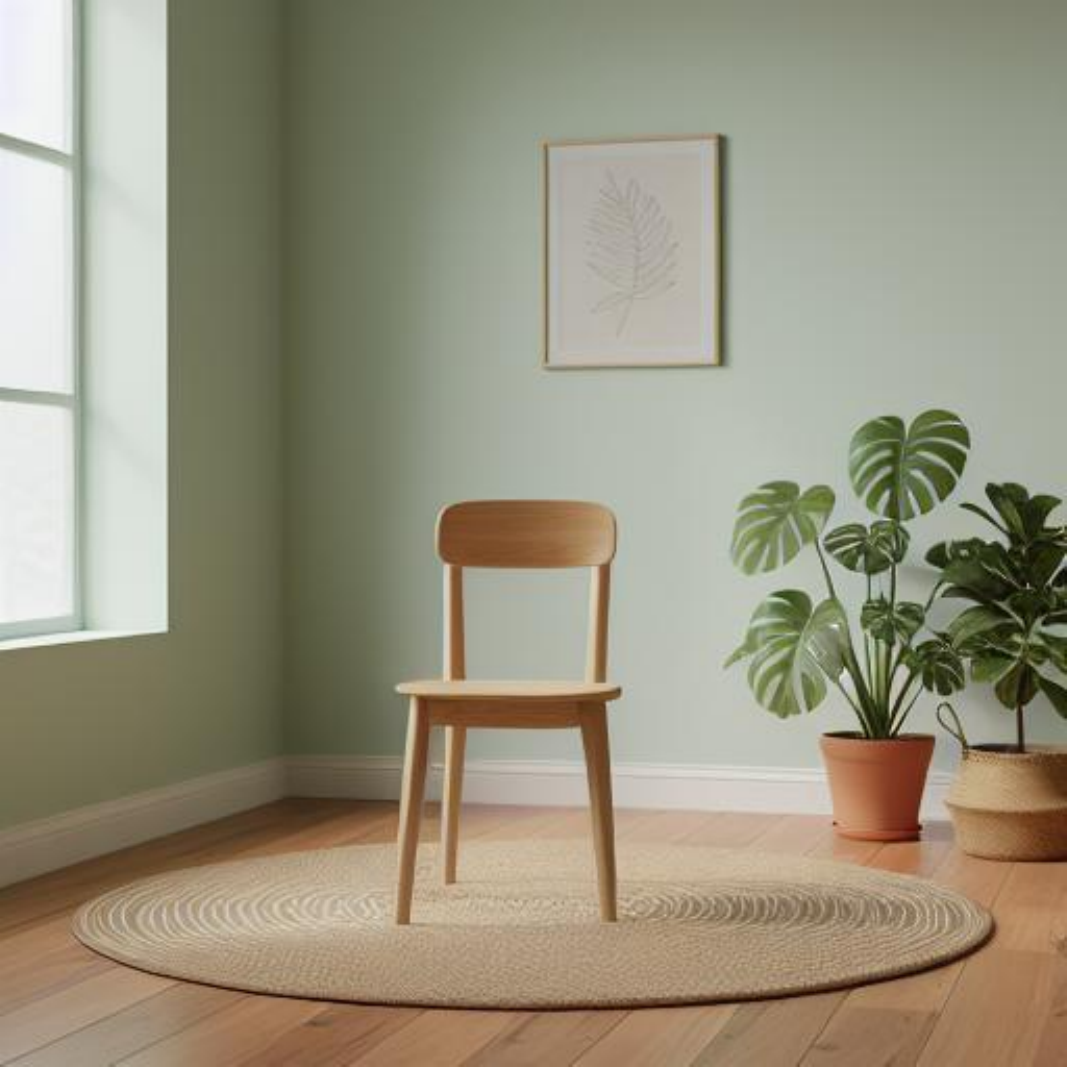} &
        \includegraphics[width=\imgwidth]{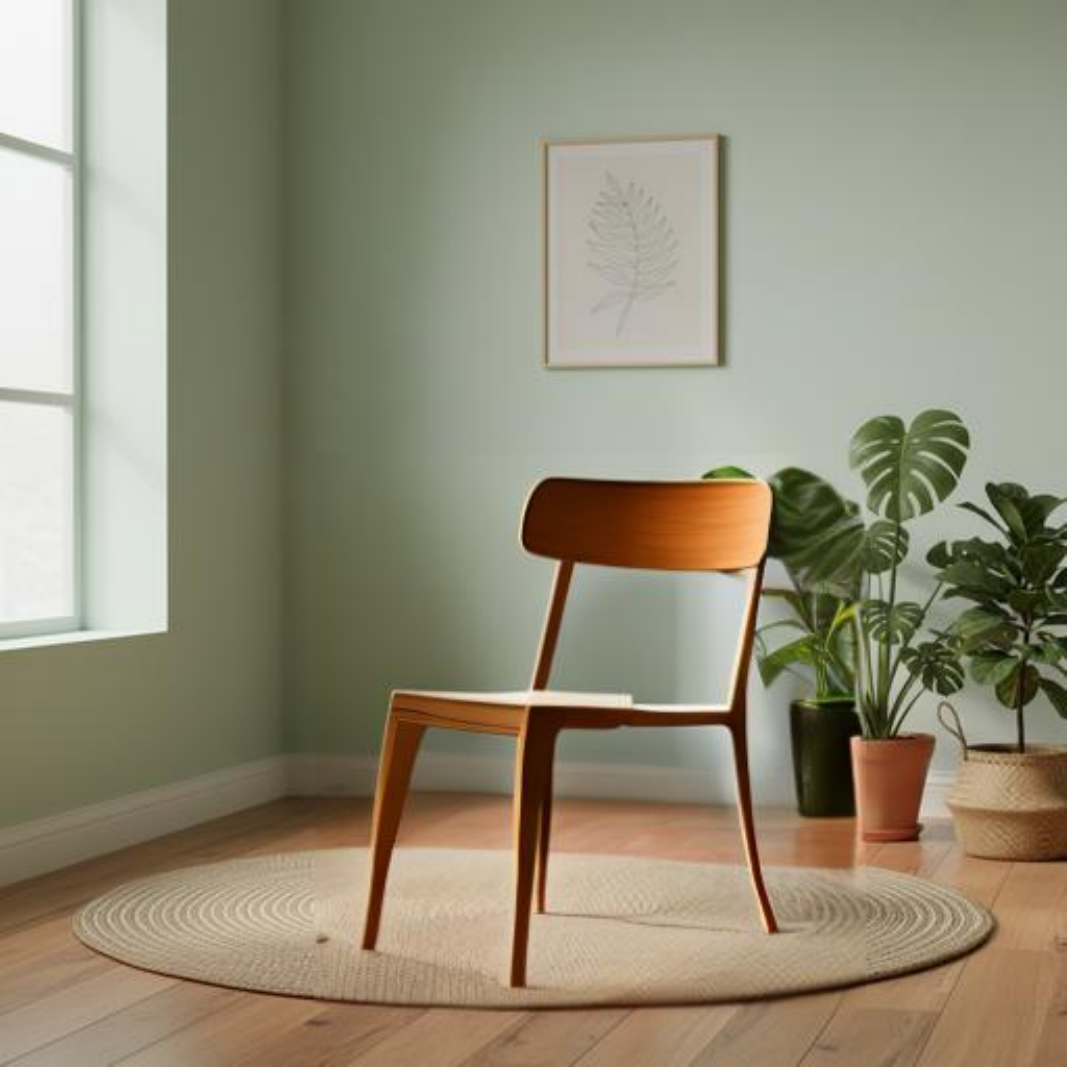} &
        \includegraphics[width=\imgwidth]{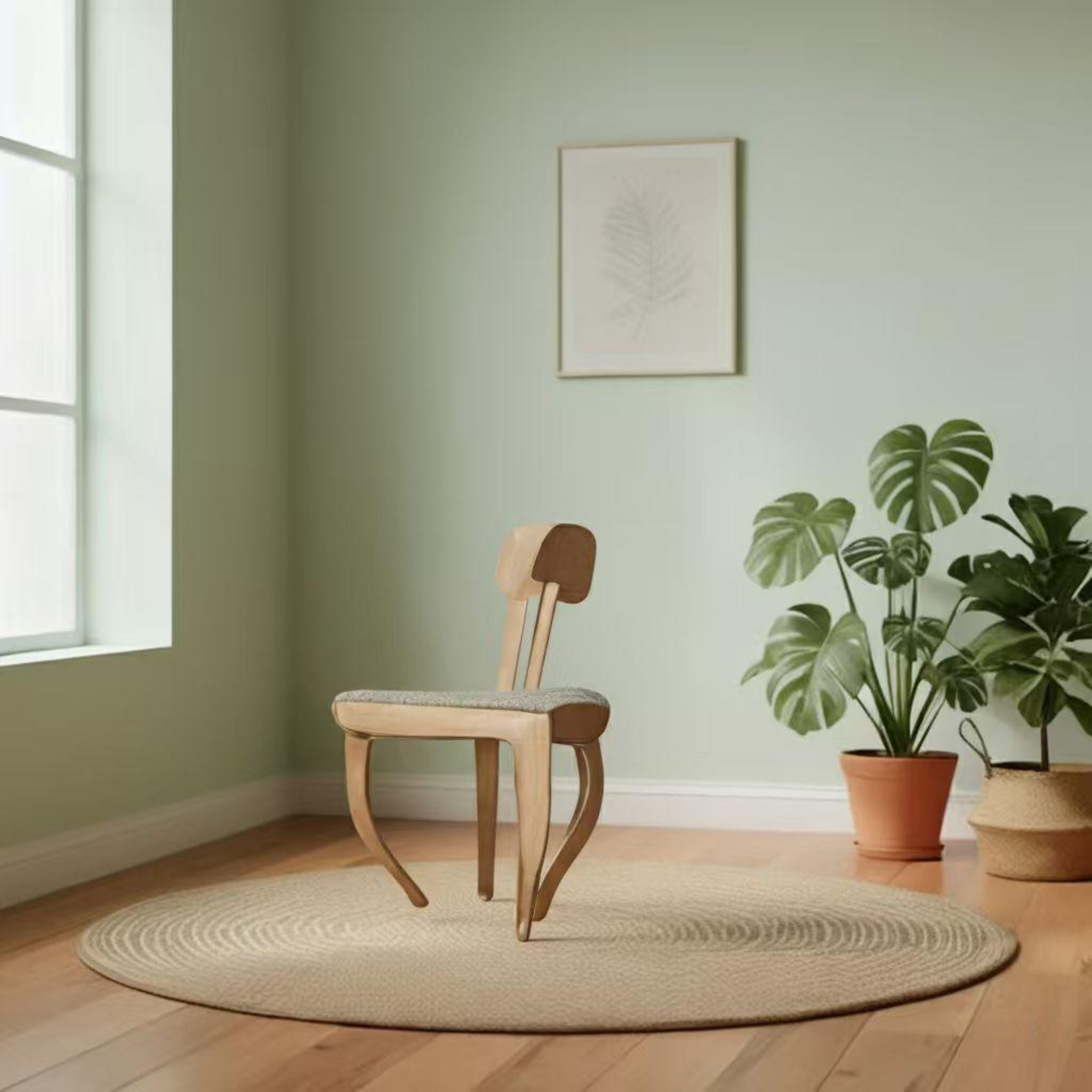} &
        \includegraphics[width=\imgwidth]{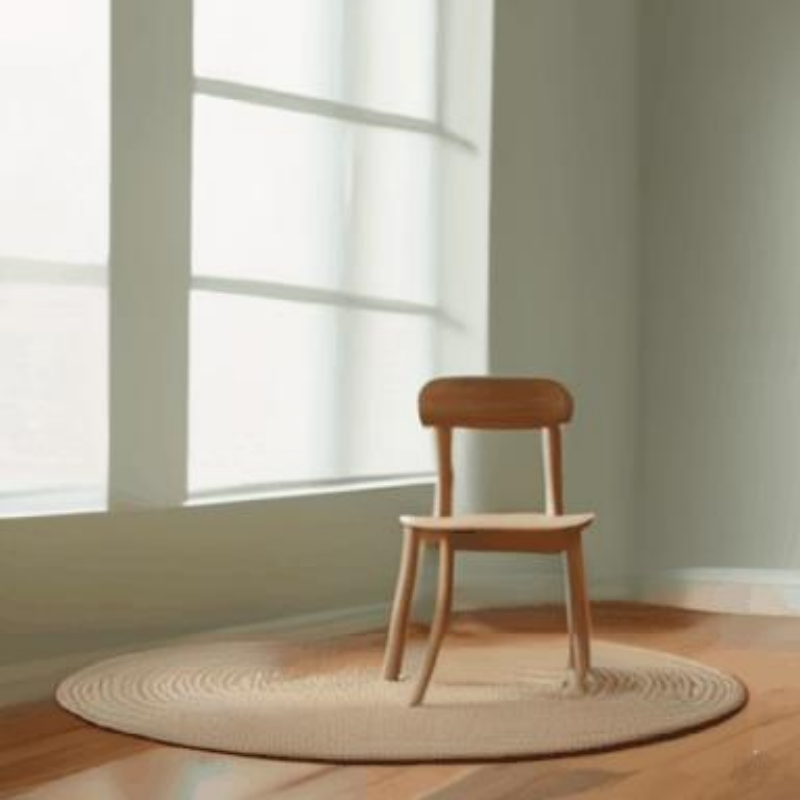} &
        \includegraphics[width=\imgwidth]{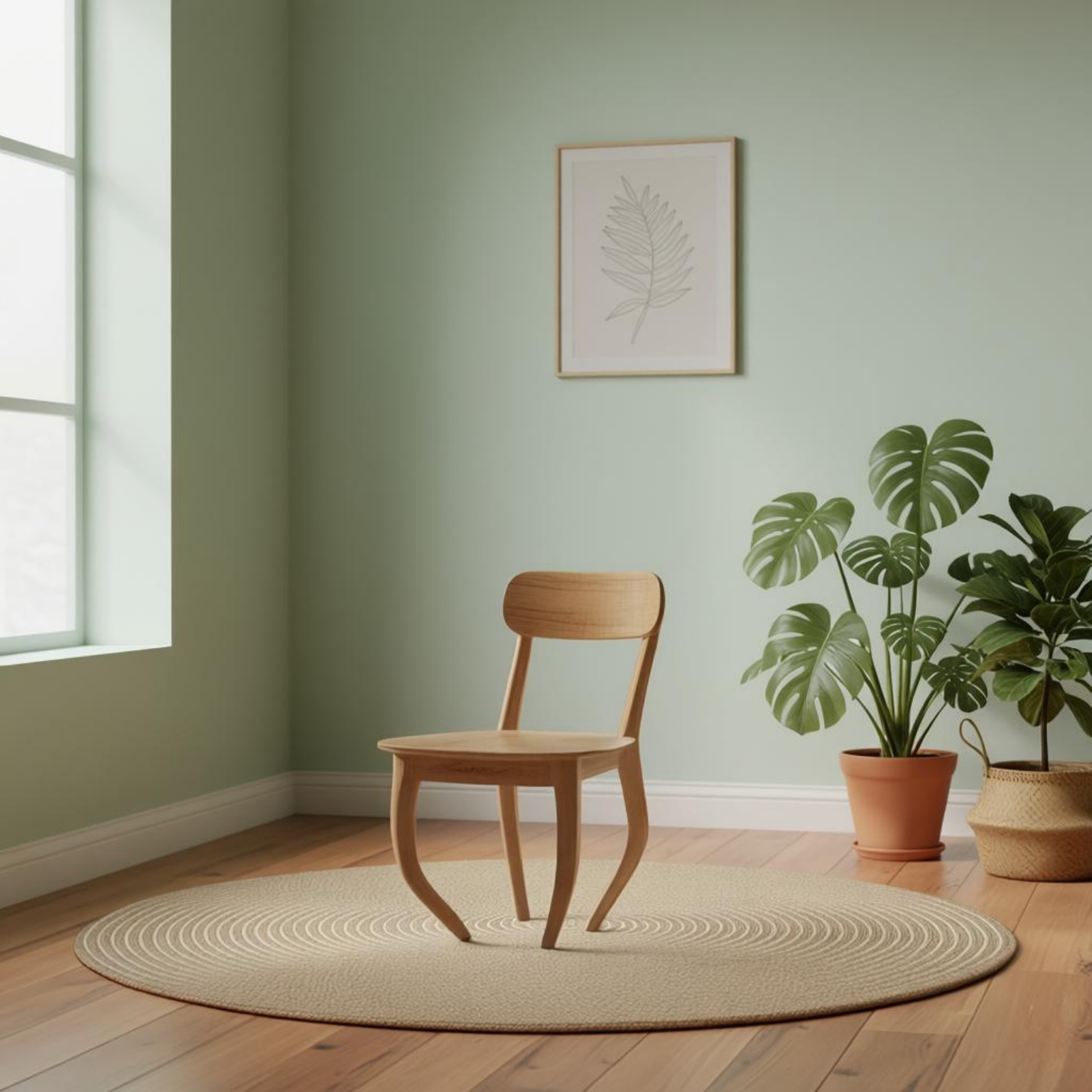} \\
        
        \raisebox{14pt}{\rotatebox{90}{\small Dog}} &
        \includegraphics[width=\imgwidth]{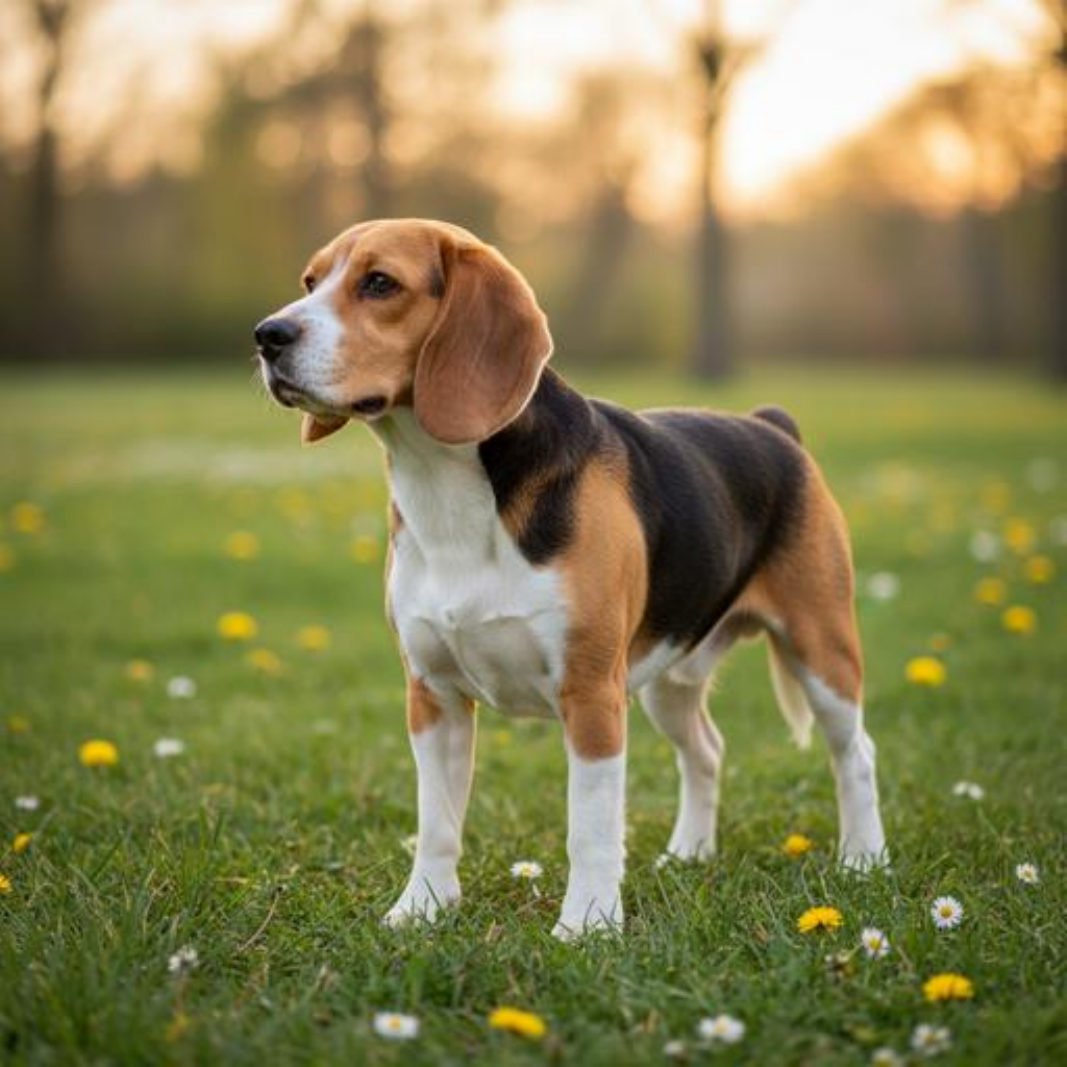} &
        \includegraphics[width=\imgwidth]{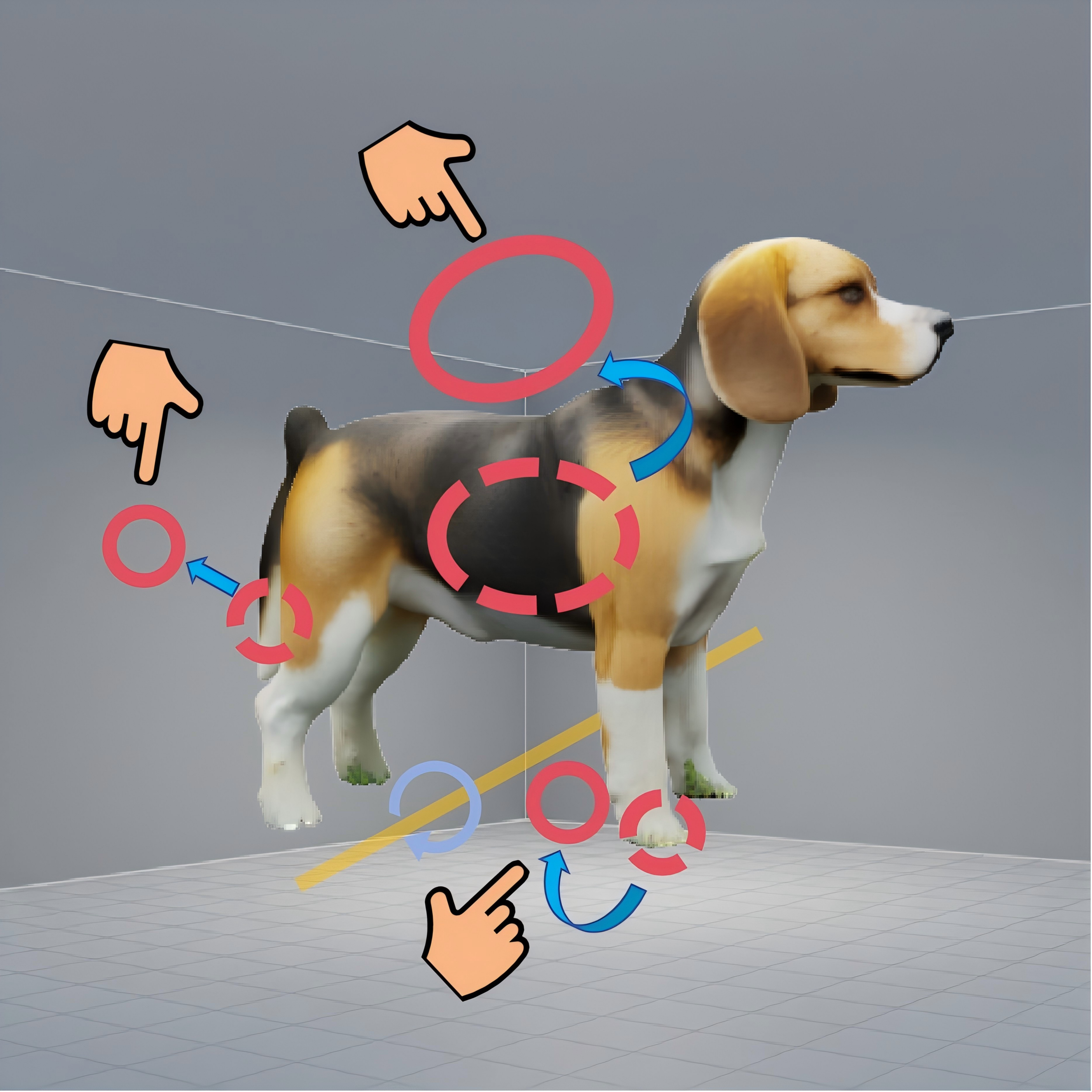} &
        \includegraphics[width=\imgwidth]{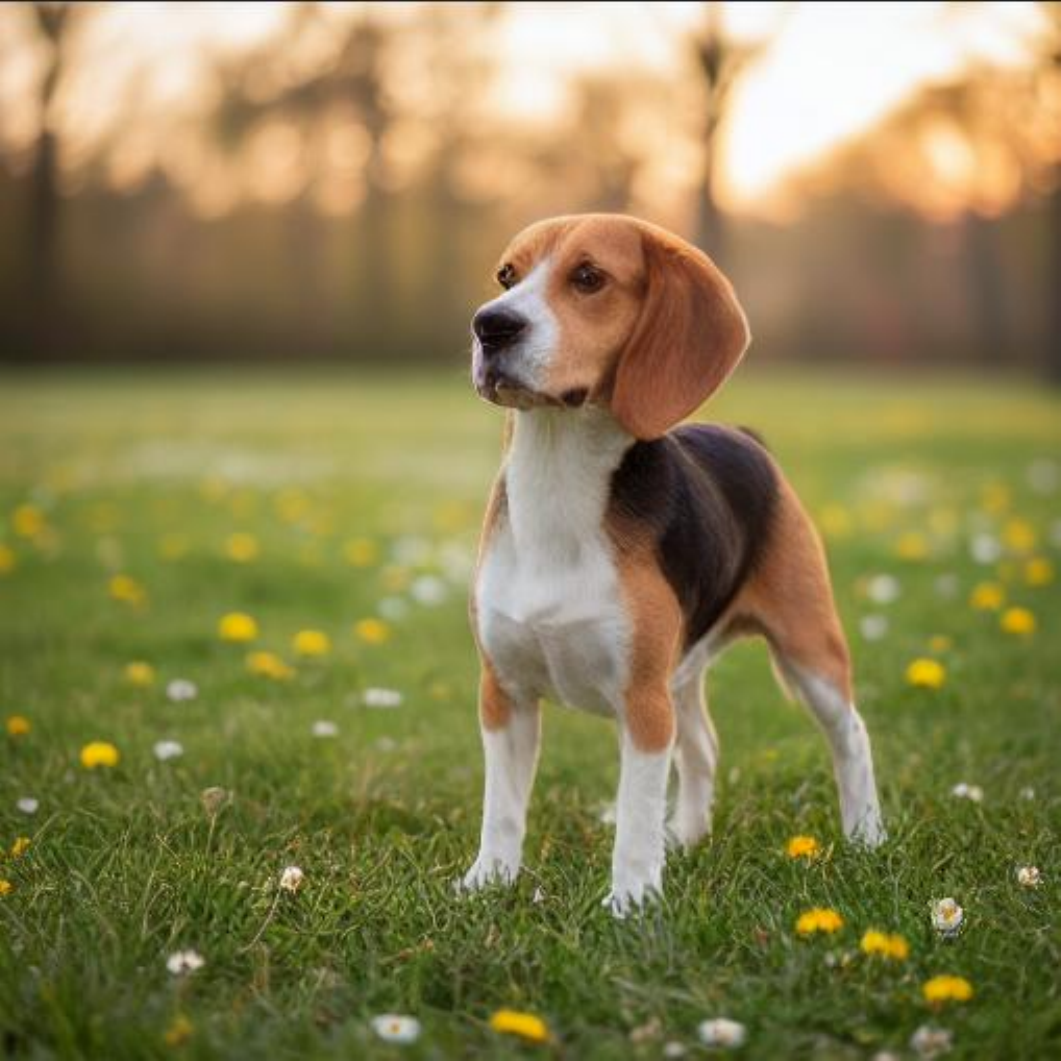} &
        \includegraphics[width=\imgwidth]{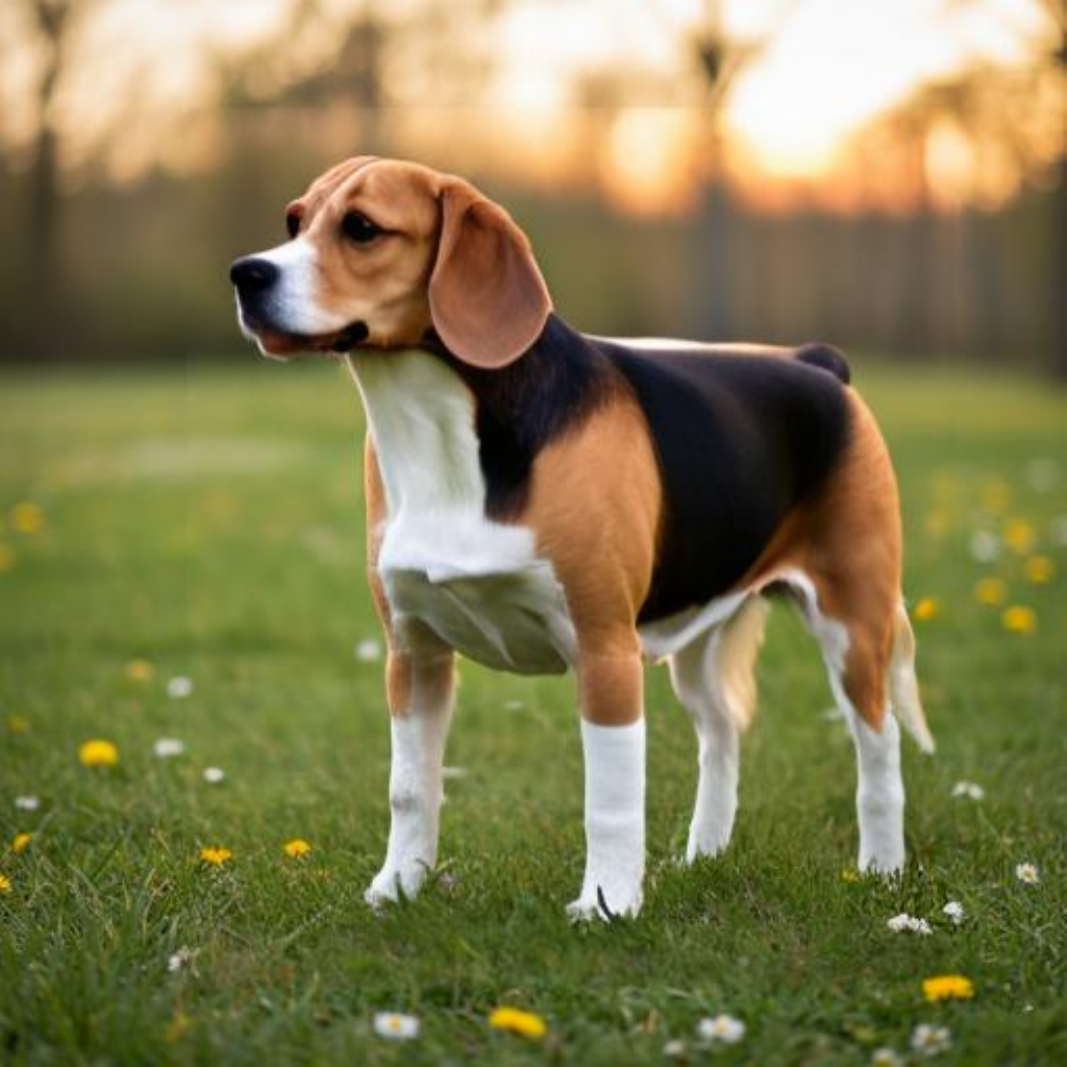} &
        \includegraphics[width=\imgwidth]{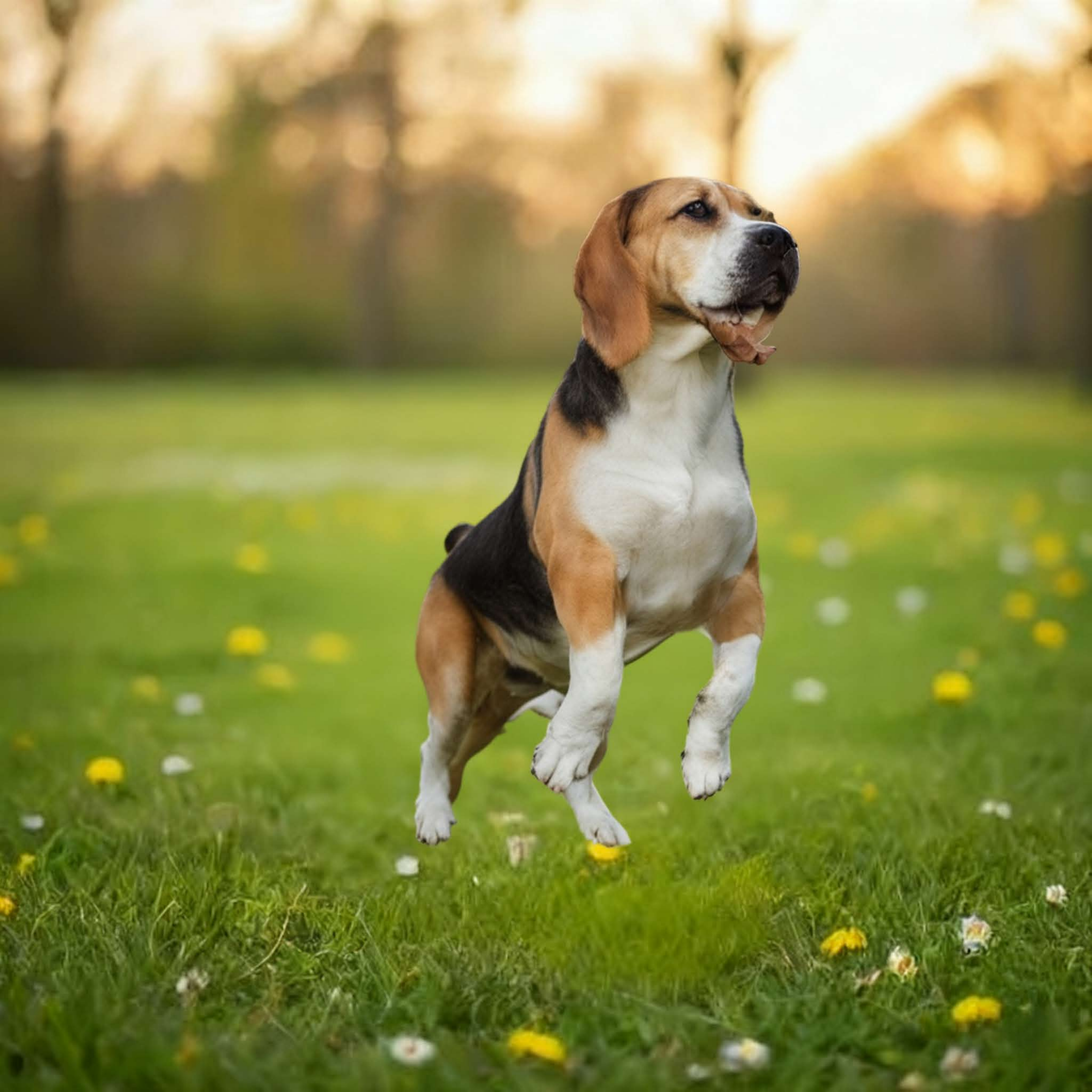} &
        \includegraphics[width=\imgwidth]{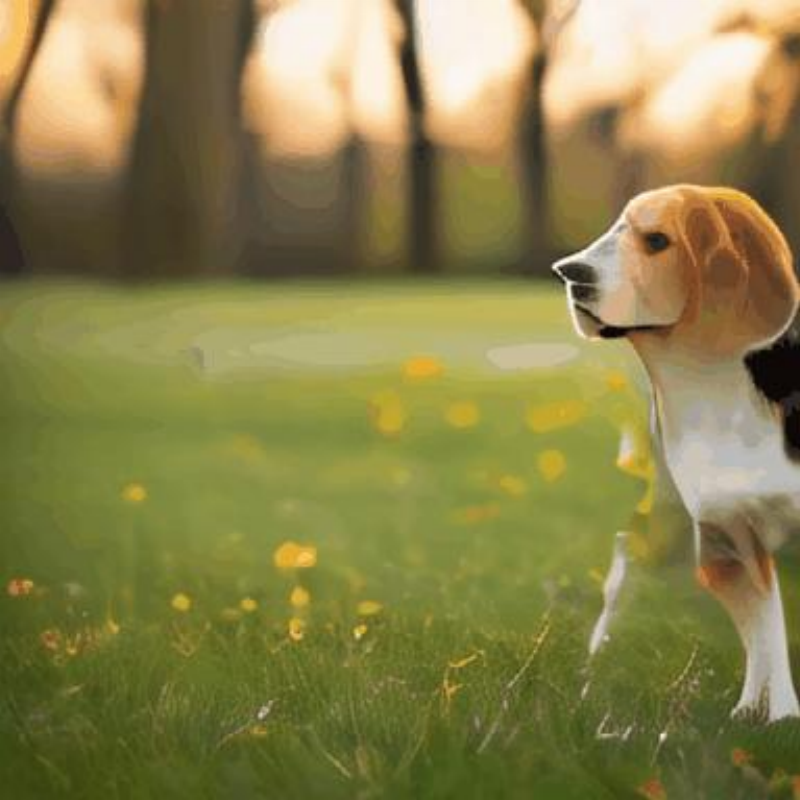} &
        \includegraphics[width=\imgwidth]{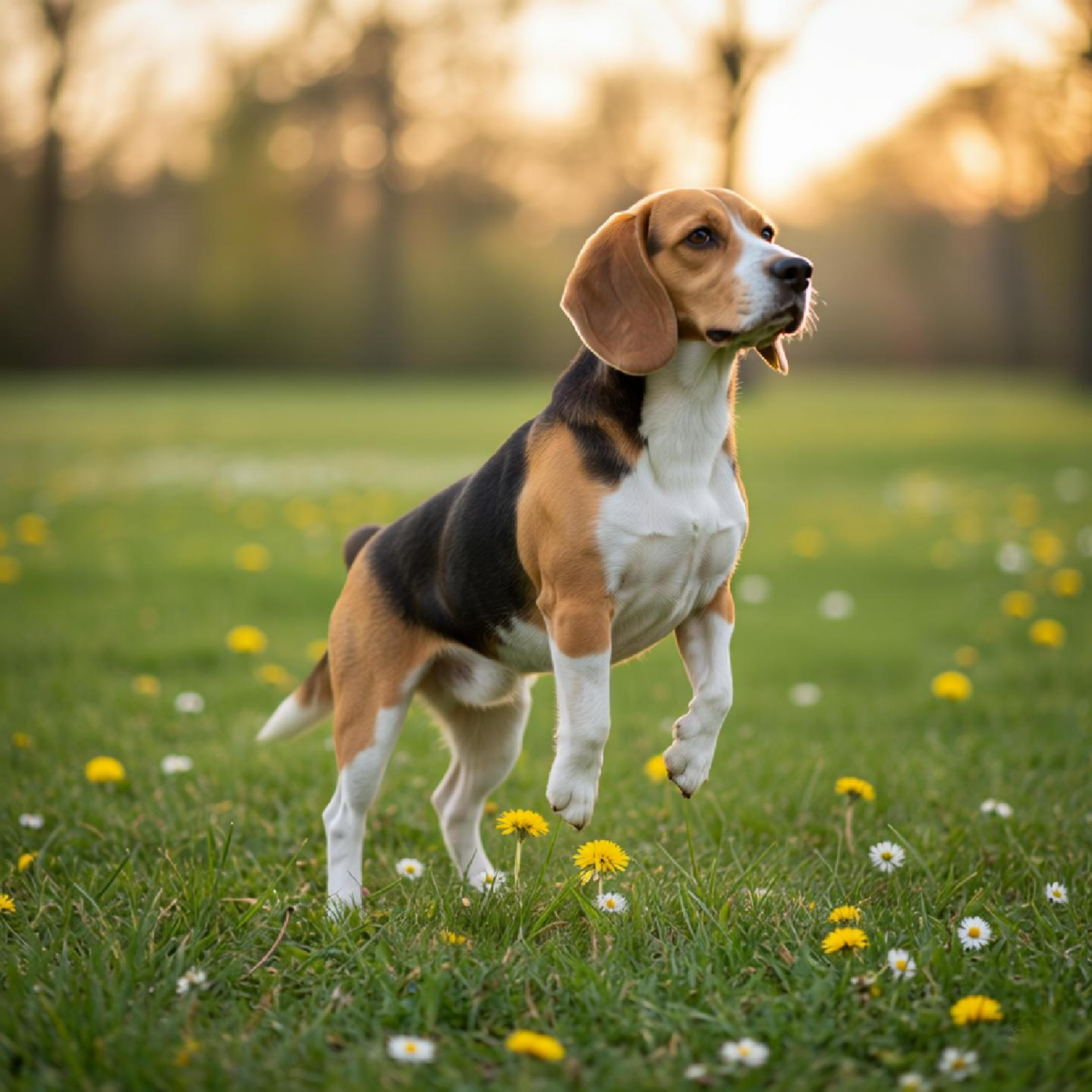} \\
        
        \raisebox{14pt}{\rotatebox{90}{\small Eagle}} &
        \includegraphics[width=\imgwidth]{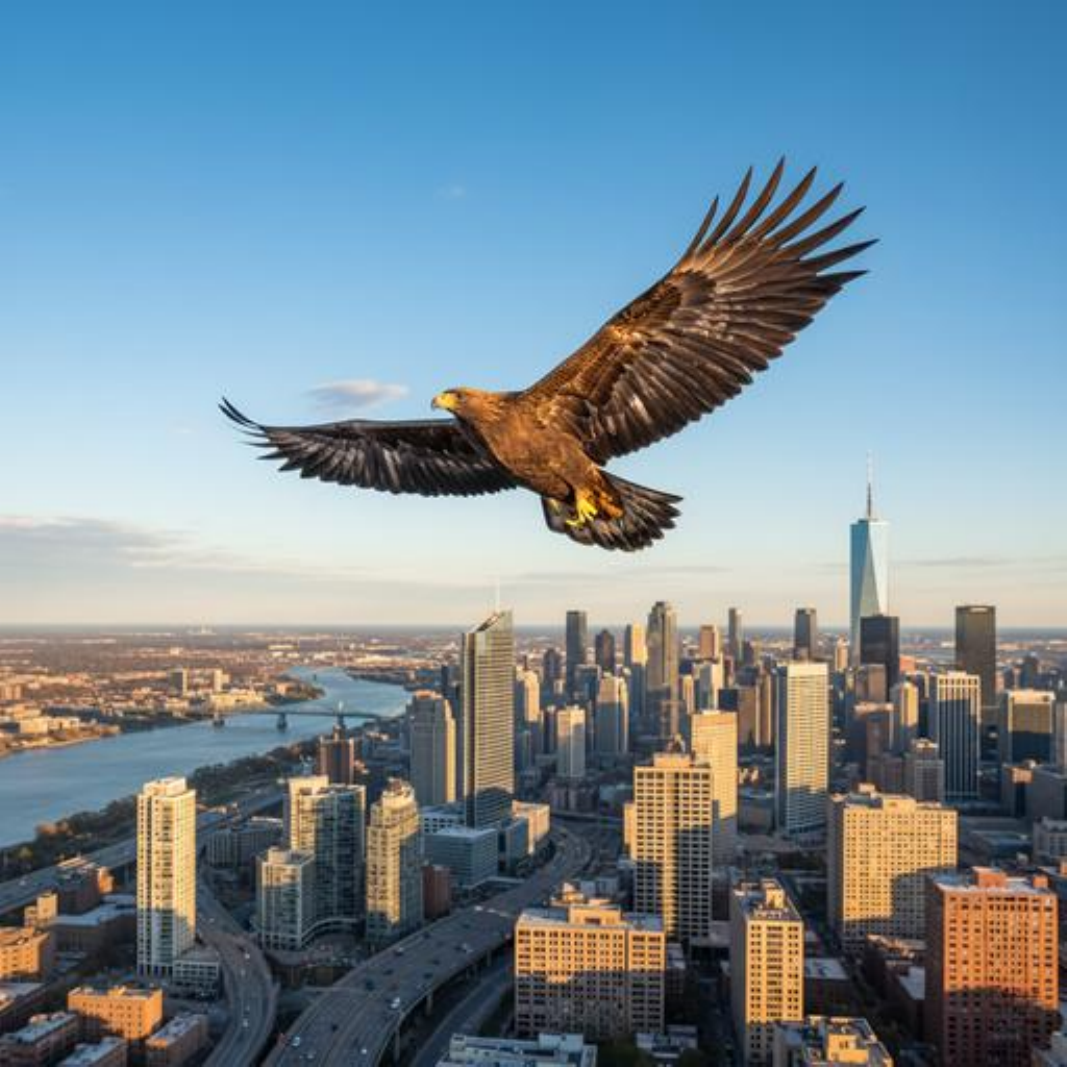} &
        \includegraphics[width=\imgwidth]{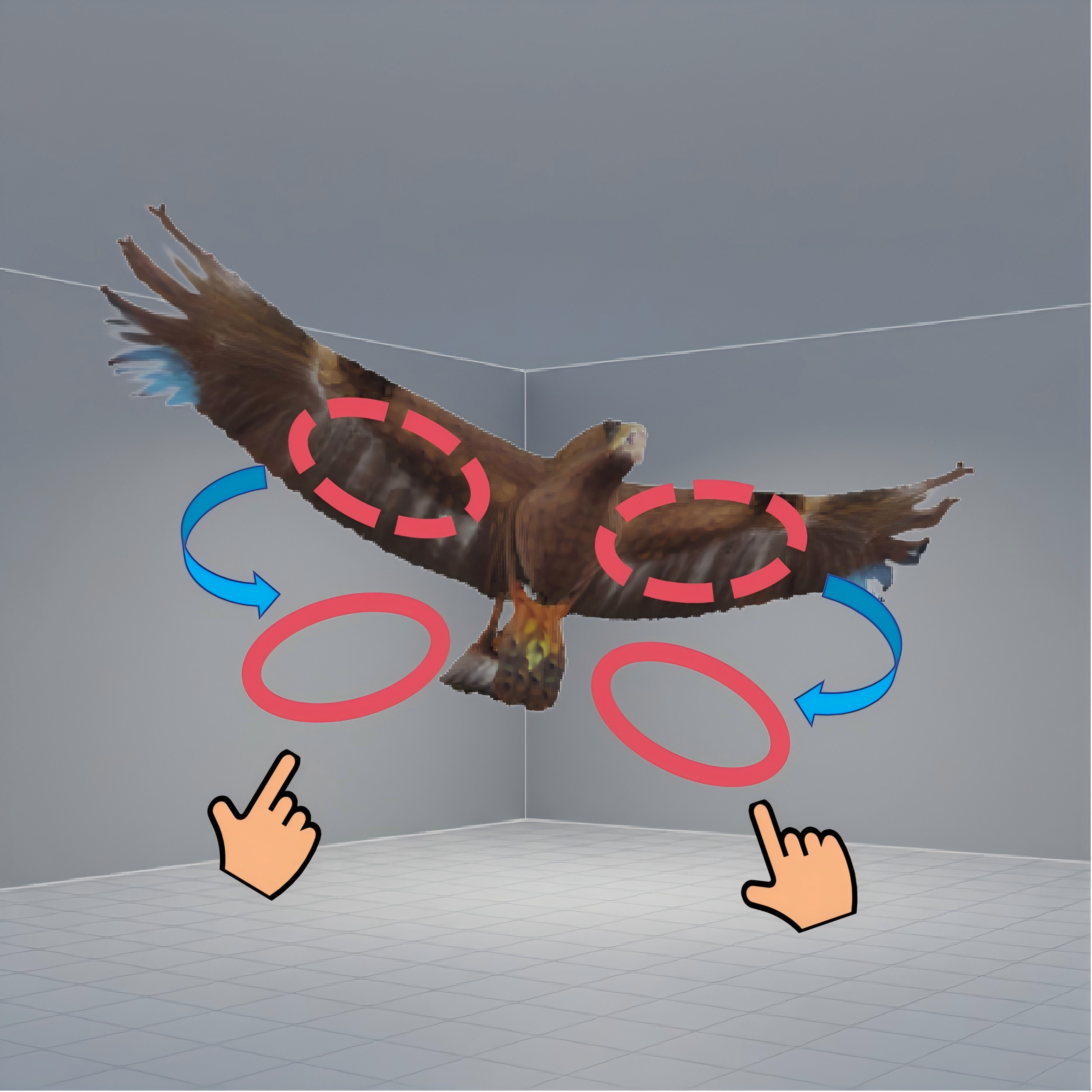} &
        \includegraphics[width=\imgwidth]{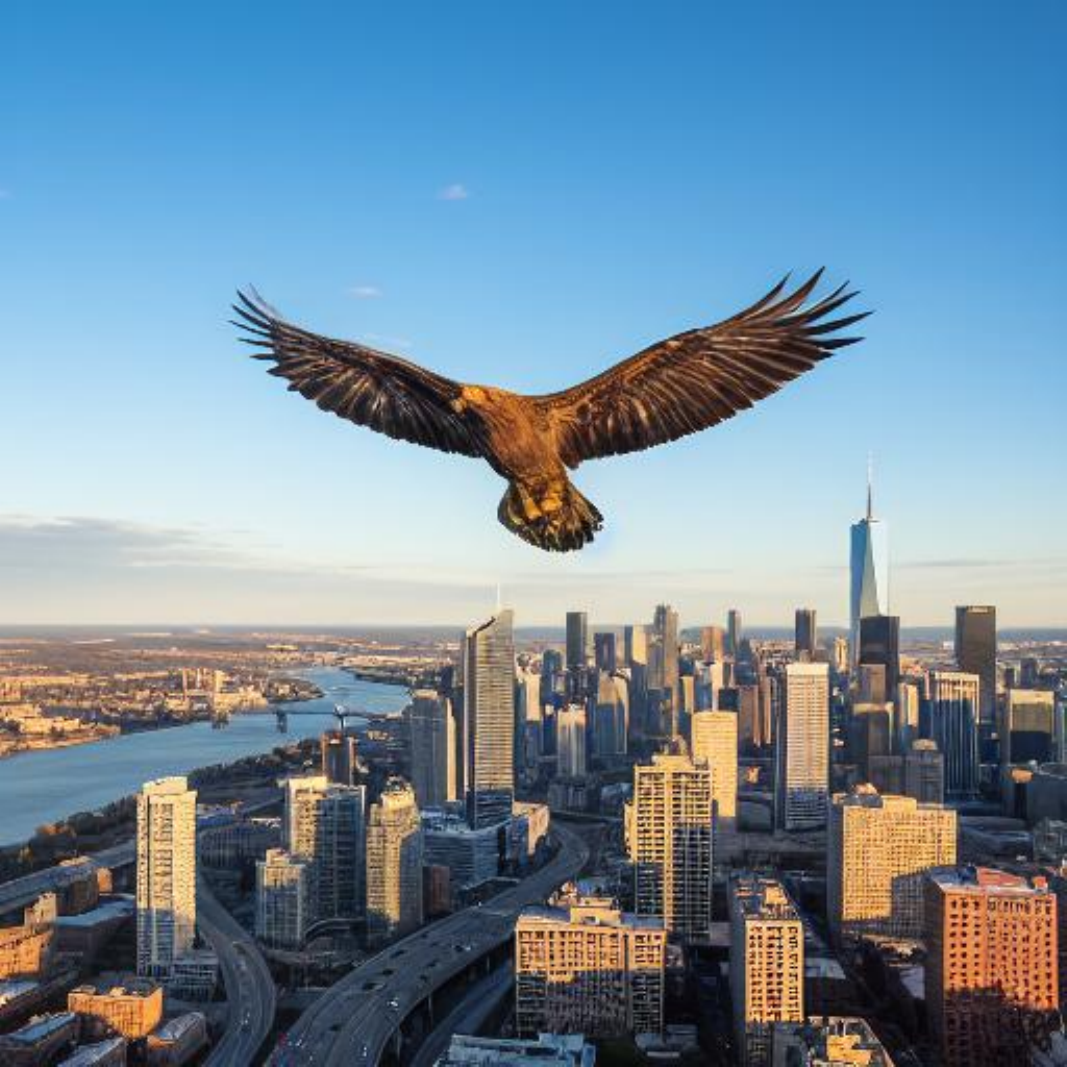} &
        \includegraphics[width=\imgwidth]{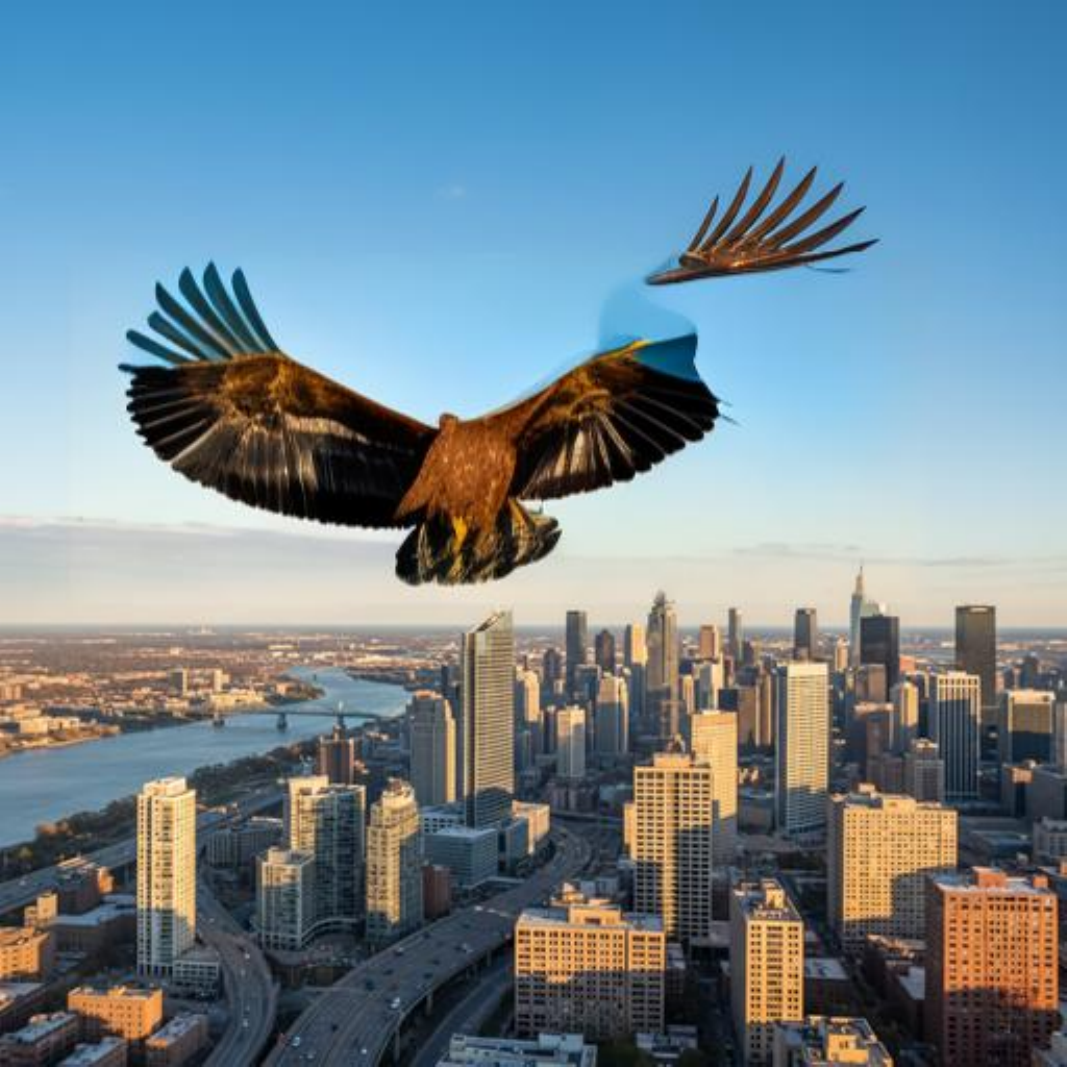} &
        \includegraphics[width=\imgwidth]{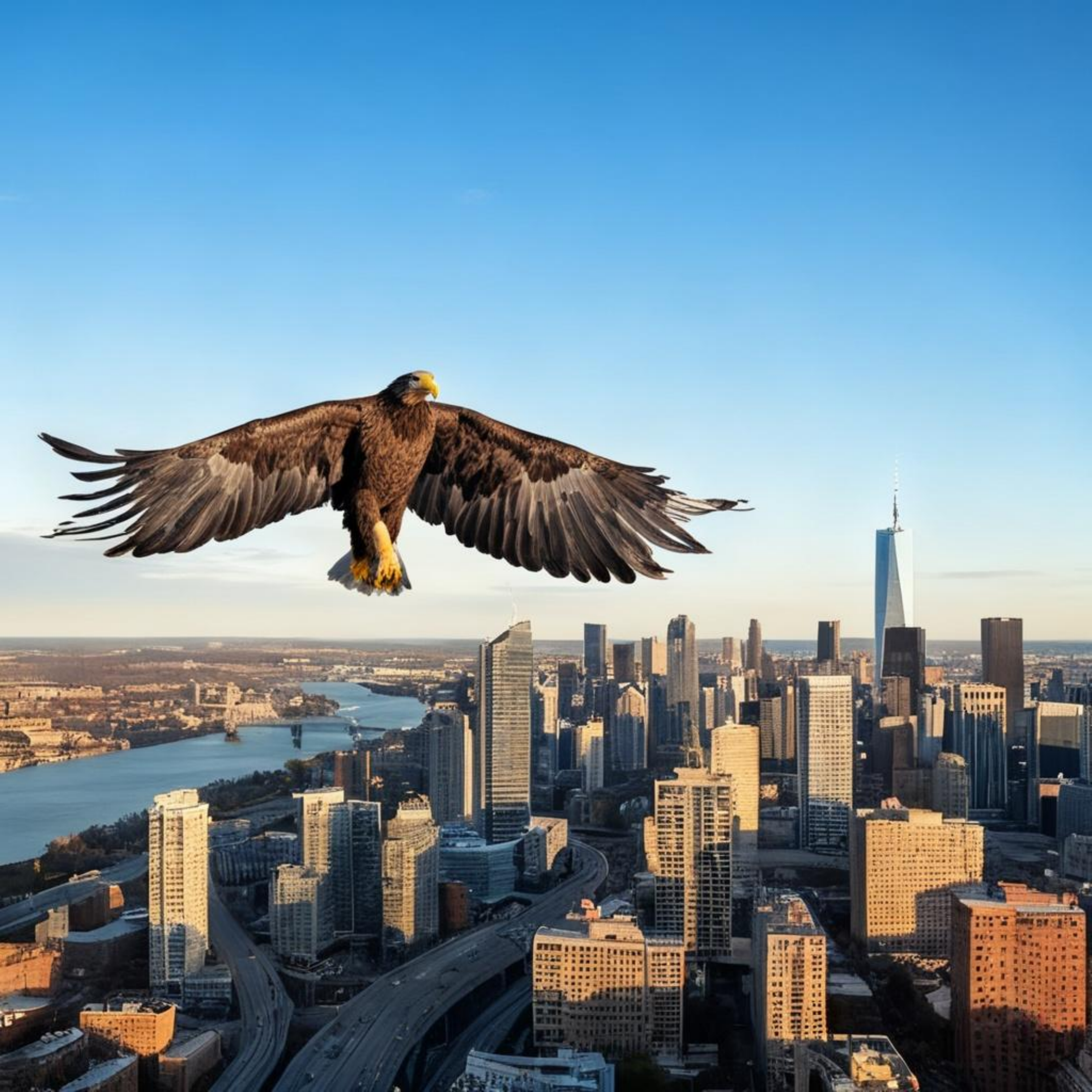} &
        \includegraphics[width=\imgwidth]{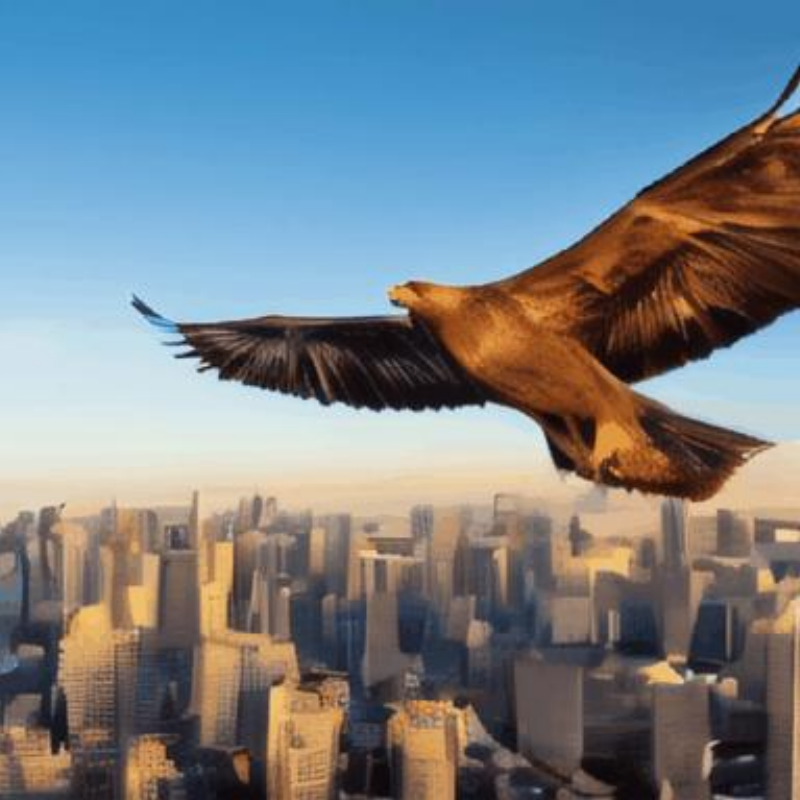} &
        \includegraphics[width=\imgwidth]{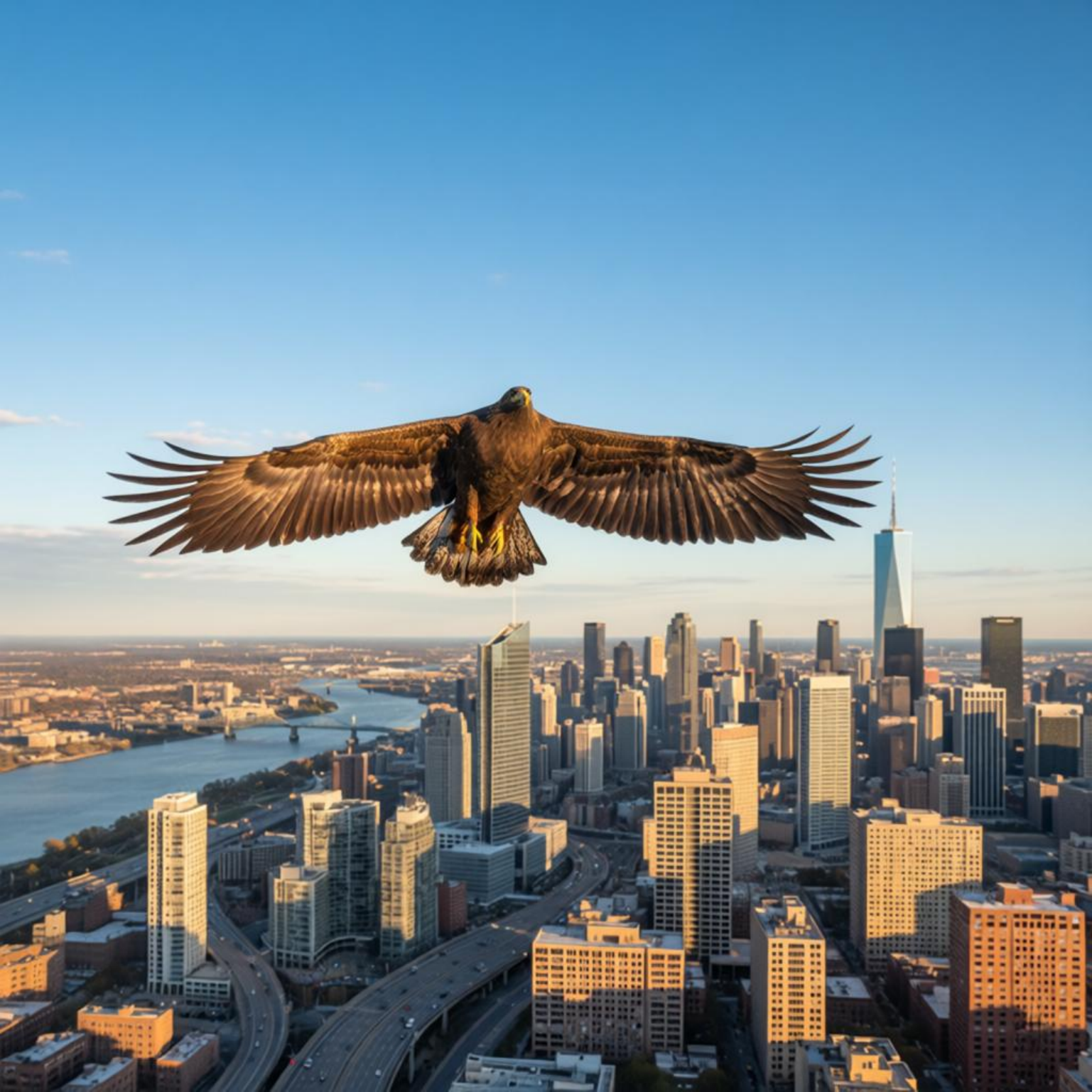} \\
        
        \raisebox{14pt}{\rotatebox{90}{\small Astronaut}} &
        \includegraphics[width=\imgwidth]{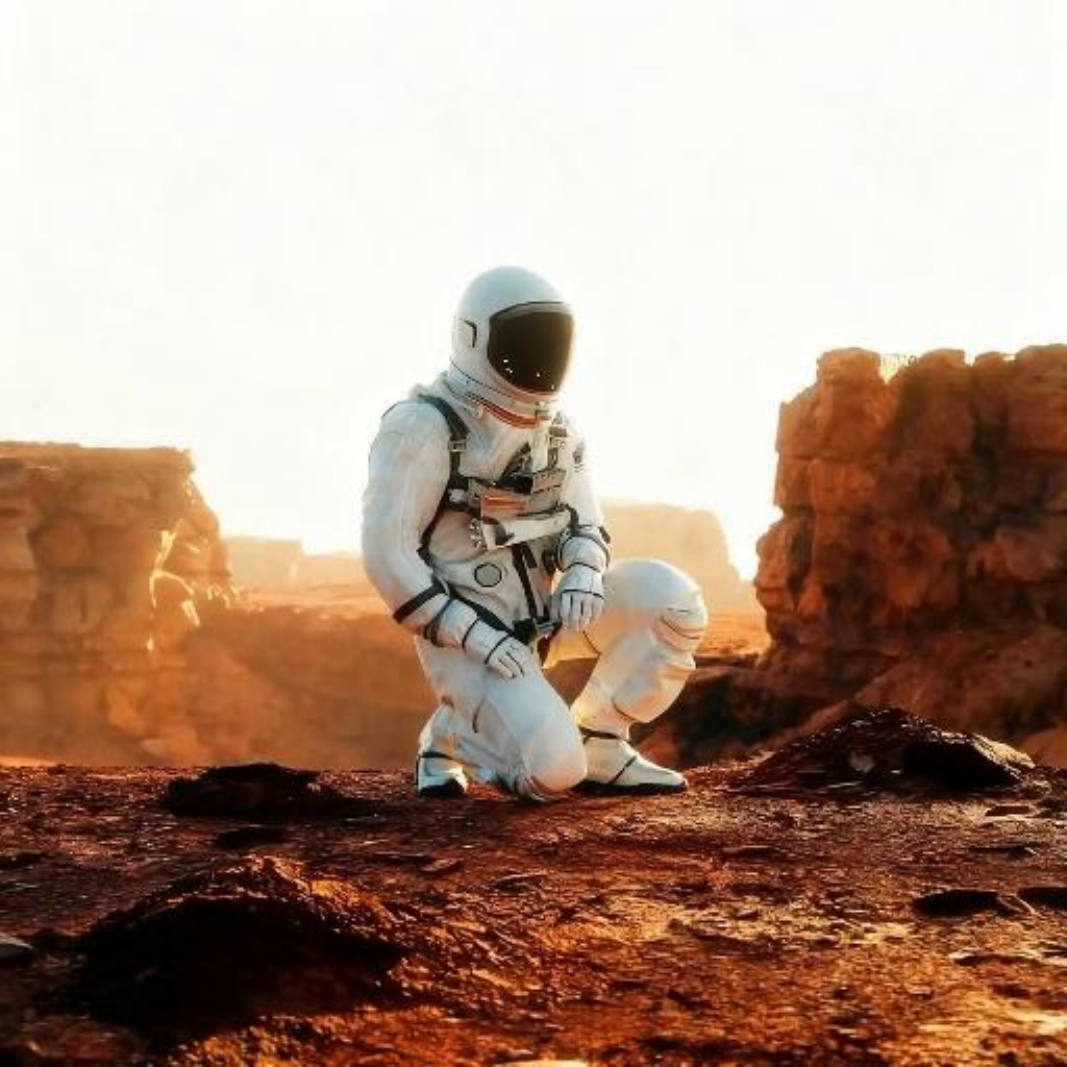} &
        \includegraphics[width=\imgwidth]{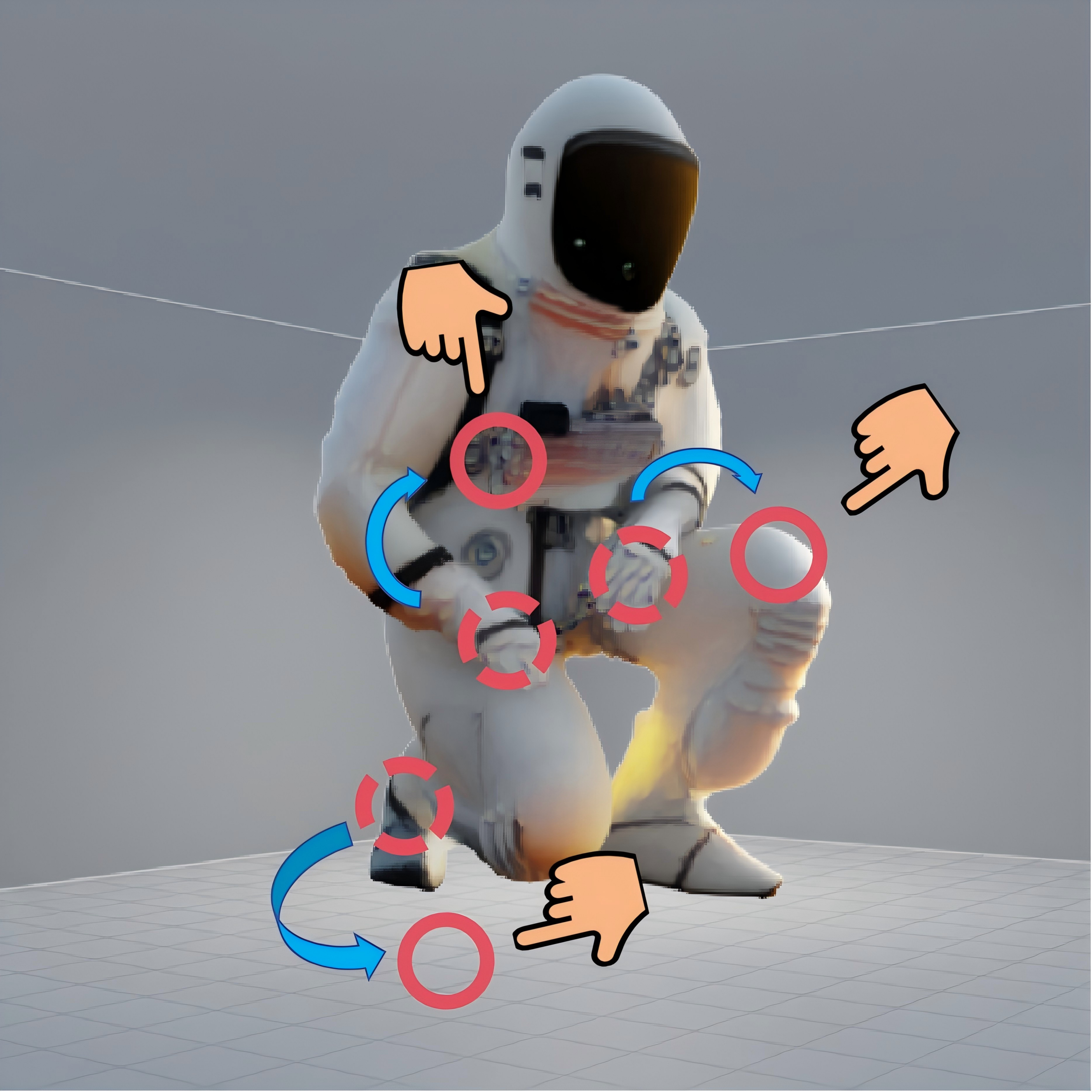} &
        \includegraphics[width=\imgwidth]{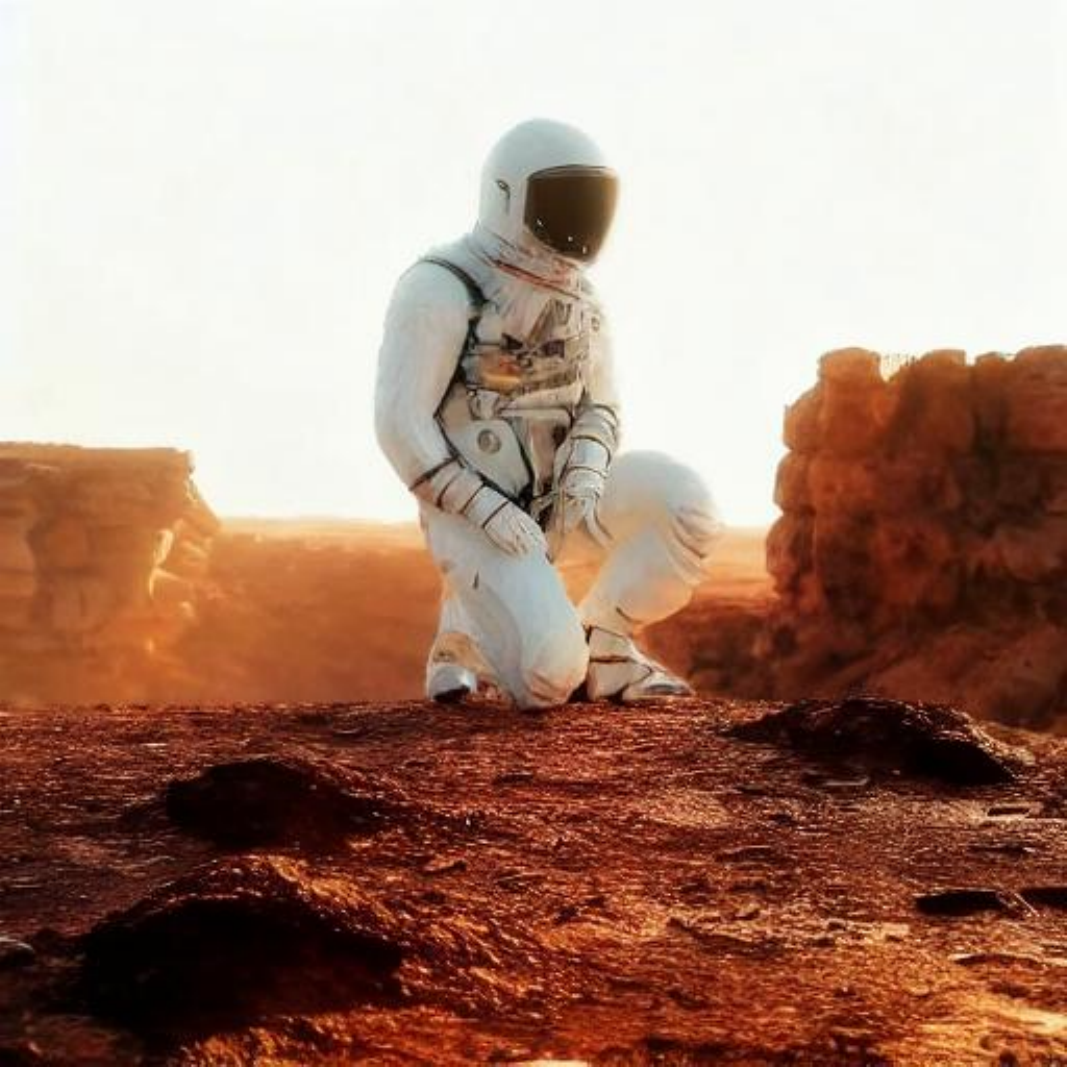} &
        \includegraphics[width=\imgwidth]{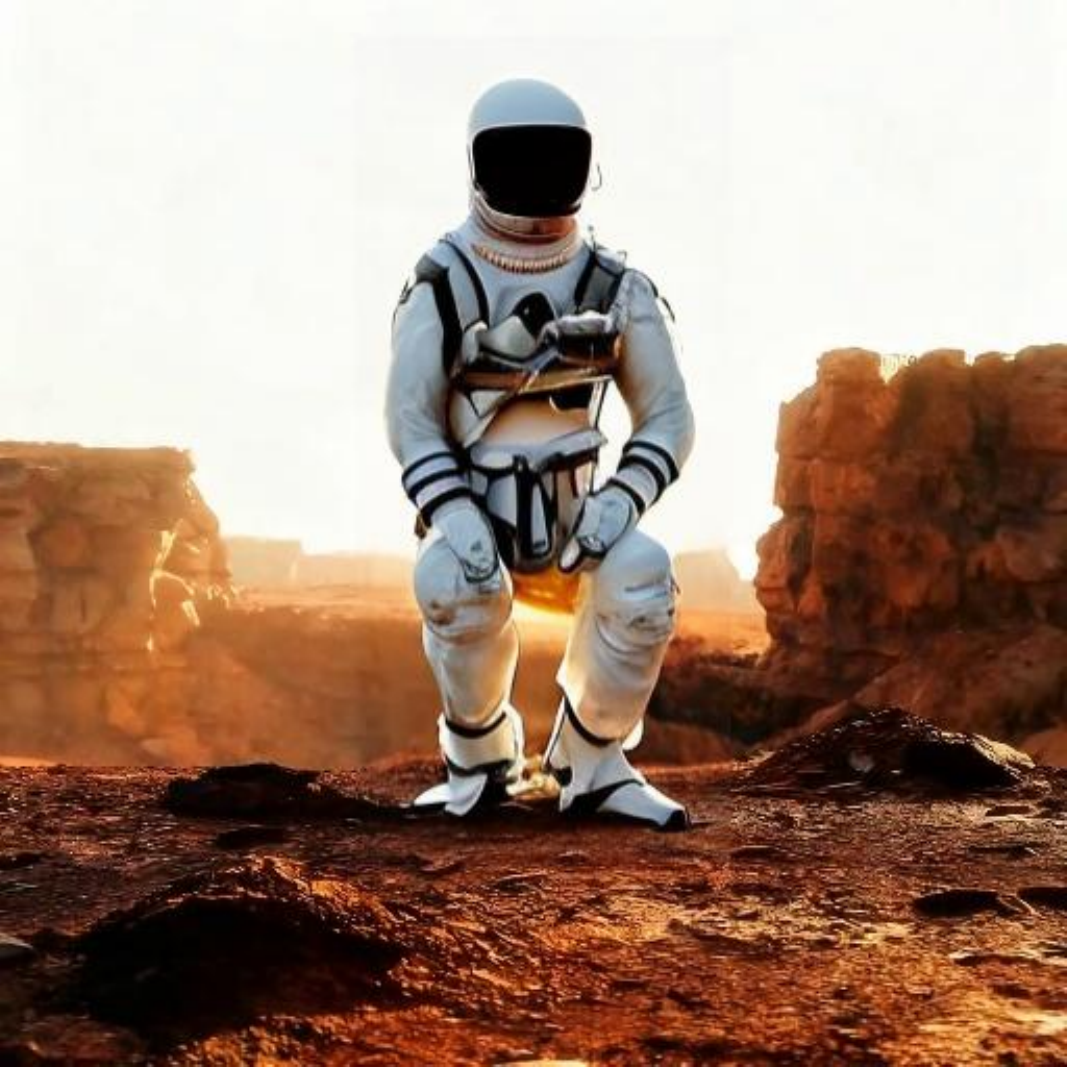} &
        \includegraphics[width=\imgwidth]{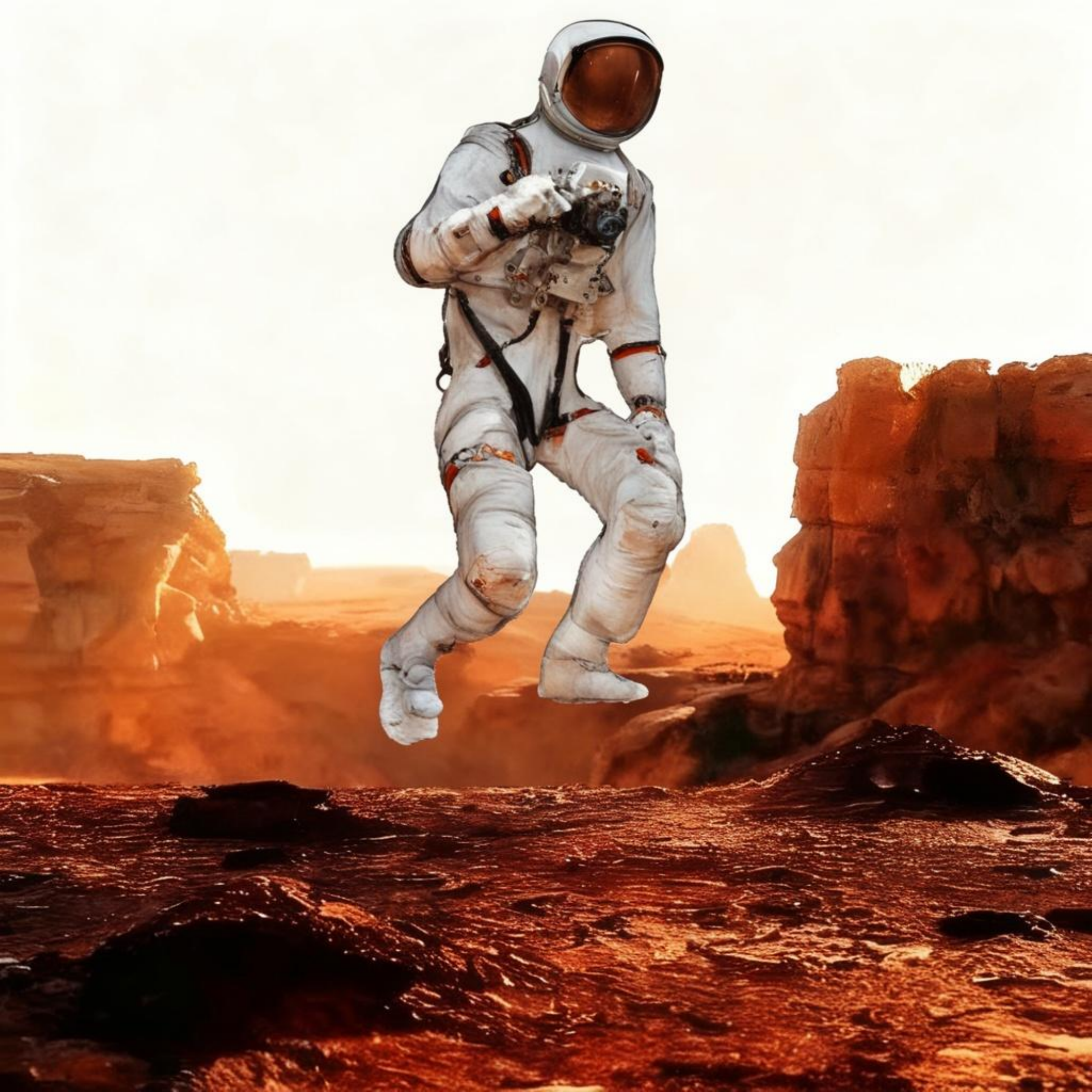} &
        \includegraphics[width=\imgwidth]{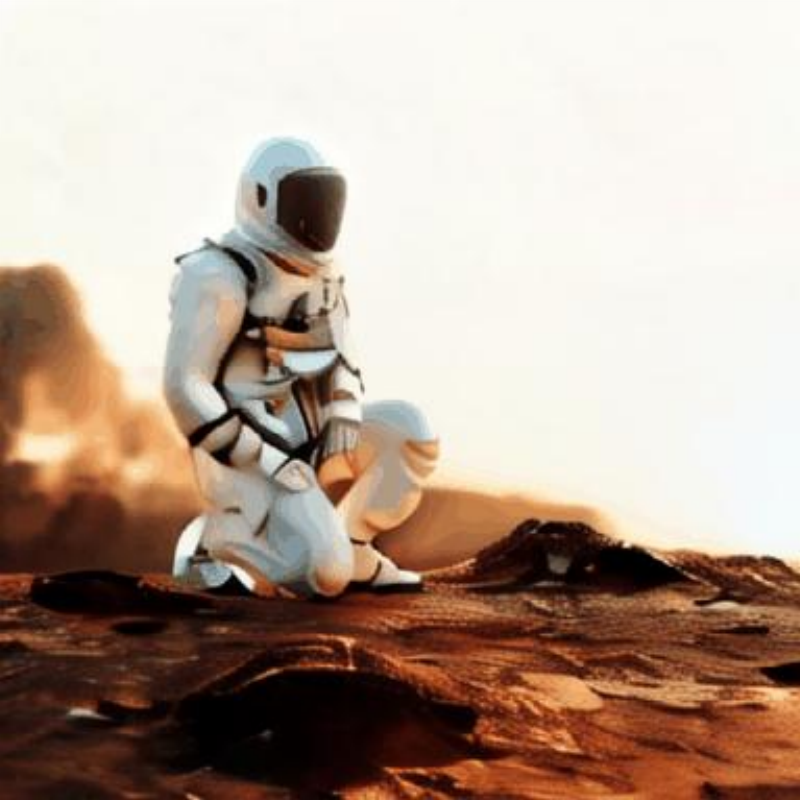} &
        \includegraphics[width=\imgwidth]{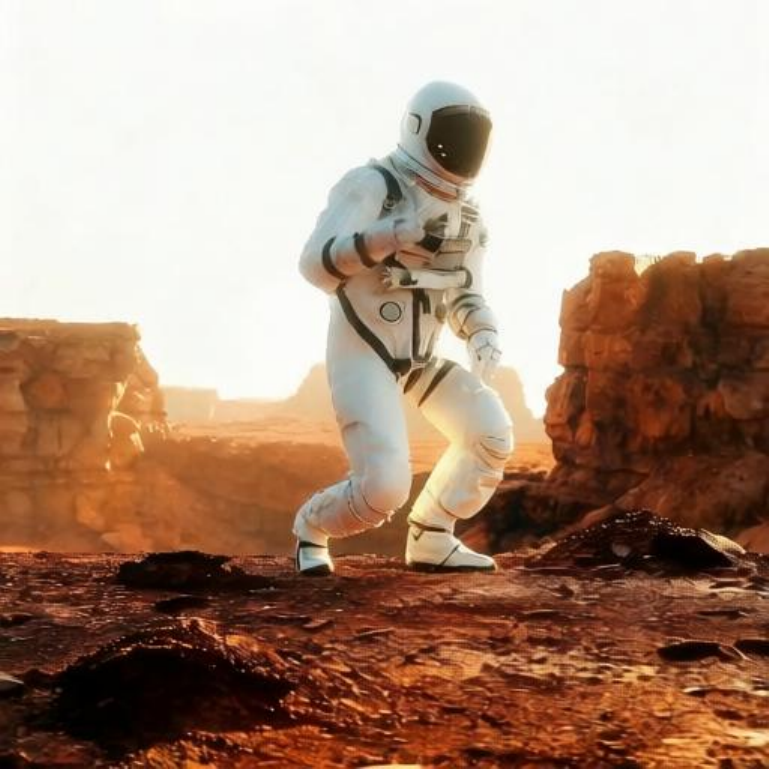} \\
        
        \raisebox{14pt}{\rotatebox{90}{\small Walle}} &
        \includegraphics[width=\imgwidth]{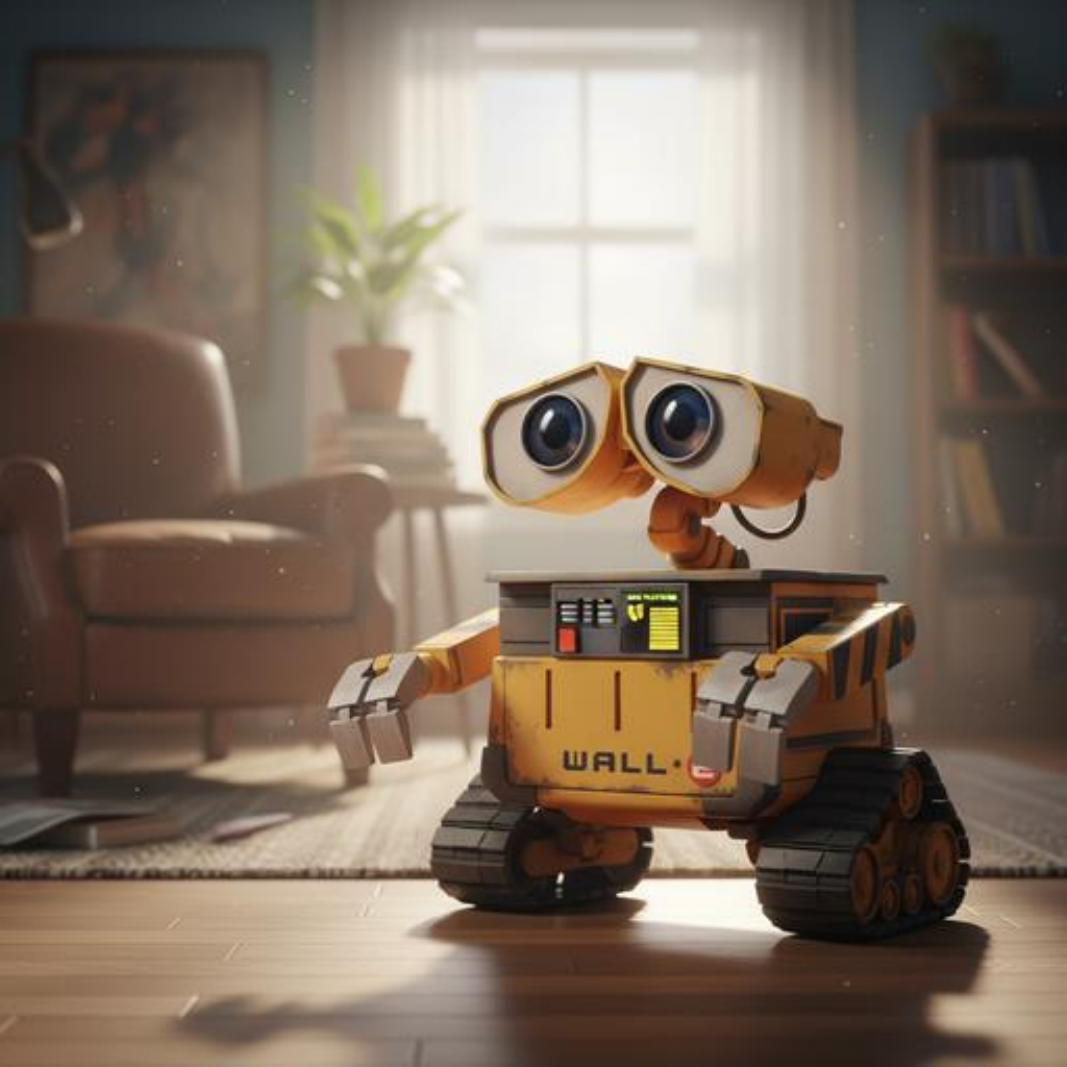} &
        \includegraphics[width=\imgwidth]{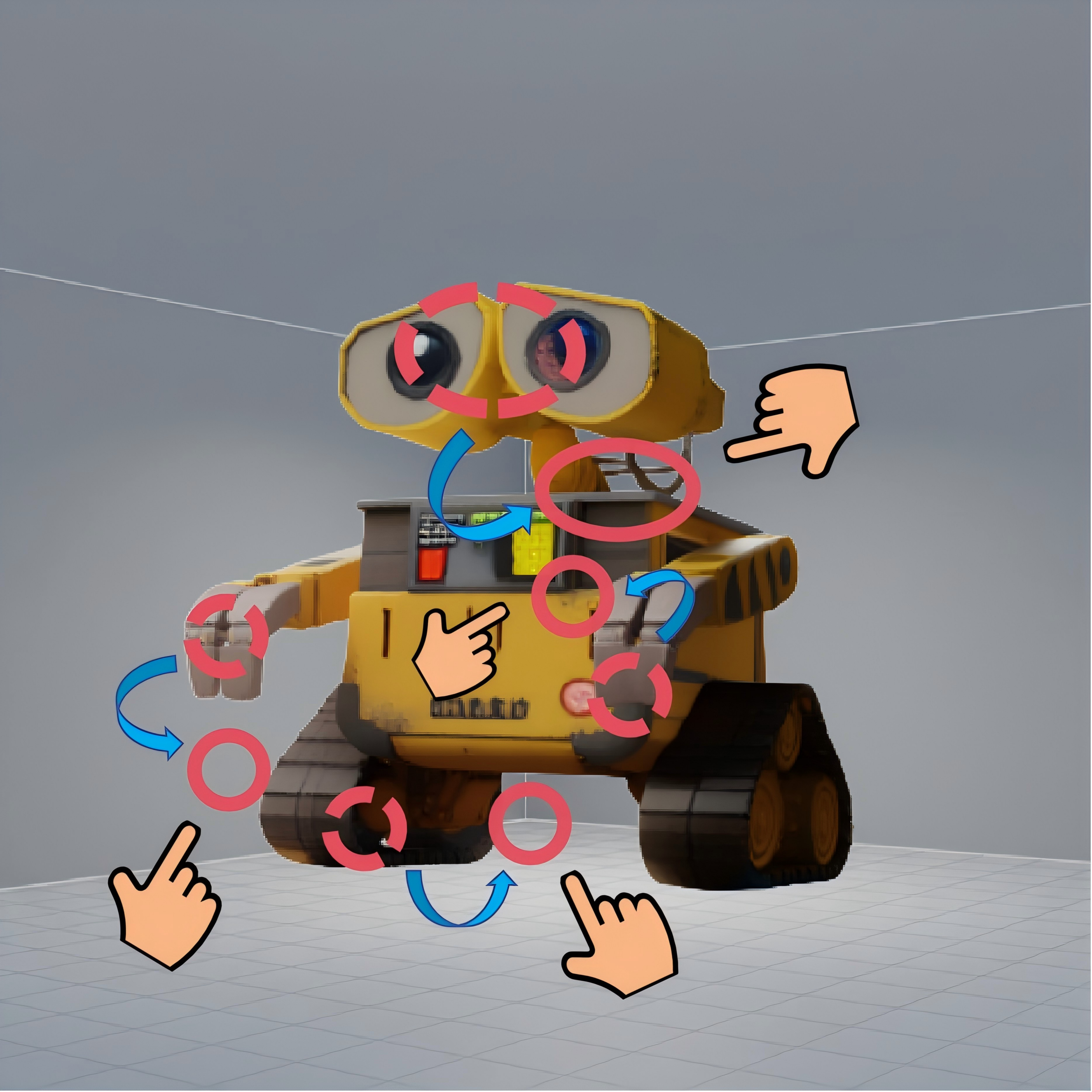} &
        \includegraphics[width=\imgwidth]{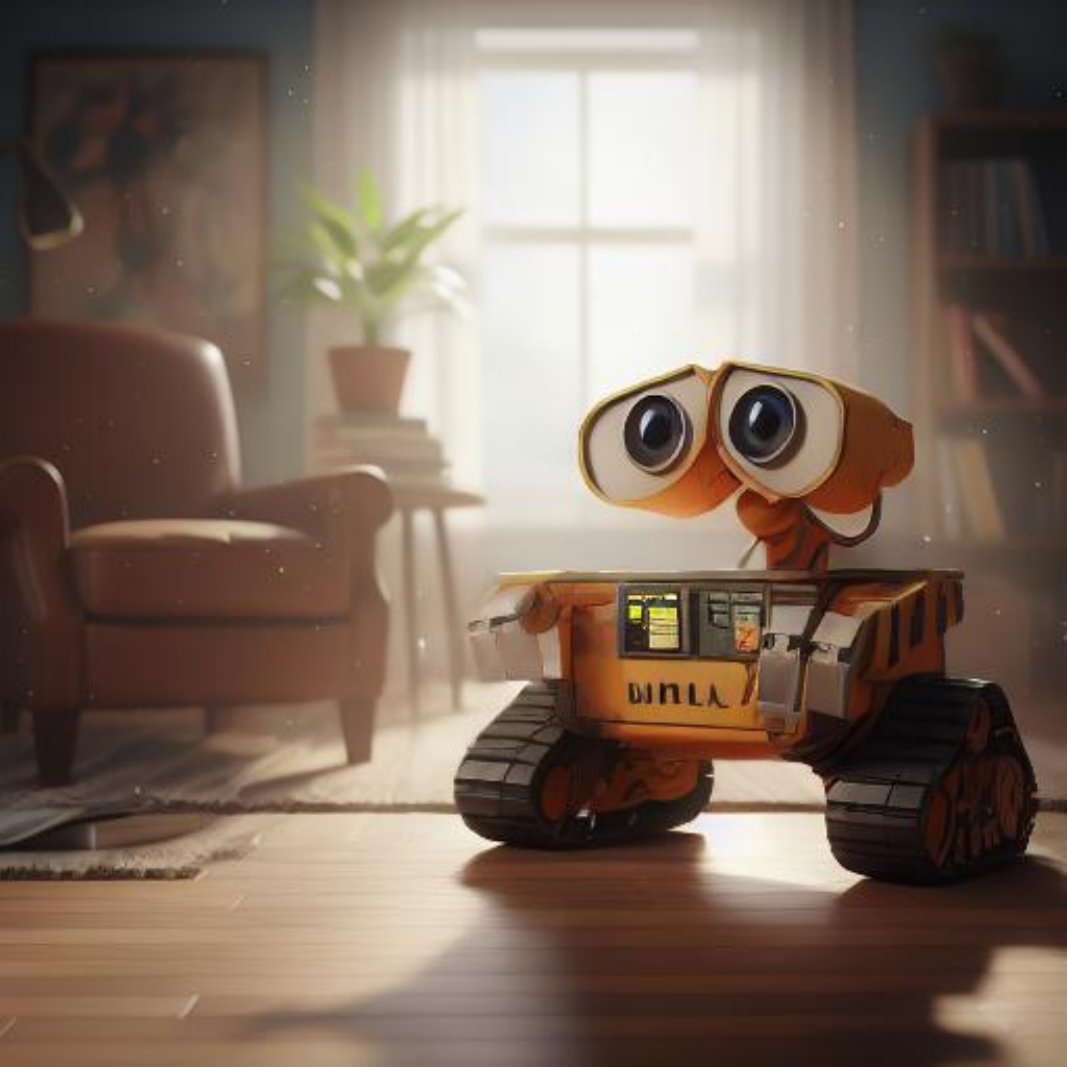} &
        \includegraphics[width=\imgwidth]{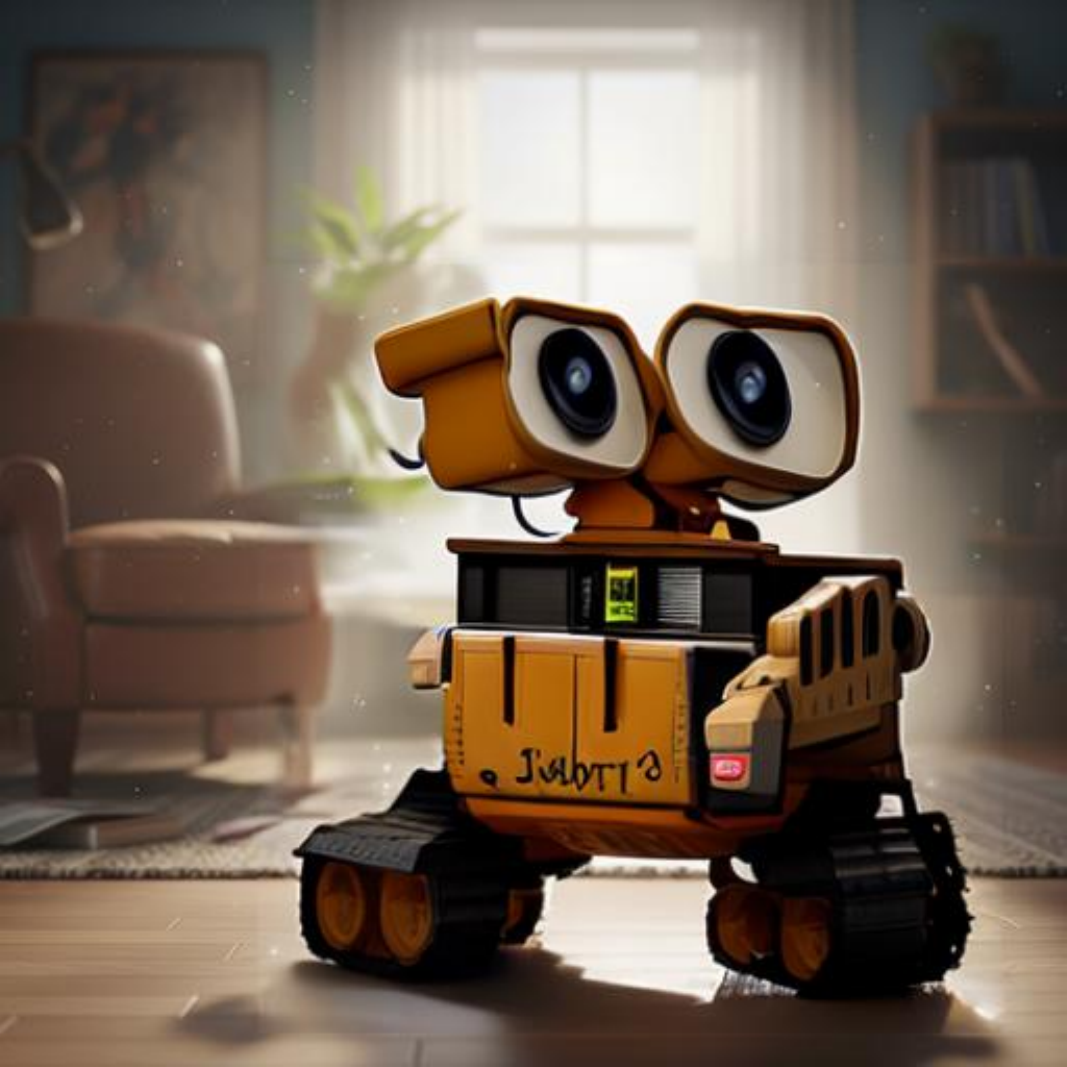} &
        \includegraphics[width=\imgwidth]{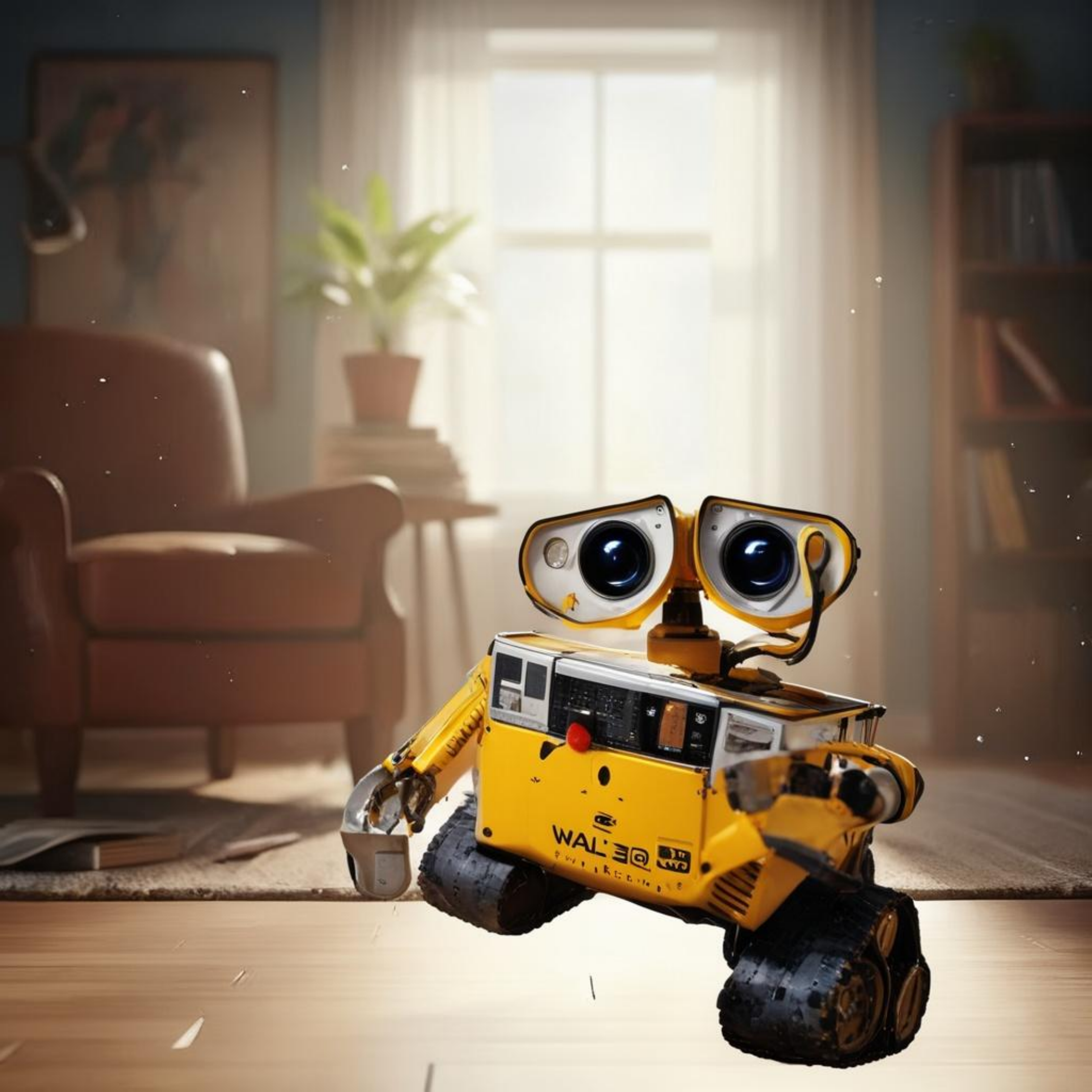} &
        \includegraphics[width=\imgwidth]{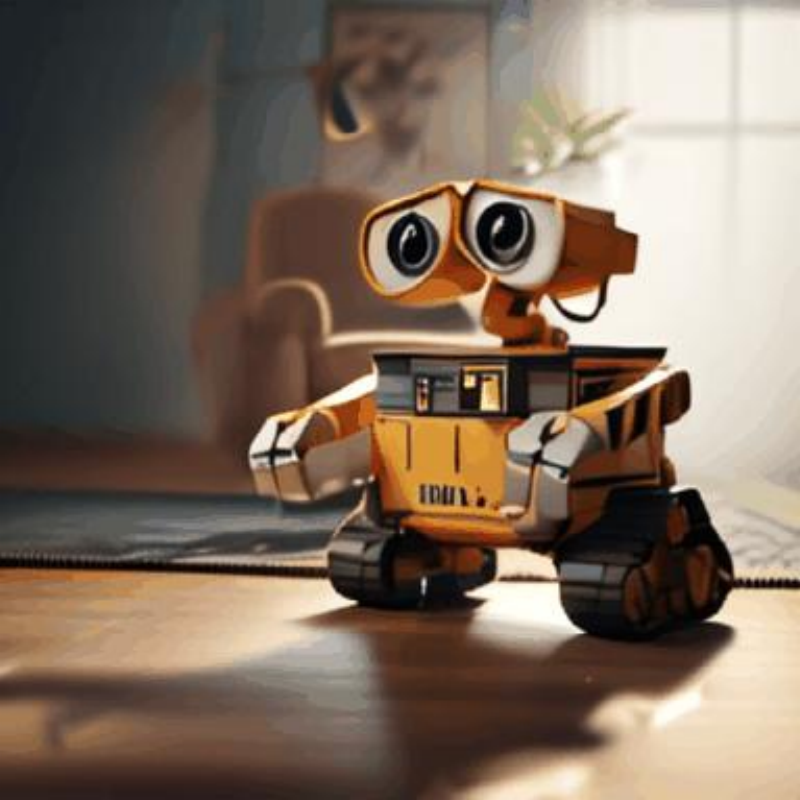} &
        \includegraphics[width=\imgwidth]{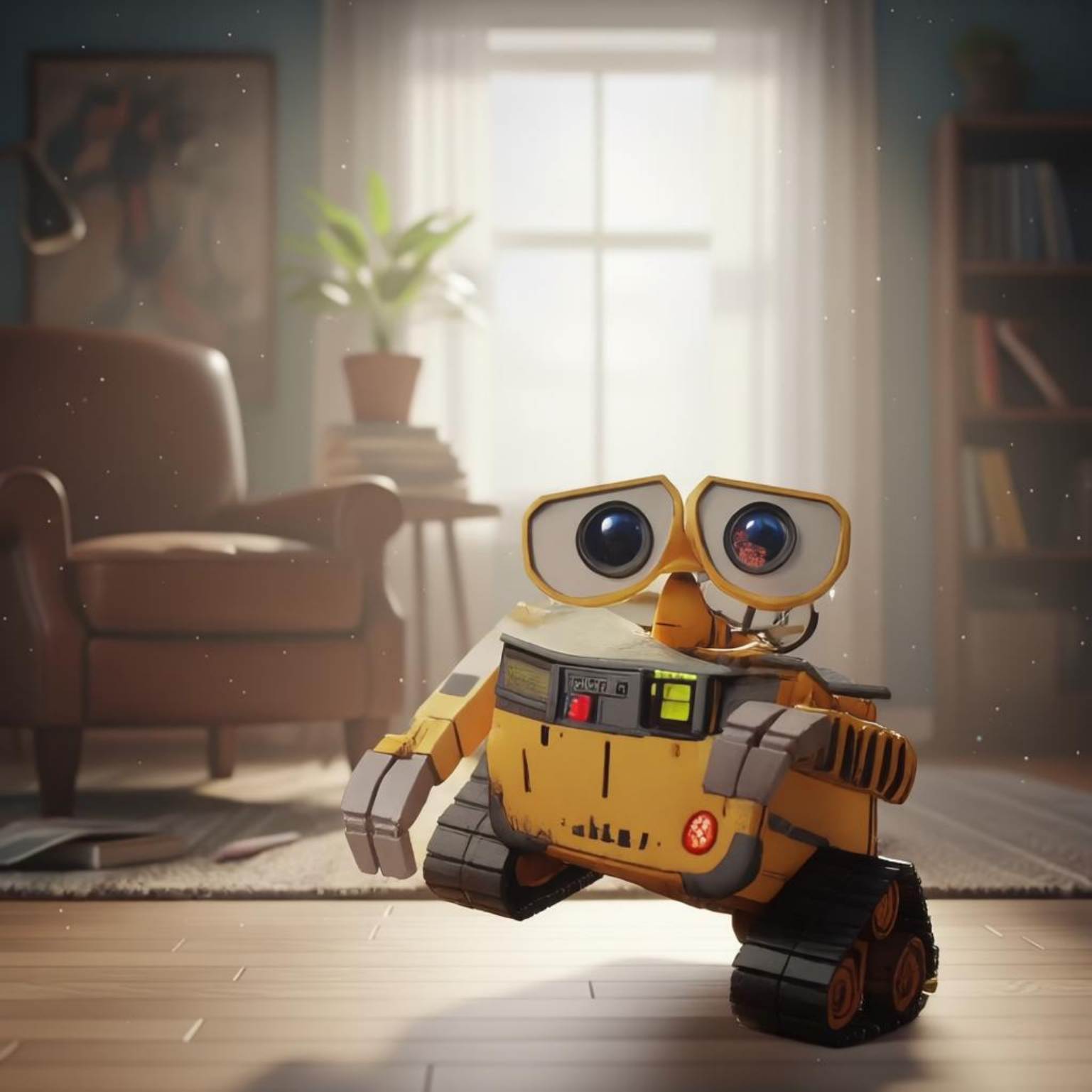} \\

    \end{tabular}
    \vspace{-2mm}
    \caption{Qualitative comparisons. We show results on 8 subjects (rows) across 5 methods (columns). Our method (Ours) is the only one that faithfully follows the non-rigid user guidance while maintaining photorealism. 2D methods fail to perform the 3D-aware edit, producing results nearly identical to the origin.}
    \label{fig:qualitative comparisons}
    \vspace{-4mm}
\end{figure*}

\section{Experiment}

\subsection{Implementation Details}
\label{sec:implementation_details}

For the composition module, we fine-tune a LoRA (rank=16) on the Qwen-Image-Edit base model using the AdamW optimizer. The training operates at a 1024×1024 resolution in bf16 precision. We utilize the LoRA+ framework with a base learning rate of 0.0001 and a multiplier of 3. Generation is guided by a task-specific text instruction that explicitly enforces contour preservation and lighting harmonization based on the reference image.

\subsection{Evaluation Datasets and Baselines}
\label{sec:datasets_and_baselines}

\paragraph{Datasets.} For quantitative and qualitative evaluation, we curate a diverse benchmark of 50 challenging subjects, and 8 of them are illustrated in Fig.~\ref{fig:qualitative comparisons}. This benchmark spans a wide range of categories, including animals (Butterfly, Dog, Eagle), humanoids (Knight, Astronaut), inanimate objects (Chair), and animated characters (Cartoon Dog, Walle). These high-quality subjects feature clear foreground objects, varied poses, and complex backgrounds, providing a challenging and consistent testbed for evaluating both non-rigid controllability and compositional realism.
\vspace{-3mm}
\paragraph{Baselines.} We compare ObjectMorpher against a suite of state-of-the-art 2D and 3D-aware editing methods. These include 2D drag-based models (DragDiffusion~\cite{shi2024dragdiffusion} and DragAnything~\cite{wu2024draganything}), 2D mask-based editing (Anydoor~\cite{chen2023anydoor}), and 3D-aware editing (ImageSculpting~\cite{yenphraphai2024image}).
\vspace{-3mm}
\paragraph{Evaluation Protocol.}
To establish a fair comparison, we define a precise non-rigid editing task for each subject (the "Guidance" column in Fig.~\ref{fig:qualitative comparisons}). This exact guidance is provided to all baselines and ObjectMorpher. The original image serves as the primary reference for evaluating both preservation and realism, while the user guidance serves as the target to assess overall controllability.

\subsection{Quantitative Evaluation}
\label{sec:quantitative_evaluation}

The core dimensions to measure the performance of user-controlled editing include (1) the quality of the result images and (2) the satisfaction with the editing effects. We evaluate the first point with quantitative metrics. To assess the second point, we conducted a user study where participants scored the results from different methods

\subsubsection{Human Evaluation}
    \label{sec:human_evaluation}
    Standard automatic metrics (e.g., FID) are often misleading in controllable editing, as they can perversely reward methods that fail to execute the user's edit. Therefore, we conduct a rigorous user study as our primary evaluation.
    \vspace{-4mm}
    \paragraph{Setup.}
    We presented 20 participants with the 8 editing tasks from our benchmark (Fig.~\ref{fig:qualitative comparisons}). For each task, participants were shown the original image, guidance, and the results from all baselines and our method (in randomized order). We excluded Image Sculpting on the 'Butterfly' subject due to its failure on this instance. Participants rated each result on a 1-5 Likert scale across three criteria:
    \begin{itemize}
    \item \textbf{Guidance Following (GF):} How well does the edit match the requested pose/shape change? 
    \item \textbf{Realism / Style Consistency (RE):} How photorealistic is the result, and does it match the original image's style? 
    \item \textbf{Identity Preservation (ID):} Does the object retain its identity (e.g., the same dog, the same knight)? 
\end{itemize}

\begin{figure}[t]
    \centering
    \includegraphics[width=1.0\columnwidth]{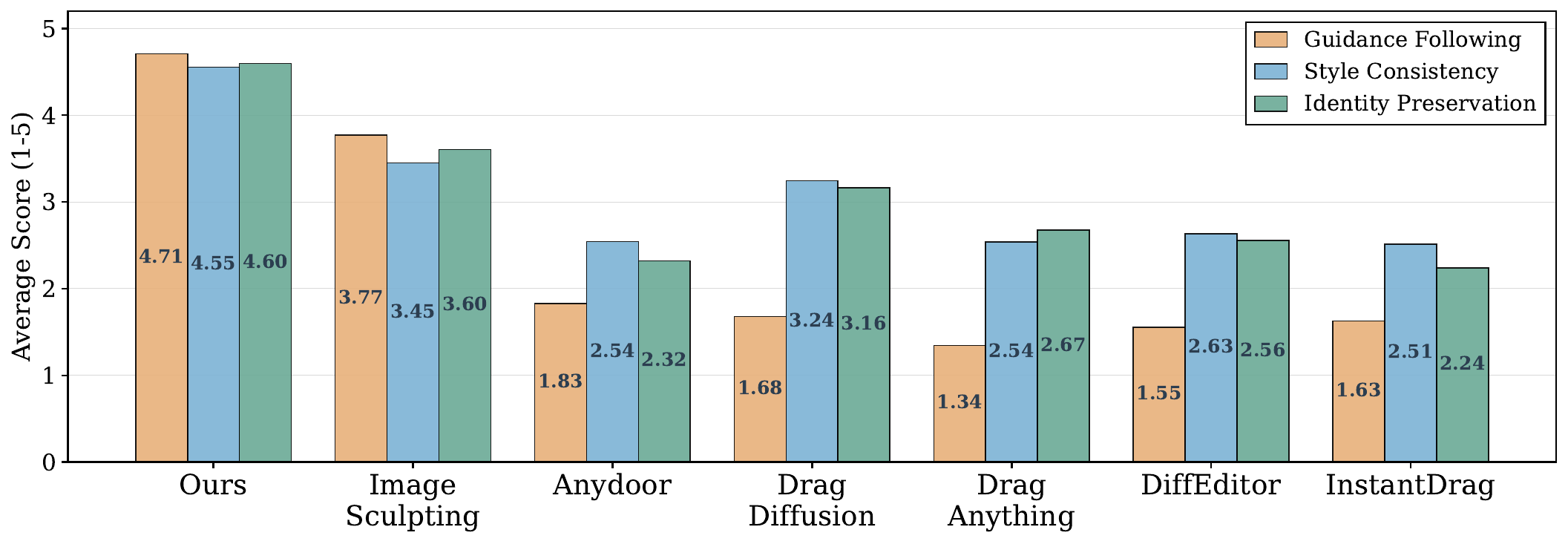}
    \vspace{-7mm}
    \caption{Visual results of our user study. Our method is consistently preferred across all three metrics (Guidance Following, Style Consistency, and Identity Preservation) over all baselines.}
    \label{fig:user_study_chart}
    \vspace{-5mm}
\end{figure}

The results are summarized in Fig.~\ref{fig:user_study_chart}. It demonstrates that our method can accomplish challenging editing tasks, including large-scale pose editing, rotation, etc., while maintaining a high degree of realism and object consistency.

\subsubsection{Quantitative Comparison}
We use the automatic perceptual to evaluate the fidelity and quality of edited images with the original images as reference:
Kernel Inception Score (KID)~\cite{binkowski2018demystifying}, Learned Perceptual Image Patch Similarity (LPIPS)~\cite{zhang2018unreasonable}, and Single Image Fréchet Inception Distance (SIFID)~\cite{shaham2019singan}. Tab.~\ref{tab:baseline} summarizes our comparisons with baselines.

\begin{table}[h]
  \centering
  \caption{Quantitative evaluation. RI represents real-time interaction and MT represents model inference time.($^*$DragDiffusion often edits images with minimal modifications.)}
  \label{tab:baseline}
  \vspace{-2.5mm}
  \setlength\tabcolsep{0pt}
    \begin{tabular*}{\linewidth}{@{\extracolsep{\fill}} lSSSSSSSSSSSSSSSS}
        \toprule[1pt]{\footnotesize{Methods}} & {\footnotesize{LPIPS $\downarrow$}} & {\footnotesize{SIFID $\downarrow$}} & {\footnotesize{KID $\downarrow$}} & {\footnotesize{RI}} & {\footnotesize{MT $\downarrow$}} \\
          \midrule
        \footnotesize{DragGAN}
        & \footnotesize{0.550} & \footnotesize{16.091} & \footnotesize{-0.056} & \footnotesize{\checkmark} & \footnotesize{$<$ 10s}\\
        \footnotesize{ImageSculpting}
        & \footnotesize{0.178} & \footnotesize{14.372} & \footnotesize{-0.055} & \footnotesize{\checkmark} & \footnotesize{$\sim$ 30m}\\
        \footnotesize{DragDffusion$^*$}
        & \footnotesize{\textbf{0.117}} & \footnotesize{\textbf{6.573}} & \footnotesize{\textbf{-0.075}} & \footnotesize{$\times$} & \footnotesize{$\sim$ 2m}\\
        \footnotesize{DragAnything}
        & \footnotesize{0.655} & \footnotesize{22.476} & \footnotesize{-0.047} & \footnotesize{$\times$} & \footnotesize{$\sim$ 70s}\\
        \footnotesize{Anydoor}
        & \footnotesize{0.173} & \footnotesize{\underline{9.512}} & \footnotesize{-0.049} & \footnotesize{$\times$} & \footnotesize{$\sim$ 10s}\\
        \footnotesize{InstantDrag}
        & \footnotesize{0.218} & \footnotesize{{13.744}} & \footnotesize{-0.042} & \footnotesize{$\times$} & \footnotesize{$\sim$ 10s}\\
        \footnotesize{DiffEditor}
        & \footnotesize{0.142} & \footnotesize{{12.589}} & \footnotesize{-0.051} & \footnotesize{$\times$} & \footnotesize{$\sim$ 10s}\\
        \footnotesize{ObjectMorpher}
        & \footnotesize{\underline{0.127}} & \footnotesize{{10.896}} & \footnotesize{\underline{-0.059}} & \footnotesize{\textbf{\checkmark}} & \footnotesize{$\sim$ 20s}\\

        \bottomrule[1pt]
    
      \end{tabular*}
  \vspace{-6mm}
\end{table}

\paragraph{Analysis.}

Our analysis reveals that automatic fidelity metrics(LPIPS/SIFID) are misleading, as they perversely reward methods that fail to execute the user's edit. Baselines like DragDiffusion and Anydoor achieve deceptively low error scores (Table~\ref{tab:baseline}) precisely because they fail to perform the specified manipulation: DragDiffusion makes minimal modifications; DiffEditor and InstantDrag struggle with identity preservation while Anydoor uses imprecise mask guidance. This is confirmed by our User Study (Fig.~\ref{fig:user_study_chart},where both received catastrophic "Guidance Following" score (DragDiffusion:1.68; Anydoor: 1.83). In contrast, our method achieves the highest GF score (4.71) while maintaining competitive fidelity, providing a superior balance.

We also evaluated the ease of editing, summarized in Tab.~\ref{tab:baseline}, using two metrics: real-time interaction (RI) and model inference time (MT). RI measures the ability to provide users with instant visual feedback in response to input controls, such as clicking and dragging, either through final results or intermediate outputs. DragGAN achieves borderline RI capability, with a few seconds of delay in responding to user controls. ImageSculpting and ObjectMorpher utilize proxy 3D assets (mesh and 3DGS, respectively) to enable real-time interaction, allowing for more flexible adjustments and efficient editing.

While RI measures the efficiency of interaction, MT measures the computational time of the model and is equally important, as it affects the total time consumed alongside the user's editing. ImageSculpting is notably unstable in editing and slow on MT due to its per-image optimization process, which includes 3D reconstruction with SDS and DreamBooth~\cite{ruiz2023dreambooth} fine-tuning. DragDiffusion requires a trained LoRA given the input image, which significantly raises the MT. DragAnything infers according to the trajectories specified by the user at a low speed. Anydoor, InstantDrag and DiffEditor can rapidly produces the edited output. The runtime of ObjectMorpher is primarily spent on Object Lifting ($\sim$10 s) and Diffusion Refinement ($\sim$10 s), while the core Interactive Drag stage achieves a near-instantaneous latency of 10 ms,

\subsection{Ablation Study}

\begin{figure}[t]
    \centering
    
    \captionsetup{justification=centering}
    \def\subfigwidth{0.19\linewidth}
    \def\rowvspace{1mm} 

    \begin{subfigure}[b]{\subfigwidth}
        \includegraphics[width=\linewidth]{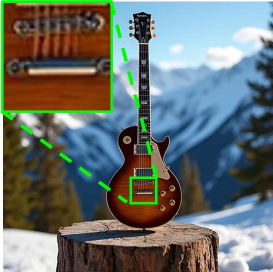} \vspace{\rowvspace}
        \includegraphics[width=\linewidth]{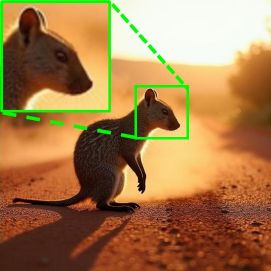}
        \caption{Input \\Image}
    \end{subfigure}
    \hfill 
    \begin{subfigure}[b]{\subfigwidth}
        \includegraphics[width=\linewidth]{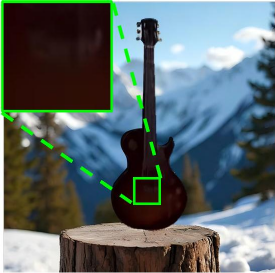} \vspace{\rowvspace}
        \includegraphics[width=\linewidth]{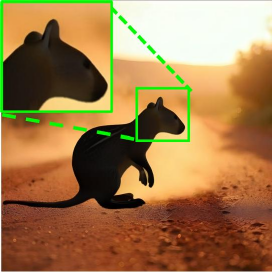}
        \caption{Mesh \\w/o GC}
    \end{subfigure}
    \hfill
    \begin{subfigure}[b]{\subfigwidth}
        \includegraphics[width=\linewidth]{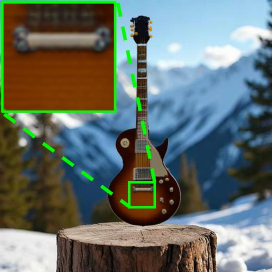} \vspace{\rowvspace}
        \includegraphics[width=\linewidth]{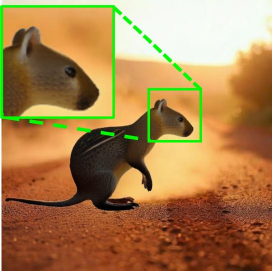}
        \caption{3DGS \\w/o GC}
    \end{subfigure}
    \hfill
    \begin{subfigure}[b]{\subfigwidth}
        \includegraphics[width=\linewidth]{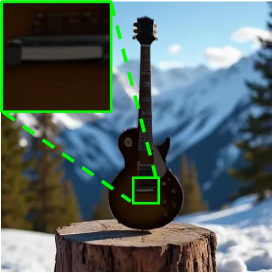} \vspace{\rowvspace}
        \includegraphics[width=\linewidth]{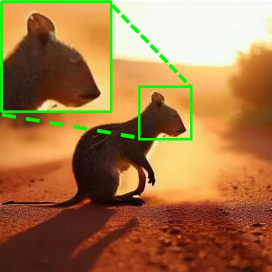}
        \caption{Mesh \\w/ GC}
    \end{subfigure}
    \hfill
    \begin{subfigure}[b]{\subfigwidth}
        \includegraphics[width=\linewidth]{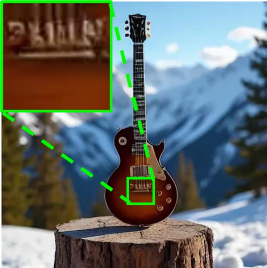} \vspace{\rowvspace}
        \includegraphics[width=\linewidth]{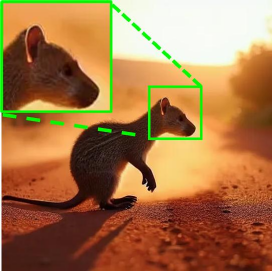}
        \caption{3DGS \\w/ GC}
    \end{subfigure}
    \vspace{-2mm}
    \caption{Ablation on 3D representations (3D GS vs. Mesh) and the use of Generative Composition (w/ GC vs. w/o GC).}
    \label{fig:ablation on representations}
\end{figure}

\paragraph{Mesh and GS.}
We evaluated different representations for 3D reconstruction from a single image, including mesh and 3D GS. First, the object is segmented from the image and then reconstructed into mesh and 3D GS representations, respectively. These representations are rendered from the initial viewpoint, and the resulting object images are embedded back into the original background. As shown in Fig \ref{fig:ablation on representations} (columns b and c), the results indicate that the 3D GS representation achieves a higher-quality rendering.

\paragraph{Generative Composition.} 
Directly embedding the rendered 3D object into the background often introduces unnatural artifacts. As observed in Fig.~\ref{fig:ablation on representations}(columns b and c), the results without Generative Composition (w/o GC) exhibit visible boundary seams and lighting inconsistencies between the foreground object and the background. To address this, our Generative Composition module utilizes diffusion-based inpainting to harmonize the object with the scene. Comparing column (c) with column (e), the ``w/ GC'' results show significantly improved visual coherence, where the object seamlessly blends into the environment while maintaining its structural integrity.

\begin{figure}[t]
    \centering
    \label{fig:deformation_constraints} 
    \vspace{0mm}
    \setlength{\tabcolsep}{3pt}

    \def\imgheightrowone{2cm} 
    \def\imgheightrowtwo{2cm} 

    \begin{tabular}{ccc}
        
        \includegraphics[width=\linewidth]{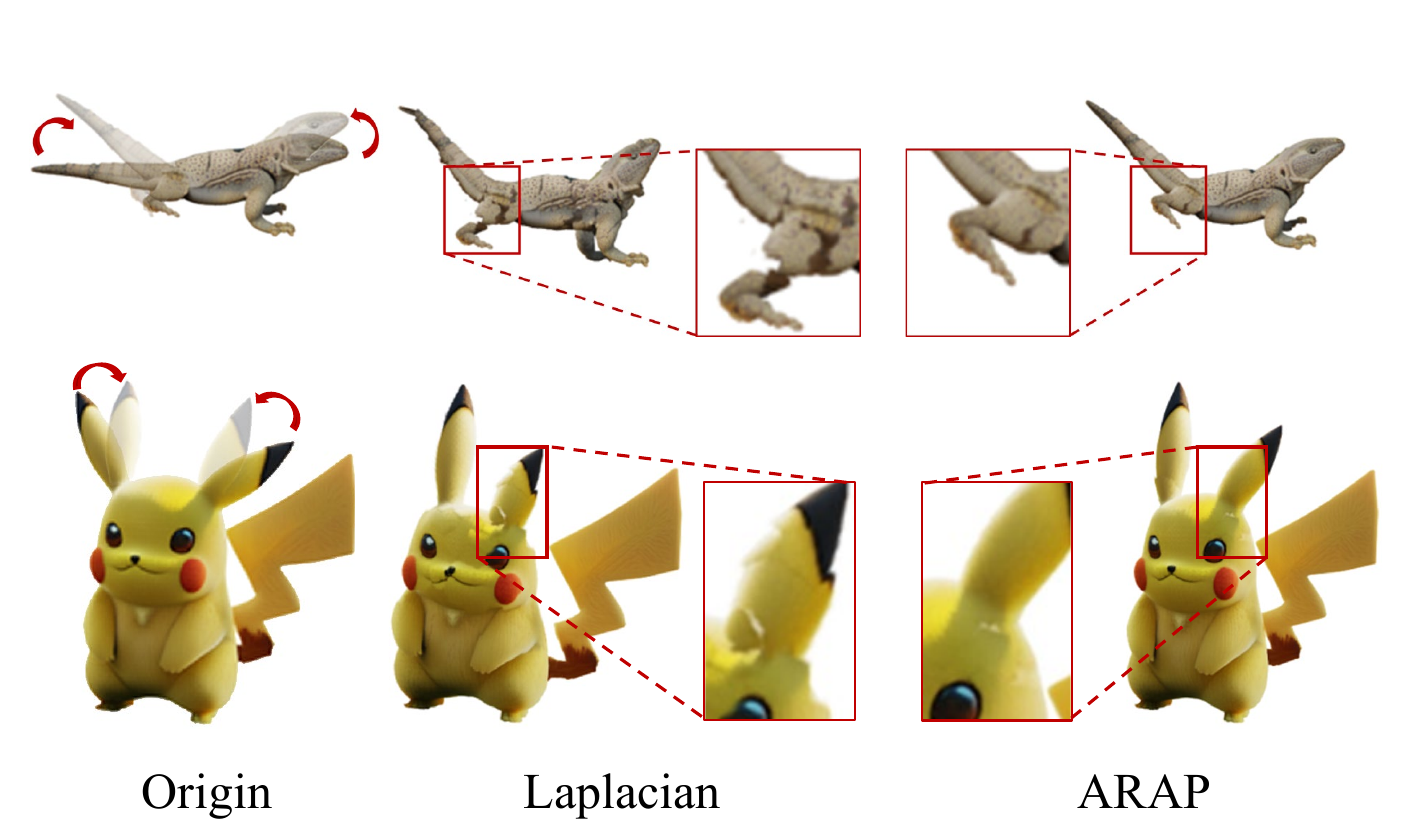} &

    \end{tabular}
    \vspace{-3mm}
    \caption{Ablation on the deformation constraints.}
    \label{fig:ablation_arap}
    \vspace{-5mm}
\end{figure}

\paragraph{Editing Constraints.}
We further evaluate the effectiveness of the editing constraints for mesh deformation. As shown in Fig.\ref{fig:ablation_arap}, using only Laplacian deformation, which optimizes Eq.\ref{eq:arap_energy} with fixed identity rotation $R_i$ and optimizable $p'_i$, often distorts the local geometry. This is particularly evident under large rotations. In contrast, the rotation-invariant ARAP constraint effectively maintains local rigidity and geometric details during deformation, resulting in more natural and plausible manipulation outcomes.

\section{Conclusion}

In this paper, we propose \name, a unified framework for image editing that integrates 3D geometric manipulation with real-time interaction, physically plausible deformation, and harmonious composition. Our pipeline utilizes an off-the-shelf foundational 3D generation model and 3DGS to achieve fast and high-fidelity rendering. Sparse deformation proxies and local rigidity constraints enable real-time, realistic deformation. Additionally, a Generative Composition module ensures seamless integration of edited components.
Extensive quantitative and qualitative evaluations show that our approach significantly outperforms state-of-the-art baselines. User studies further validate that our method provides superior adherence to user guidance compared to existing dragging-based editing approaches, while consistently maintaining high visual quality.

\section*{Acknowledgment}
The work has been supported by Hong Kong Research Grant Council - General Research Fund Scheme (Grant No. 17202422, 17212923, 17215025) Theme-based Research (Grant No.T45-701/22-R), and Strategic Topics Grant (Grant No.STG3/E-605/25-N). Part of the described research work is conducted in the JC STEM Lab of Robotics for Soft Materials funded by The Hong Kong Jockey Club Charities Trust.
\vspace{-2mm}
{
    \small
    \bibliographystyle{ieeenat_fullname}
    \bibliography{main}
}


\end{document}